\documentclass[10pt,twocolumn,letterpaper]{article}

\usepackage{iccv}              

\usepackage{algorithm}
\usepackage{algpseudocode}
\usepackage{graphicx}
\usepackage{amsmath}
\usepackage{amssymb}
\usepackage{booktabs}
\usepackage{xcolor,colortbl}
\usepackage{nicefrac, xfrac}

\usepackage{caption} 
\usepackage{arydshln} 
\usepackage{subcaption} 
\usepackage{tikz}
\usetikzlibrary{matrix,positioning}
\usetikzlibrary{calc,spy}
\tikzset{
  closeup/.style={
    opacity=1.0,
    height=1.6cm,
    width=1.6cm,
    connect spies, green,
  },
  largewindow/.style={
    red, line width=0.50mm
  },
  smallwindow/.style={
    blue, line width=0.20mm
  }
}
\usepackage{placeins}

\definecolor{iccvblue}{rgb}{0.21,0.49,0.74}
\usepackage[pagebackref,breaklinks,colorlinks,allcolors=iccvblue]{hyperref}

\def\paperID{12280} 
\def\confName{ICCV}
\def\confYear{2025}

\title{RayGaussX: Accelerating Gaussian-Based Ray Marching for Real-Time and High-Quality Novel View Synthesis\\[1em]
}

\author{Hugo Blanc, Jean-Emmanuel Deschaud, Alexis Paljic\\
Mines Paris, PSL University, Centre for robotics, 75006 Paris, France\\
{\tt\small \{hugo.blanc,jean-emmanuel.deschaud,alexis.paljic\}@minesparis.psl.eu}
}

\begin{document}

\twocolumn[{
  \renewcommand\twocolumn[1][]{#1}
  \maketitle
  \begin{center}
    \includegraphics[trim={1cm 6cm 1cm 3cm},clip,width=\linewidth]
      {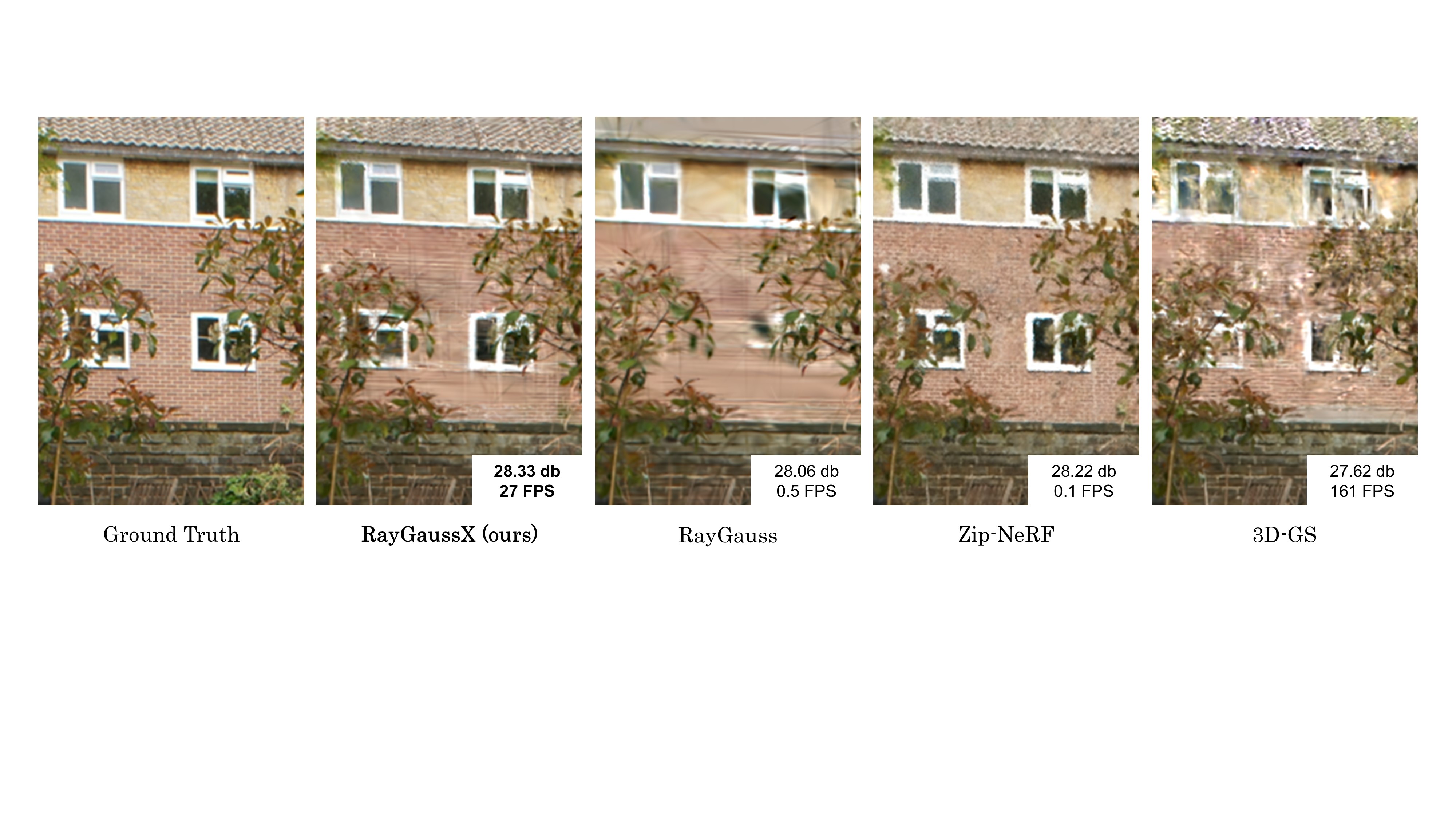}
    \captionof{figure}{Comparison of RayGaussX with RayGauss, Zip-NeRF and 3D-GS. Test image from the \textit{garden} scene of Mip-NeRF360.}
    \label{fig:figure1}
  \end{center}
}]

\begin{abstract}
RayGauss has achieved state‑of‑the‑art rendering quality for novel‑view synthesis on synthetic and indoor scenes by representing radiance and density fields with irregularly distributed elliptical basis functions, rendered via volume ray casting using a Bounding Volume Hierarchy (BVH). However, its computational cost prevents real-time rendering on real-world scenes. Our approach, RayGaussX, builds on RayGauss by introducing key contributions that accelerate both training and inference. Specifically, we incorporate volumetric rendering acceleration strategies such as empty-space skipping and adaptive sampling, enhance ray coherence, and introduce scale regularization to reduce false-positive intersections. Additionally, we propose a new densification criterion that improves density distribution in distant regions, leading to enhanced graphical quality on larger scenes. As a result, RayGaussX achieves 5× to 12× faster training and 50× to 80× higher rendering speeds (FPS) on real-world datasets while improving visual quality by up to +0.56 dB in PSNR. The associated code is available at: \href{https://github.com/hugobl1/raygaussx}{github.com/hugobl1/raygaussx}.
\end{abstract}

\section{Introduction}
\label{sec:intro}

Novel view synthesis (NVS) methods enable the generation of new images of a scene from a set of known viewpoints. This field has made significant progress with the introduction of Neural Radiance Fields (NeRF) \cite{Neural_Radiance_Fields}. This groundbreaking approach introduced a differentiable volumetric rendering equation coupled with a neural network to learn the scene's radiance fields. Subsequent research has further refined this method by enhancing both rendering quality and speed. One promising direction involves representing radiance and density fields using irregularly distributed local parametric functions. These approaches replace the computationally expensive neural networks used by NeRF with more localized structures that are less demanding and better suited to reconstruct the scene’s geometry and appearance. Among these approaches, two main categories can be identified according to their rendering algorithms. The first relies on a differentiable rasterization algorithm for local primitives, such as Gaussians, as in the 3D Gaussian Splatting approach~\cite{3D_Gaussian_Splatting}. This rasterization algorithm enables high rendering speeds. However, it can introduce artifacts such as flickering, which is inherent to the rasterization process. The second category employs the volumetric ray-marching rendering algorithm introduced by NeRF, providing a more exhaustive sampling of the space and being less prone to rendering artifacts. The recent RayGauss approach~\cite{raygauss} leverages this type of algorithm to render scenes represented with irregularly distributed elliptical basis functions, achieving state-of-the-art graphical results on synthetic and indoor scenes. However, its performance slightly degrades in outdoor environments, and its training and inference times remain significant for complex scenes. Thus, this paper aims to address these limitations and enhance the method's applicability to a broader range of scenarios. To achieve this, we focus on key aspects to improve computational efficiency. First, we draw inspiration from acceleration strategies in classic volumetric ray marching, such as empty‐space skipping and adaptive sampling, to speed up the rendering equation computation. Then, by analyzing the implementation specifics, we introduce a reordering of primitives and rays to better adapt the algorithm for parallel computation on GPUs, enhancing ray coherence and memory‐access efficiency. Furthermore, we propose a scale‐regularization function to minimize false‐positive intersections. Finally, to improve rendering quality in large-scale scenes, we present a new densification method that evenly increases the density of Gaussians in the scene, regardless of their distance from the camera. Thus, our contributions can be summarized as follows:
\begin{itemize}
  \item Integrating empty-space skipping and adaptive sampling strategies to accelerate the evaluation of the rendering equation.
  \item Enhancing ray coherence by reordering rays and primitives for efficient GPU parallelization.
  \item Introducing a scale-regularization function to mitigate false-positive intersections.
  \item Formulating a novel criterion for improving densification in distant regions.
\end{itemize}

\begin{figure*}[ht]
  \centering
   \includegraphics[trim={0.5cm 1cm 0.5cm 1cm},clip,width=\linewidth]{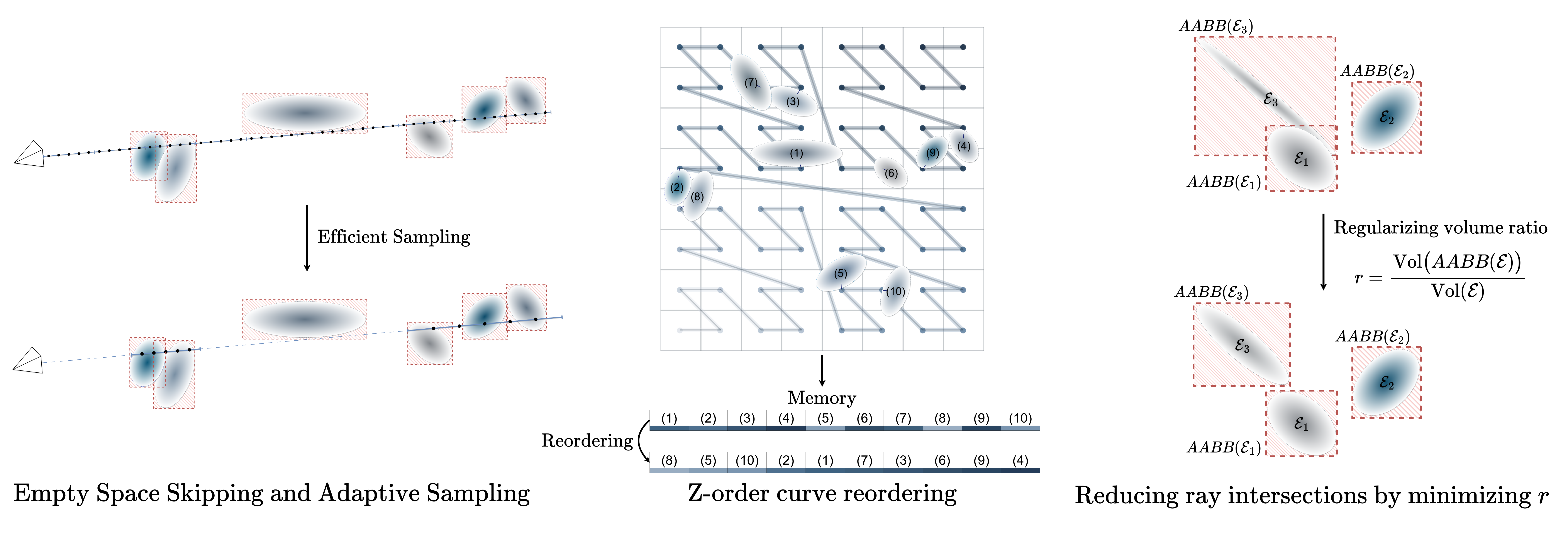}
      \caption[Main contributions]{Main contributions of RayGaussX approach to improve training and rendering speed.}
   \label{fig:figure2}
\end{figure*}
\section{Related Work}
\label{sec:related_work}
Novel view synthesis has been studied in the computer vision community for decades, with early approaches known as Image-Based Rendering (IBR), including~\cite{levoy1996}, which interpolate new images from the closest known views. Another strategy involves using a proxy, such as a mesh~\cite{debevec1998}, to enhance interpolation or even leveraging neural networks~\cite{Hedman2018}. Point clouds can also serve as proxies, as seen in NPBG~\cite{NPBG}, NPBG++~\cite{NPBG++}, and ADOP~\cite{ADOP}. However, these image-space interpolation methods suffer from artifacts and flickering.

In 2020, NeRF~\cite{Neural_Radiance_Fields} showed that scene density and radiance fields can be encoded within a Multi-Layer Perceptron (MLP), with its weights learned thanks to differentiable volumetric rendering. Numerous improvements have made NeRF faster, such as using grids of small MLPs in KiloNeRF~\cite{KiloNeRF}, caching in FastNeRF~\cite{FastNeRF}, dense voxel grids~\cite{Direct_Voxel_Grid_optimization}, sparse voxel structures~\cite{NSVF,Baking}, octree-based representations in PlenOctrees~\cite{PlenOctrees}, or hash tables in Instant-NGP~\cite{Instant_NGP}. Other approaches have improved rendering quality by addressing aliasing, as in Mip-NeRF~\cite{Mip_NeRF} and Mip-NeRF360~\cite{Mip_NeRF_360}. More recently, Zip-NeRF~\cite{zipnerf}, which combines Mip-NeRF360 and Instant-NGP, has set the state of the art in rendering quality but suffers from slow training and does not allow interactive rendering. Furthermore, Plenoxels~\cite{Plenoxels} demonstrated that NeRF’s key aspect is the differentiability of the rendering equation. Instead of using a network to encode scene density and color parameters, they employ a voxel grid with spherical harmonics. Since then, other data structures have been explored, including point clouds in Point-NeRF~\cite{Point_NeRF}, PointNeRF++~\cite{sun2023pointnerfpp}, and tetrahedra in Tetra-NeRF~\cite{Tetra_NeRF}.

DSS~\cite{DSS2019}, Point-Based Radiance~\cite{Differentiable_Point_Based}, and the recent, widely adopted 3D Gaussian Splatting (3D-GS)~\cite{3D_Gaussian_Splatting} also use point clouds to encode scenes. However, instead of volumetric ray marching, they use a rasterization-based approximation: splat-based differentiable rendering enables training and rendering speeds several orders of magnitude faster than NeRF-like methods. Many approaches have since built upon 3D-GS, including Mip-Splatting~\cite{Yu2023MipSplatting} for anti-aliasing, Scaffold-GS~\cite{Scaffold-GS} using anchors to better regularize Gaussian parameters, and Spec-Gaussian~\cite{Spec-gaussian}, which incorporates a MLP to better model specular and anisotropic colors. However, rasterization‑based radiance‑field rendering suffers from two key limitations~\cite{Gaussians_accurate}: flickering arising from image‑space primitive ordering, which induces popping artifacts~\cite{stopthepop}, and inherent approximations in the rendering equation (each primitive contributes only once per pixel, irrespective of its size). Furthermore, rasterization algorithms struggle to accommodate complex camera models and to simulate a broader range of physical phenomena, such as volumetric scattering.

A final category relies on rendering primitives defined by radial basis functions (RBFs) via volumetric ray casting. NeuRBF~\cite{chen2023neurbf} uses a hybrid representation with adaptive RBFs for fine
details, but is limited to a small number of primitives per sample along a ray. More recently, 3DGRT~\cite{3dgrt2024}, RayGauss~\cite{raygauss} and EVER~\cite{EVER} have leveraged a Bounding Volume Hierarchy (BVH) structure and the OptiX library for fast ray casting in a scene using Gaussian-based primitives. RayGauss~\cite{raygauss} showed state-of-the-art rendering quality on synthetic datasets like the NeRF dataset~\cite{Neural_Radiance_Fields}, but suffers from slow training and rendering on real-world datasets.

Our approach, RayGaussX, extends RayGauss~\cite{raygauss} by significantly reducing training and rendering times through key contributions that exploit irregularly distributed elliptical primitives rendered via volume ray marching.
\section{RayGaussX approach}
\subsection{Preliminary: gaussian-based ray-marching}
The RayGauss method offers a physically consistent formulation of the emitted radiance \( c \) and density \( \sigma \), expressed through a decomposition using Gaussian functions combined with Spherical Gaussians/Harmonics (SG/SH) for an all-frequency colorimetric representation. Specifically, it represents the density \( \sigma \) and radiance \( c \) as:

\begin{equation}
\begin{aligned}
    \sigma(\mathbf{x}) &= \sum_{l=1}^N \tilde{\sigma}_l \mathcal{G}(\mathbf{x}; \mu_l, \mathbf{q}_l, \mathbf{s}_l) \\
    c(\mathbf{x}, \mathbf{d}) &= \frac{\sum_{l=1}^N c_l({\mathbf{d}}) \tilde{\sigma}_l \mathcal{G}(\mathbf{x}; \mu_l, \mathbf{q}_l, \mathbf{s}_l)}{\sum_{m=1}^N \tilde{\sigma}_m \mathcal{G}(\mathbf{x}; \mu_m, \mathbf{q}_m, \mathbf{s}_m)}
\end{aligned}
\end{equation}
where \( c_l \) is the radiance emitted by the \( l \)-th primitive, parameterized with SH/SG, \( \tilde{\sigma}_l \) is a density parameter of the \( l \)-th primitive, and \( \mathcal{G} \) is an unnormalized Gaussian. To render these fields, RayGauss relies on volumetric rendering theory, expressing the interaction of light with the medium through the classical volumetric rendering equation:

\begin{equation}
\begin{aligned}
C(\mathbf{r}) &= \int_{t_n}^{t_f} c(\mathbf{r}(t), \mathbf{d}) \sigma(\mathbf{r}(t)) T(t) \, dt,\\
T(t) &= e^{- \int_{t_n}^{t} \sigma(\mathbf{r}(s)) \, ds}
\end{aligned}
\label{eq:volrender}
\end{equation}
To numerically compute this integral, the usual approach subdivides it into uniform segments and computes an approximation for each segment, yielding the Volume Ray Marching equation used in recent works~\cite{raygauss,Neural_Radiance_Fields}:
\begin{equation}
\begin{aligned}
C(\mathbf{r}) &= \sum_{i=0}^{N} \left(1 - \exp(-\sigma_i \Delta t)\right) c_i T_i \\
T_i &= \exp \left( -\sum_{j=0}^{i-1} \sigma_j \Delta t \right)
\label{eq:vol_raycasting}
\end{aligned}
\end{equation}
where the step size \(\Delta t\) is constant. In the RayGauss approach, this equation is efficiently computed using an algorithm that leverages a Bounding Volume Hierarchy (BVH) to rapidly compute ray-primitive intersections. 

In this section, we present our main contributions aimed at improving this algorithm: empty-space skipping, adaptive sampling, enhanced ray coherence through ray and primitive reordering, scale regularization to reduce false positive intersections, and finally, a novel densification criterion improving rendering quality in distant regions.

\subsection{Efficient Ray Sampling: Empty Space Skipping and Adaptive Sampling} 
As illustrated by Eq.~\eqref{eq:vol_raycasting}, the computational cost of evaluating the color per ray is directly related to the number of samples $N$.
Thus, to reduce computation time, a key approach is to minimize the samples needed for ray color computation. Two main types of approaches can be distinguished: the first, empty space skipping, aims to avoid sampling completely transparent regions with null density. Since these regions do not contribute to the integration, such methods do not affect rendering quality. A second type of approach consists in adjusting the sampling step based on the integrated space region, focusing samples on areas that contribute the most to color computation. These two types of approaches are complementary, and in the following, we introduce advancements in both areas that accelerate the computation times of the RayGauss approach.
\paragraph{Empty Space Skipping:}As previously discussed, the objective is to design a method that avoids sampling in regions with zero density. A key advantage of RayGauss~\cite{raygauss} is its representation based on sparsely distributed truncated Gaussians, which explicitly encode regions of non-zero density. In contrast, other representations, such as those based on neural networks, do not define these regions explicitly. Additionally, RayGauss stores these primitives in a Bounding Volume Hierarchy (BVH) for efficient intersection computations. Ray integration is performed sequentially: at step \( i \), a segment \( S_i = [t_i, t_i + N_s \cdot \Delta t] \) containing a fixed number \( N_s \) of samples is considered. The BVH is used to compute the intersection of the scene's Gaussians with \( S_i \). If no intersection is found, the next segment is evaluated; otherwise, the contribution of this segment to the color is computed. Thus, at the segment level, RayGauss can detect whether the region is empty. Our approach enhances this empty space skipping by leveraging the precomputed BVH, which is already required by the algorithm. 
\definecolor{softblue}{RGB}{100, 150, 255} 
\algnewcommand{\LineComment}[1]{\State \(\triangleright\) #1}
\definecolor{softred}{RGB}{220, 80, 80} 

\algnewcommand{\New}[1]{\State \textcolor{softblue}{#1}} 
\begin{algorithm}[h]
\small
\caption{Volume Ray Marching}
\textbf{Input:}  $(o,d)$: ray origin/direction, $\Delta t$: step size, $N_s$: buffer size, $T_{\epsilon}$: transmittance threshold, $(t_n,t_f)$: integration bounds, $\mathcal{H}_b$: hit primitive index buffer, $\mathcal{G}_p$: gaussians parameters
 \\
\textbf{Output:}  $C_R$
\begin{algorithmic}[1]
    \State $\Delta S\gets \Delta t \times N_s$ \Comment{Segment size}
    \State $T \gets 1.0$ \Comment{Ray transmittance}
    \State $C_R \gets (0.0,0.0,0.0)$ \Comment{Ray color}
        \New{$t_{min} \gets \text{BVHClosestHit}(o, d, t_n, t_f)$}
        \New{$t_S \gets t_{min}$} \Comment{Segment distance along the ray }
        
        \While{$t_S < t_f$ \textbf{and} $T > T_{\epsilon}$}
            \State $n_p \gets 0$ \Comment{Number of primitives}
            \LineComment{Collect the intersected primitives}
                \State $(\mathcal{H}_b,n_p) \gets \text{BVHTraversal}(o, d, t_S, t_S+\Delta S)$
                \If{$n_p == 0$}
                    \State $t_S \gets t_S + \Delta S$
                    \LineComment{Recompute closest hit}
                    \New{$t_S \gets \text{BVHClosestHit}(o, d, t_S, t_f)$}
                    \State \textbf{continue}
                \EndIf
                
                \LineComment{Update ray color and transmittance}
                \State $(C_R, T)\gets\text{RayUpdate}(n_p, \Delta t, t_S, o, d, \mathcal{G}_p,\mathcal{H}_b,C_R, T)$
            \State $t_S \gets t_S + \Delta S$
        \EndWhile
\end{algorithmic}
\label{alg:ray_march}
\end{algorithm}
Our method follows the procedure described in Algorithm \ref{alg:ray_march}. Ray tracing libraries, such as the OptiX API used by RayGauss, are designed to efficiently compute the closest intersection point for a given ray, as this is a fundamental operation in classical ray tracing algorithms. Our idea is to leverage this by alternating between two types of processes during scene traversal. On the one hand, the ClosestHit shader computes the nearest intersection point with primitives for a ray constrained to a given region, e.g., \( [t_n, t_f] \). On the other hand, the AnyHit shader (referred to as BVHTraversal in line 9 of Alg.~\ref{alg:ray_march}), already implemented in RayGauss~\cite{raygauss}, collects all primitives intersected along the segment.
Thus, starting from the initial integration bound \( t_n \), we first compute the closest hit with the primitives at \( t_{\min} \), which allows us to avoid sampling empty regions between the camera and the closest primitive. Indeed:
\begin{equation}
\begin{split}
    \sigma(\mathbf{r}(t)) = 0, \quad \forall t \in [t_n, t_{\min}], \\
    \int_{t_n}^{t_{\min}} c(\mathbf{r}(t), \mathbf{d}) \sigma(\mathbf{r}(t)) T(t) \, dt = 0.
\end{split}
\end{equation}
Once the closest intersection point is found, we begin the integration process as described in~\cite{raygauss}. However, other large empty regions may still exist along \( [t_{\min}, t_f] \). Moreover, the AnyHit shader not only collects primitives but also determines their number. If no primitives are detected within the current segment \( [t_S, t_S + \Delta S] \), it indicates that the region has zero density. In this case, the ClosestHit program is invoked again to locate the next point of interest. This approach effectively avoids exhaustive sampling of empty regions while adapting to the specific requirements of the Volume Ray Casting algorithm developed in~\cite{raygauss}.

\paragraph{Adaptive Sampling:}A second approach to accelerating the computation of the volumetric rendering equation is to adapt the spatial subdivision of the integral to reduce the number of samples. Several studies have explored this problem. Danskin et al.~\cite{10.1145/147130.147155} leveraged importance sampling principles to construct a pyramidal structure enabling adaptive sampling. More recently, Morrical et al.~\cite{8933539} employed a kd-tree to adjust the sampling step based on the variance within the subpartitions. Campagnolo et al.~\cite{7314541} proposed an adaptive Simpson integration scheme to tackle this problem. Thus, adaptive sampling strategies are designed to align with the specific constraints of the scene representation and rendering algorithm to determine an adaptive step size. In our case, two criteria are important. During training, the structure is updated at each optimization step, meaning that an approach storing the sampling criterion in an auxiliary structure, as in Morrical et al.~\cite{8933539}, would require frequent reconstruction, increasing computational cost. Furthermore, the RayGauss rendering algorithm sequentially accumulates properties segment by segment, with each segment containing multiple samples. To retain the algorithm's advantages, we adopt adaptive subdivision at the segment level while keeping a fixed number of samples $N_S$ per segment for improved ray coherence. Since our primary goal is to accelerate rendering, we introduce a criterion with minimal computational overhead. Danskin et al.~\cite{10.1145/147130.147155} suggest that adaptive sampling should depend on transmittance. Here, we define a criterion using the transmittance at the end of \((i-1)\)-th segment to determine the size of the \(i\)-th segment. The idea is that for a low resulting transmittance, the subsequent segments will contribute less to the rendering integral, allowing for a larger step size. Furthermore, we also make the sampling criterion dependent on the distance to the camera, assuming that distant regions require less precision than closer ones.

Consequently, the size of the \(i\)-th segment, \(\Delta S_i\), can be dynamically adapted using a criterion based on the current distance from the origin, \(d_i = \|s_i - o\|\), and the transmittance, \(T_i = T(s_i)\), where \(s_i\) is the position at the beginning of the segment and \(o\) is the camera origin:
\begin{equation}
\Delta S_i = N_S\!\cdot\!\min\! \left( \max\! \left(\frac{d_i}{\beta},\Delta t_{\min}\right)\!\cdot\! T_i^{-\tfrac{1}{3}},\Delta t_{\max}\right)
\end{equation}
where \(\beta\) quantifies the variation of the step size with respect to the distance from the origin, \(\Delta t_{\min}\) is the minimum step size and \(\Delta t_{\max}\) is the maximum step size to prevent excessively large variations in the sampling step.

\subsection{Optimizing Ray Coherence and Memory Access Efficiency}
This section explores improvements enabled by the specifics of parallel GPU computation, particularly through enhanced ray coherency, ensuring more consistent data access patterns and execution paths.

\paragraph{Enhancing Memory Access Efficiency through Spatial Reordering of Gaussians:}
The Volume Ray Marching algorithm from~\cite{raygauss} efficiently renders irregularly distributed Gaussians using the OptiX API. Scene parameter optimization and inference are performed per image, which is significant because rays from the same image tend to traverse spatially adjacent regions, a property not guaranteed if originating from different viewpoints. Consequently, during rendering, these rays pass through contiguous spatial areas and intersect nearby Gaussians. In this context, it is important to note that global memory access, where Gaussian parameters are stored, benefits from spatial locality in modern frameworks like CUDA, used by the OptiX API. Memory accesses occur via the transfer of contiguous blocks, and spatial locality in memory enables more efficient coalescing and higher data throughput. Moreover, spatially adjacent Gaussians are accessed simultaneously, suggesting that grouping their parameters in memory and reordering them based on their spatial distribution could enhance performance by improving memory coalescing.

\begin{figure}[htb]
  \centering
  \begin{subfigure}[b]{0.48\columnwidth}
    \centering
    \includegraphics[trim={3cm 0cm 0cm 0cm},clip,width=\linewidth]{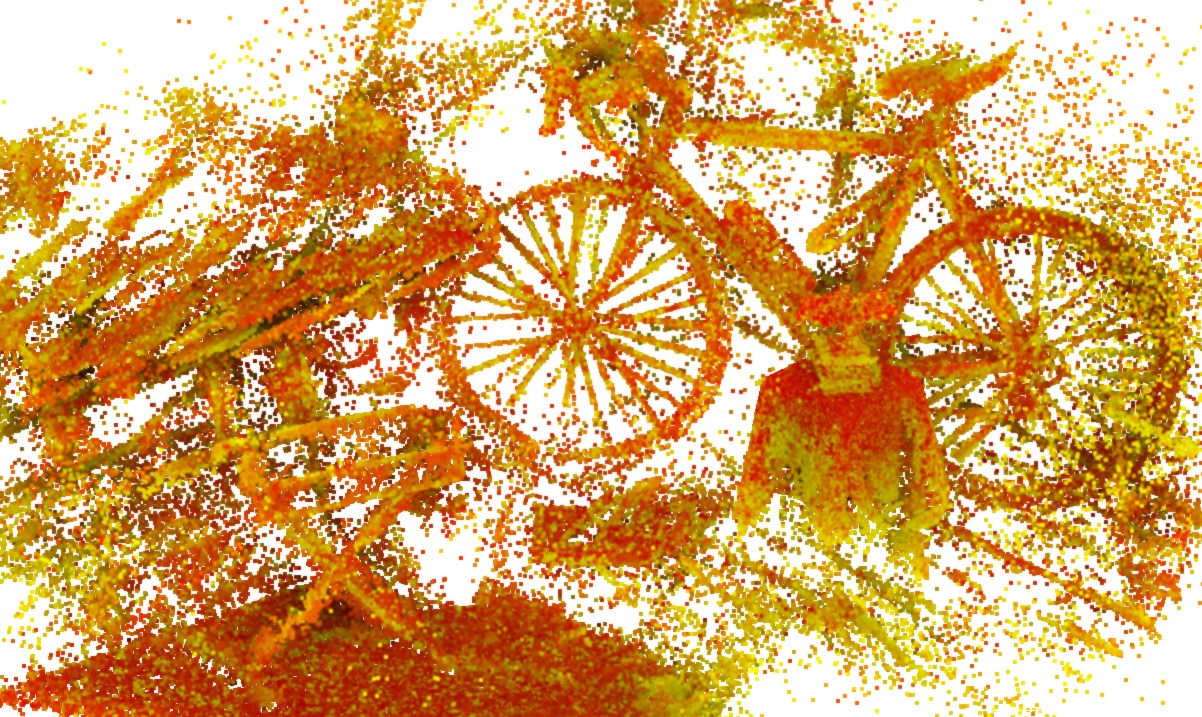}
    \caption{Default indexing}
    \label{fig:bonsai_default}
  \end{subfigure}
  \hfill
  \begin{subfigure}[b]{0.48\columnwidth}
    \centering
    \includegraphics[trim={3cm 0cm 0cm 0cm},clip,width=\linewidth]{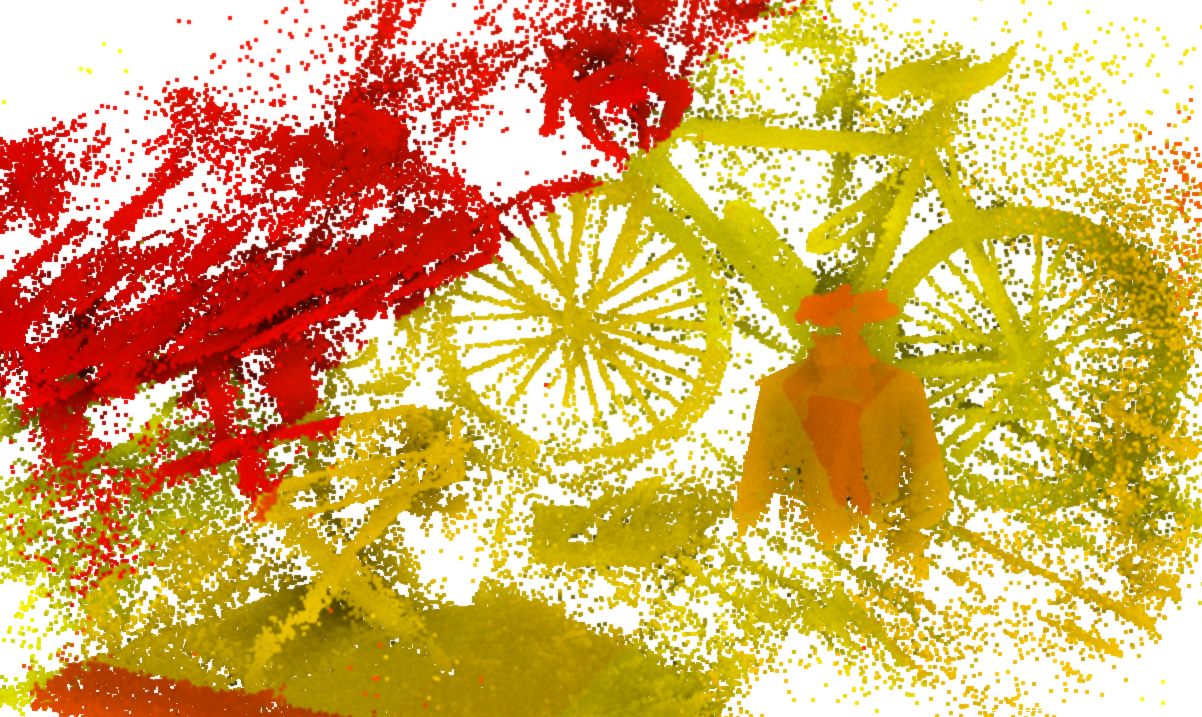}
    \caption{Z-order curve indexing}
    \label{fig:bonsai_zorder}
  \end{subfigure}
  \caption{Comparison of the \textit{bonsai} scene from the Mip-NeRF360 dataset: the left image shows the scene with default indexing, while the right image uses indexing obtained via a Z-order curve. Note that the color scale ranges from yellow to red, and that colorimetric similarity corresponds to a proximity of indices, hence contiguous memory allocation.}
  \label{fig:bonsai_comparison}
\end{figure}

The objective is to efficiently reorder data, so that spatially proximate Gaussians are stored in contiguous memory regions. To achieve this, we employ Z-order curve theory, which maps multidimensional integer data point to an integer index while preserving data locality. This means that points that are close together in the multidimensional space are more likely to have nearby indices \cite{morton1966}. 
In practice, the Gaussian mean serves as an indicator of its spatial position, which is normalized, quantized into an integer, and mapped to a Z-order index. The Gaussians are then reordered in ascending Z-index order, allowing spatially close Gaussians to be stored in neighboring memory regions (illustrated in Fig.~\ref{fig:figure2} and applied to the \textit{bonsai} scene in Fig.~\ref{fig:bonsai_comparison}).

\paragraph{Improving Ray Coherence:}In the classical parallel computing model supported by CUDA \cite{cuda_ppm}, computations are performed in parallel by threads. In the CUDA architecture, these threads are organized into groups of 32, called warps. Within a warp, all threads execute the same instruction simultaneously, although each may process different data. Maximum efficiency is achieved when all 32 threads follow the same execution path. However, if divergence occurs, such as through a data-dependent conditional branch, the warp must sequentially execute each path taken, temporarily disabling inactive threads. While necessary for handling different execution scenarios, this mechanism significantly degrades performance by reducing parallelism efficiency. This phenomenon is known as warp divergence.

If each ray is processed in parallel, minimizing warp divergence requires maximizing coherence among rays within a warp, ensuring that as many rays as possible execute the same operations on similar data. However, achieving this coherence depends on how rays are organized. RayGauss~\cite{raygauss} does not perform any specific ray ordering, leading to an initial grouping in scanline order. Without prior scene information, dividing the image into tiles and grouping rays accordingly generally reduces variance among rays compared to a scanline-order approach. Rays within the same tile typically sample similar regions (and thus similar Gaussians), whereas scanline ordering may group rays observing significantly different parts of the scene, increasing divergence. In practice, we leverage the OptiX API, which transparently manages this initial distribution by launching rays on a 2D grid. Although threads may later be dynamically reordered, making the process more opaque, this approach provides an effective initial grouping.

\subsection{Limiting Highly Anisotropic Gaussian}
We observe that differentiable rendering algorithms relying on Gaussian kernels often produce highly anisotropic Gaussians, with substantial variations in their scale values (e.g., \( s_{\text{min}} \ll s_{\text{max}} \)). This phenomenon negatively impacts the computational efficiency of RayGauss~\cite{raygauss}. Indeed, the algorithm leverages a BVH for efficient intersection computations. However, for custom primitives like ellipsoids, an Axis-Aligned Bounding Box (AABB) is first constructed around each ellipsoid, and a BVH is then built around these AABBs. Subsequently, during intersection computations, the optimized OptiX API first computes the intersection with the AABB and then calls the user-code intersection program for the ellipsoid. Here, the smallest enclosing AABB is used for each ellipsoid. If this AABB's volume is significantly larger than that of the ellipsoid, more false-positive calls to the user-code intersection program occur.
This slows down BVH traversal and is particularly problematic in the context of hardware-accelerated ray intersection~\cite{wald_owltubes_20}. Thus, to accelerate rendering, reducing false-positive hits is beneficial. Also, their occurrence is correlated with the following volume ratio:

\begin{equation}
    r=\frac{\text{Vol}(\text{AABB}(\mathcal{E}))}{\text{Vol}(\mathcal{E})}
\end{equation}
with \(\text{Vol}\) the volume function, \(\mathcal{E}\) a given ellipsoid, and \(\text{AABB}(\mathcal{E})\) its associated axis-aligned bounding box. When \( r \) is high, many false-positive intersections occur. Conversely, when \( r \) is low, false-positive intersections decrease. To ensure a rotation-invariant criterion and avoid bias toward any specific Gaussian orientation, we constrain the worst-case scenario with respect to rotation by computing an upper bound. (see supplementary material for details):

\begin{equation}
\frac{\text{Vol}(\text{AABB}(\mathcal{E}))}{\text{Vol}(\mathcal{E})}
\;\le\;
\frac{2}{\pi\,\sqrt{3}}
\;\frac{\bigl(s_{1}^{2} + s_{2}^{2} + s_{3}^{2}\bigr)^{\tfrac{3}{2}}}{s_{1}\,s_{2}\,s_{3}}
\end{equation}
Thus, to minimize false-positive intersections, we add a loss term (hereafter referred to as the isotropic loss) that constrains this ratio for all Gaussians:
\begin{equation}
L_s = \frac{1}{N} \sum_{l}(\max \left( 
    r_{\mathrm{max}, l}, 
    r_0 
\right) - r_0)
\end{equation}
with:
\begin{equation}
    r_{\mathrm{max}, l}=\frac{2}{\pi\,\sqrt{3}}
\;\frac{\bigl(s_{l,1}^{2} + s_{l,2}^{2} + s_{l,3}^{2}\bigr)^{\tfrac{3}{2}}}{s_{l,1}\,s_{l,2}\,s_{l,3}}
\end{equation}
and ${r_0}$ is a threshold ratio beyond which a penalty is applied. Then, the total loss function is: $ L =  (1 - \lambda)\cdot L_{1} + \lambda \cdot \text{DSSIM}+\lambda_s L_s$ 
where $\lambda_s$ is a fixed hyperparameter. 

\subsection{A Novel Densification Criterion}
RayGauss densification process \cite{raygauss} is inspired by the adaptive density control heuristic introduced in 3D Gaussian Splatting~\cite{3D_Gaussian_Splatting}. This heuristic enables cloning or splitting the Gaussians in the scene by evaluating a criterion every $I$ iterations. For a Gaussian $l$ with mean position $\mu_l$, let $\mu_l^P$ be its projection in normalized device coordinates (NDC). If this Gaussian has been observed in $I_l < I$ images during recent iterations, the heuristic determines whether to densify it by comparing the average norm of the loss gradient with respect to $\mu_l^P$ across these $I_l$ images to a given threshold $\tau$:
\begin{equation}
    \label{eq:densification_3Dsplat}
    \frac{1}{I_l} \sum_{i=1}^{I_l} \left\| \nabla_{\mu_l^{P}} L_i \right\| > \tau.
\end{equation}
where $L_i$ is the loss computed for the $i$-th image in which Gaussian $l$ appears. RayGauss does not compute $\mu_l^P$, since rendering uses volumetric ray casting rather than projection and splatting, and replaces the densification criterion with:
\begin{equation}
    \label{eq:densification_ray_gauss}
    \frac{1}{I_l} \sum_{i=1}^{I_l} \left\| \nabla_{\mu_l} L_i \right\| > \tau.
\end{equation}
where the gradient with respect to $\mu_l$ is computed during optimization. However, a limitation of the RayGauss formulation (Eq.~\ref{eq:densification_ray_gauss}) is that it introduces a dependency on the distance between the Gaussian and the camera. Specifically, distant Gaussians are more likely to exhibit low gradients compared to closer ones, even if their contributions to the image are similar. This occurs because a slight deviation of the mean in world space results in a smaller variation in the image contribution of a distant Gaussian compared to a closer one, due to perspective projection. As a result, Gaussians farther from the camera exhibit lower gradients and are less likely to be densified. This may explain the lower performance of RayGauss on outdoor scenes compared to indoor environments. In contrast, the original densification criterion from~\cite{3D_Gaussian_Splatting} does not suffer from this issue, as it relies on the view-space position gradient. This prevents the criterion from being biased by the Gaussian’s distance to the camera. Indeed, a slight deviation in view space does not favor closer Gaussians over distant ones. For this reason, we modify the criterion from the RayGauss approach~\cite{raygauss} by introducing a corrective factor that weights the gradients in 3D space, ensuring densification of distant Gaussians:
\begin{equation}
    \label{eq:new_densification_ray_gauss}
    \frac{1}{I_l} \sum_{i=1}^{I_l} \alpha_i \left\| \nabla_{\mu_l} L_i \right\| > \tau.
\end{equation}
where \( \alpha_i = \frac{\left\| \mu_l - o_i \right\|}{f} \), with \( o_i \) being the center of camera \( i \), and $f$ the focal distance. Empirically, we observe that $\nabla_{\mu_l} L_i \cdot \frac{\mu_l - o_i}{\|\mu_l - o_i\|}$ is generally negligible compared to its tangential component. Under this approximation, our criterion corresponds to averaging the norms of the gradients with respect to the mean projected onto the sphere of radius $f$ (see supplementary for details). This criterion therefore remains closer to the original criterion of 3D Gaussian Splatting, accounts for the scaling factor, and is efficiently computable.

\section{Experiments and Results}
\label{sec:results}

\subsection{Experiments}

The experiments are conducted on two synthetic datasets, the NeRF Synthetic dataset~\cite{Neural_Radiance_Fields} and the NSVF Synthetic dataset~\cite{NSVF}, as well as on three real-world datasets: Mip-NeRF360~\cite{Mip_NeRF_360}, Tanks\&Temples~\cite{Knapitsch2017}, and Deep Blending~\cite{Hedman2018}. We evaluate our approach using standard image synthesis metrics, including rendering quality (PSNR, SSIM, and LPIPS computed with a VGG network), training time (Train), and rendering speed (FPS, Frames Per Second). Our RayGaussX approach is compared to the RayGauss method~\cite{raygauss} as well as to state-of-the-art novel view synthesis techniques. Specifically, we include Zip-NeRF~\cite{zipnerf}, which achieves the highest visual quality but requires a long training time and does not support real-time rendering, and 3D-GS~\cite{3D_Gaussian_Splatting}, a rasterization-based Gaussian rendering method that offers the fastest performance. Additionally, we compare against more recent variants of 3D-GS: Mip-Splatting~\cite{Yu2023MipSplatting}, which addresses aliasing, and Spec-Gaussian~\cite{Spec-gaussian}, which incorporates an MLP network to better model high-frequency color details. All methods were retrained from their official repositories on a single NVIDIA RTX 4090 to ensure a fair comparison of training and rendering performance (see supplementary).

\subsection{Implementation details}

The training parameters for the RayGaussX method are identical to those used for the RayGauss method~\cite{raygauss}. The only differences concern the following contributions:

\begin{itemize}
    \item \textbf{Sampling strategy:} We use \textbf{uniform sampling} with $\Delta t = 0.0025$ for synthetic datasets (same as RayGauss), whereas for real-world datasets, we adopt \textbf{adaptive sampling} with $\Delta t_{\min} = 0.005, \quad \Delta t_{\max} = 4 \times \Delta t_{\min}, \quad \beta = 1024$. In contrast, to benchmark RayGauss, we kept a fixed $\Delta t = 0.005$ for real-world datasets.
    \item \textbf{Loss for isotropic Gaussians:} This loss is applied to real-world datasets with the following hyperparameters $\lambda_s = 0.00025, \quad r_0 = 10.$
    \item \textbf{New densification:} We use the parameter $\tau = 0.00015.$ in all experiments.
\end{itemize}

No major hyperparameters are introduced by the \textbf{Empty Space Skipping}, \textbf{Ray Coherence}, or \textbf{Memory Reordering of Gaussians} techniques.

\subsection{Results on synthetic datasets}

Tab.~\ref{tab:blender} and Tab.~\ref{tab:nsvf} present the results of our RayGaussX approach in comparison with RayGauss~\cite{raygauss} and other novel view synthesis methods on the synthetic NeRF~\cite{Neural_Radiance_Fields} and NSVF~\cite{NSVF} datasets. All methods were trained with a white background, a standard practice for comparisons on synthetic data. 

It can be observed that for both datasets, RayGaussX achieves the best rendering quality (slightly superior to RayGauss) while offering a rendering speed \textbf{three times faster} in terms of FPS. For synthetic datasets, we applied empty space skipping, z-curve indexing, ray reordering, and uniform sampling (to maximize rendering quality), which already provide significant improvements in rendering speed.

\begin{table}[h]
\centering
\resizebox{\columnwidth}{!}{%
\begin{tabular}{l | c | c | c | c | c }
\multicolumn{6}{c}{\textbf{NeRF Synthetic}} \\
\textbf{Method $\backslash$ Metric}  & PSNR$\uparrow$ & SSIM$\uparrow$ & LPIPS$\downarrow$ & Train$\downarrow$ & FPS$\uparrow$ \\
\hline
Instant-NGP~\cite{Instant_NGP} & 33.18 & 0.963 & 0.045 & - & - \\ 
Mip-NeRF360~\cite{Mip_NeRF_360} & 33.24 & 0.961 & 0.042 & - & - \\
Point-NeRF~\cite{Point_NeRF} & 33.30 & 0.962 & 0.049 & - & - \\
3D-GS~\cite{3D_Gaussian_Splatting}* & 33.39 & 0.968 & 0.031 & 4min & 681 \\
Zip-NeRF~\cite{zipnerf}* & 33.69 & \cellcolor{red!40}0.974 & \cellcolor{yellow!40}0.028 & 445min & 0.3 \\
Mip-Splatting~\cite{Yu2023MipSplatting}* & 33.74 & \cellcolor{orange!40}0.971 & 0.029 & 5min & 468 \\
Spec-Gaussian~\cite{Spec-gaussian}* & 33.80 & \cellcolor{yellow!40}0.970 & \cellcolor{orange!40}0.027 & 11min & 142 \\
3DGRT~\cite{3dgrt2024}* & 33.86 & \cellcolor{yellow!40}0.970 & 0.037 & 9min & 304 \\
NeuRBF~\cite{chen2023neurbf} & \cellcolor{yellow!40}34.47 & \cellcolor{red!40}0.974 & 0.035 & - & $\simeq$1 \\
\hdashline
RayGauss~\cite{raygauss}* & \cellcolor{orange!40}34.53 & \cellcolor{red!40}0.974 & \cellcolor{red!40}0.024 & 33min &  \textbf{25} \\
\hline
RayGaussX (ours) & \cellcolor{red!40}34.54 & \cellcolor{red!40}0.974 & \cellcolor{red!40}0.024 & 22min &  \textbf{85} \\
\end{tabular}
}
\caption{Results on NeRF Synthetic dataset~\cite{Neural_Radiance_Fields}. Methods marked * were retrained on an NVIDIA RTX4090 for fair comparison of training and rendering (FPS) times.}
\label{tab:blender}
\end{table}

\begin{table}[h]
\centering
\resizebox{\columnwidth}{!}{%
\begin{tabular}{l | c | c | c | c | c }
\multicolumn{6}{c}{\textbf{NSVF Synthetic}} \\
\textbf{Method $\backslash$ Metric}  & PSNR$\uparrow$ & SSIM$\uparrow$ & LPIPS$\downarrow$ & Train$\downarrow$ & FPS$\uparrow$ \\
\hline
TensoRF~\cite{TensoRF} & 36.53 & 0.982 & 0.026 & - & - \\
3D-GS~\cite{3D_Gaussian_Splatting}* & 37.07 & 0.987 & 0.014 & 6min & 498 \\
Mip-Splatting~\cite{Yu2023MipSplatting}* & 37.28 & 0.988 & 0.014 & 6min & 445 \\
NeuRBF~\cite{chen2023neurbf} & 37.80 & 0.988 & 0.019 & - & $\simeq$1 \\
Spec-Gaussian~\cite{Spec-gaussian}* & \cellcolor{yellow!40}37.97 & \cellcolor{yellow!40}0.989 & \cellcolor{yellow!40}0.012 & 15min & 102 \\
\hdashline
RayGauss~\cite{raygauss}* & \cellcolor{orange!40}38.72 & \cellcolor{orange!40}0.990 & \cellcolor{orange!40}0.011 & 42min &  \textbf{22} \\
\hline
RayGaussX (ours) & \cellcolor{red!40}38.75 & \cellcolor{red!40}0.991 & \cellcolor{red!40}0.010 & 26min & \textbf{71} \\
\end{tabular}
}
\caption{Results on NSVF Synthetic dataset~\cite{NSVF}. Methods marked * were retrained on an NVIDIA RTX4090 for fair comparison of training and rendering (FPS) times.}
\label{tab:nsvf}
\end{table}

\subsection{Results on real-world datasets}
\begin{table*}[ht]
\centering
\resizebox{\linewidth}{!}{%
\begin{tabular}{l| c|c|c|c|c| c|c|c|c|c| c|c|c|c|c}
\textbf{Dataset} & \multicolumn{5}{c|}{\textbf{Mip-NeRF360}} & \multicolumn{5}{c|}{\textbf{Tanks\&Temples}} & \multicolumn{5}{c}{\textbf{Deep Blending}} \\
\textbf{Method $\backslash$ Metric} & PSNR$\uparrow$ & SSIM$\uparrow$ & LPIPS$\downarrow$ & Train$\downarrow$ & FPS$\uparrow$ &    PSNR$\uparrow$ & SSIM$\uparrow$ & LPIPS$\downarrow$ & Train$\downarrow$ & FPS$\uparrow$ &    PSNR$\uparrow$ & SSIM$\uparrow$ & LPIPS$\downarrow$ & Train$\downarrow$ & FPS$\uparrow$  \\
\hline
Instant-NGP~\cite{Instant_NGP} & 25.59 & 0.699 & 0.331 & - & - & 21.92 & 0.745 & 0.305 & - & - & 24.97 & 0.817 & 0.390 & - & - \\
3DGRT~\cite{3dgrt2024}* & 26.89 & 0.807 & 0.259 & 60min & 59 & 22.74 & 0.844 & 0.197 & 28min & 111 & 29.74 & 0.905 & 0.315 & 49min & 65 \\
EVER~\cite{EVER} & 27.51 & 0.825 & 0.194 & - & 36 & - & - & - & - & - & - & - & - & - & - \\
Mip-NeRF360~\cite{Mip_NeRF_360} & 27.69 & 0.792 & 0.237 & - & - & 22.22 & 0.759 & 0.257 & - & - & 29.40 & 0.901 & 0.245 & - & -  \\
3D-GS~\cite{3D_Gaussian_Splatting}* & 27.80 & 0.825 & 0.208 & 21min & 161 & \cellcolor{yellow!40}23.72 & 0.848 & 0.178 & 14min & 197 & 29.92 & 0.905 & 0.244 & 22min & 176  \\
Mip-Splatting~\cite{Yu2023MipSplatting}* & 27.93 & \cellcolor{orange!40}0.838 & \cellcolor{red!40}0.176 & 27min & 114 & \cellcolor{yellow!40}23.72 & \cellcolor{orange!40}0.860 & \cellcolor{orange!40}0.157 & 15min & 170 & 29.51 & 0.903 & \cellcolor{yellow!40}0.243 & 24min & 139\\
Spec-Gaussian~\cite{Spec-gaussian}* & 28.05 & \cellcolor{yellow!40}0.835 & \cellcolor{orange!40}0.177 & 41min & 42 & \cellcolor{red!40}23.86 & \cellcolor{yellow!40}0.856 & \cellcolor{yellow!40}0.166 & 21min & 85 & 29.67 & \cellcolor{yellow!40}0.906 & 0.245 & 26min & 75 \\
Zip-NeRF~\cite{zipnerf}* & \cellcolor{red!40}28.56 & 0.828 & \cellcolor{red!40}0.176 & 425min & 0.1 & 10.85 & 0.326 & 0.661 & 422min & 0.3 & \cellcolor{red!40}30.76 & \cellcolor{orange!40}0.914 & \cellcolor{red!40}0.209 & 416min & 0.2\\
\hdashline
RayGauss~\cite{raygauss}* & \cellcolor{yellow!40}28.23 & 0.834 & \cellcolor{yellow!40}0.193 & 456min & \textbf{0.5} & 23.20 & 0.849 & 0.167 & 565min & \textbf{0.5} & \cellcolor{yellow!40}30.30 & \cellcolor{orange!40}0.914 & \cellcolor{orange!40}0.242 & 553min & \textbf{0.8} \\
\hline
RayGaussX (ours) & \cellcolor{orange!40}28.43 & \cellcolor{red!40}0.842 & \cellcolor{orange!40}0.177 & 85min & \textbf{24.1} & \cellcolor{orange!40}23.76 & \cellcolor{red!40}0.865 & \cellcolor{red!40}0.150 & 47min & \textbf{41.1} & \cellcolor{orange!40}30.32 & \cellcolor{red!40}0.915 & \cellcolor{orange!40}0.242 & 56min & \textbf{39.6} \\
\end{tabular}
}
\caption{Results on Mip-NeRF360~\cite{Mip_NeRF_360}, Tanks\&Temples~\cite{Knapitsch2017} and Deep Blending~\cite{Hedman2018} datasets. FPS are calculated on the resolutions of the test set images. Methods marked * were retrained on an NVIDIA RTX4090 for fair comparison of training and rendering (FPS) times.}
\label{tab:realxp}
\end{table*}

Tab.~\ref{tab:realxp} reports results on three real-world datasets: Mip-NeRF360, Tanks\&Temples, and Deep Blending~\cite{Mip_NeRF_360,Knapitsch2017,Hedman2018}.

First, we observe that compared to RayGauss, RayGaussX achieves training times that are \textbf{5 to 12 times faster} and rendering speeds that are \textbf{50 to 80 times faster} in FPS, while also improving quality by up to \textbf{+0.56 dB} in PSNR on the Tanks\&Temples dataset. Also, RayGaussX is 100x to 200x faster in FPS than the state-of-the-art method Zip-NeRF~\cite{zipnerf}, while being only -0.13 dB lower in PSNR on Mip-NeRF360. Furthermore, RayGaussX outperforms Spec-Gaussian~\cite{Spec-gaussian}, the best Gaussian splatting variant, by +0.38 dB in PSNR on Mip-NeRF360 dataset, while being only twice as slow in rendering speed. Overall, RayGaussX achieves better rendering quality than Gaussian splatting-based methods (3D-GS~\cite{3D_Gaussian_Splatting}, Mip-Splatting~\cite{Yu2023MipSplatting}, and Spec-Gaussian~\cite{Spec-gaussian}) while approaching the rendering speed of the most recent variant, Spec-Gaussian~\cite{Spec-gaussian}. As shown in Tab.~\ref{tab:realxp}, the addition of features to improve the rendering quality of 3D-GS, such as antialiasing in Mip-Splatting~\cite{Yu2023MipSplatting} and the MLP for high-frequency color prediction in Spec-Gaussian~\cite{Spec-gaussian}, has progressively slowed down rasterization-based methods, decreasing from 161 FPS for 3D-GS to 42 FPS for Spec-Gaussian on Mip-NeRF360 dataset.

Secondly, we observe that the most notable improvement in rendering quality between RayGauss and RayGaussX occurs in outdoor scenes (from the Mip-NeRF360~\cite{Mip_NeRF_360} and Tanks\&Temples~\cite{Knapitsch2017} datasets). This improvement is due to the new densification strategy, which allows densifying Gaussians far from the camera and better reconstructing distant areas. 

Qualitative results on the Mip-NeRF360~\cite{Mip_NeRF_360} dataset are shown in Figure~\ref{fig:figure1} and in the supplementary material.

\subsection{Ablation study on the main contributions}
\label{sec:ablation}

Tab.~\ref{tab:ablation_study} presents an ablation study on the different contributions of RayGaussX that significantly accelerate rendering. This study was conducted on the \textit{garden} scene from the Mip-NeRF360~\cite{Mip_NeRF_360} dataset. We decomposed RayGaussX into four main contributions: the new densification (D), empty space skipping with adaptive sampling (E+A), reordering of Gaussians with ray coherence (Z+R), and finally, the isotropic loss (L). The first row (1) of Tab.~\ref{tab:ablation_study} represents the original RayGauss method, row (1*) is our optimized RayGauss baseline with only implementation-level optimizations and the last row (8) corresponds to our complete RayGaussX approach.

It can be observed that all four contributions improve the training time and rendering speed of RayGauss. The new densification alone (row (2)) significantly improves rendering quality (28.38 dB) while also slightly enhancing training and rendering speed. This is due to the novel densification strategy, which, compared to RayGauss, adds more distant Gaussians while reducing the number of nearby ones, thereby slightly speeding up the ray marching process. The most impactful contributions for accelerating ray marching with Gaussian primitives are empty space skipping with adaptive sampling (row (3), achieving a 3.9× speedup in FPS compared to (1*)) and the isotropic loss (row (5), yielding a 6.7× increase in FPS). Additionally, the isotropic loss alone (row (5)) results in a -0.2 dB drop in PSNR compared to (1*), but it provides significant gains in training and rendering speed. For this reason, the isotropic loss is preferred for real-world datasets but is not retained for synthetic datasets. Finally, as more contributions are added, rendering speed progressively improves, reaching \textbf{27 FPS} for interactive rendering, compared to \textbf{0.5 FPS} for RayGauss at a test image resolution of 1297×840 pixels.

\begin{table}[h]
\centering
\resizebox{\columnwidth}{!}{%
\begin{tabular}{l | c | c | c | c | c | c | c | c | c }
& D & E+A & Z+R & L & PSNR$\uparrow$ & SSIM$\uparrow$ & LPIPS$\downarrow$ & Train$\downarrow$ & FPS$\uparrow$  \\
\hline
(1) & & & & & 28.06 & 0.879 & 0.103 & 585min & 0.5  \\
(1*) & & & & & 28.14 & 0.885 & 0.090 & 297min & 1.5  \\
(2) & \checkmark  & & &  & 28.38 & 0.887 & 0.090 & 294min & 1.8 \\
(3) & & \checkmark & & & 28.14 & 0.884 & 0.090 & 180min & 5.9 \\
(4) & & & \checkmark & & 28.15 & 0.885 & 0.089 & 200min & 3.6\\
(5) & & & & \checkmark & 28.12 & 0.885 & 0.090 & 133min & 10.1 \\

(6) & \checkmark & \checkmark &  &  &                    28.35 & 0.887 & 0.089 & 168min & 6.3    \\
(7) & \checkmark & \checkmark & \checkmark & &           28.36 & 0.888 & 0.089 & 116min & 10.5 \\
(8) & \checkmark & \checkmark & \checkmark & \checkmark & 28.35 & 0.887 & 0.090 & 84min & 27.4 \\
\end{tabular}
}
\caption{Ablation study of key contributions on \textit{garden} scene from Mip-NeRF360 dataset (D: new Densification criterion, E+A: Empty space Skipping + Adaptive sampling, Z+R: Z-curve re-indexing of gaussians + Ray coherence, L: Loss for isotropic gaussians). Row (1) represents the original RayGauss method~\cite{raygauss}, row (1*) is our optimized RayGauss baseline with only implementation-level optimizations (excluding D, E, A, Z, R, L), and row (8) corresponds to our full method, RayGaussX.}
\label{tab:ablation_study}
\end{table}
\section{Conclusion}
\label{sec:conclusion}
Building on the RayGauss algorithm~\cite{raygauss}, which has achieved great graphical results on synthetic and indoor scenes, we propose a new approach for real-time novel view synthesis. Our approach introduces innovative contributions: empty-space skipping, adaptive sampling, reordering of rays and primitives for improved ray coherence, scale regularization, and a novel densification criterion. These enhancements enable the real-time synthesis of novel views with state-of-the-art graphical quality across indoor and outdoor scenes from multiple datasets.

\clearpage
\newpage

{
    \small
    \bibliographystyle{ieeenat_fullname}
    \bibliography{main}
}



\end{document}


\maketitle

\section*{Supplementary Overview}
This supplementary material is organized as follows. Section~\ref{sec:limitations_future} discusses current limitations and outlines future work. Section~\ref{sec:ratio_upper_bound} derives the upper bound on the volume ratio (see Sec.~3.4 of the main paper). We first outline a general derivation of the bound in Subsection~\ref{subsec:general_proof}, then compute the minimum AABB of an ellipsoid in Subsection~\ref{subsec:minimum_aabb}, and finally establish an upper bound on its volume in Subsection~\ref{subsec:upper_bound_aabb}. Section~\ref{sec:densif_connect} connects the densification criterion with the projected‐mean derivative. Section~\ref{sec:eval_densif} experimentally demonstrates that the new densification criterion more effectively densifies regions far from camera poses. Section~\ref{sec:detailed_ablations} presents a detailed ablation study evaluating the impact of each contribution. Section~\ref{sec:comparison_related_methods} presents a targeted comparison of our approach with two related methods: 3D Gaussian Splatting (3D-GS)~\cite{3D_Gaussian_Splatting} and 3D Gaussian Ray Tracing (3DGRT)~\cite{3dgrt2024}. First, we compare rendering quality using the same appearance model (spherical harmonics / spherical Gaussians) for both 3D-GS and our method to isolate the impact of their rendering pipelines. Second, we analyze the different constraints of our approach and 3DGRT that motivate our choice of enclosing volumes. Section~\ref{sec:additional_results} presents additional experimental results in two parts: Subsection~\ref{sec:qualitative_results} shows qualitative comparisons, and Subsection~\ref{sec:detailed_quantitative_results} provides detailed quantitative evaluations.

\section{Limitations and Future Work}
\label{sec:limitations_future}
The proposed approach achieves state-of-the-art rendering quality with a reasonable training time of 60–80 minutes and enables interactive rendering.  However, this requires a high-end GPU (NVIDIA RTX 4090 in our experiments), whereas Gaussian Splatting and its variants can run on mobile devices or in web-GL environments. Future work could further accelerate rendering with novel optimizations. Additionally, our approach does not properly handle aliasing, which falls outside the scope of this paper and could be addressed by future work. Nevertheless, RayGaussX’s fast training and high-quality rendering make it a strong framework for applications requiring high accuracy, including surface reconstruction~\cite{PGSR}, inverse rendering~\cite{GS-IR}, SLAM~\cite{GS-SLAM}, camera optimization~\cite{GS-CPR}, and relighting~\cite{GS3}.

\section{Computation of the upper bound on the ratio of volumes}
\label{sec:ratio_upper_bound}
In this section, we derive the upper bound used for the volume ratio in \textbf{Section~3.4 Limiting Highly Anisotropic Gaussian} of the main paper. First, in Subsection~\ref{subsec:general_proof}, we present a general derivation of this bound by assuming intermediate results, which are then proven in Subsections~\ref{subsec:minimum_aabb} and~\ref{subsec:upper_bound_aabb}.

\subsection{Derivation of the Volume Ratio Bound}
\label{subsec:general_proof}
We consider here the computation of an upper bound for the ratio:
\begin{equation}
    r=\frac{\text{Vol}(\text{AABB}(\mathcal{E}))}{\text{Vol}(\mathcal{E})}
\end{equation}
with \(\text{Vol}\) the volume function, \(\mathcal{E}\) a given ellipsoid, and \(\text{AABB}(\mathcal{E})\) its associated axis-aligned bounding box.

In particular, if we consider the $l$-th primitive of our representation, the ellipsoid associated with the $l$-th Gaussian is the $\sigma_{\epsilon}$-level isosurface of the density function of the $l$-th primitive:

\begin{equation}
    \sigma_l(\mathbf{x}) = \tilde{\sigma_l} \cdot \exp \left( -\frac{1}{2} (\mathbf{x} - \mathbf{\mu_l})^T \mathbf{\Sigma_l}^{-1} (\mathbf{x} - \mathbf{\mu_l}) \right)
\end{equation}
where $\sigma_{\epsilon}$ is the chosen density threshold \cite{raygauss}. In order to avoid overloading the notations, we omit the index $l$ hereafter. The covariance matrix can be decomposed as: $\mathbf{\Sigma} = \mathbf{R} \mathbf{S} \mathbf{S}^T  \mathbf{R}^T$ where \(\mathbf{R}=(r_{i,j})_{1\leq i,j \leq 3}\) is a rotation matrix and \( \mathbf{S} = \text{diag}(s_{1}, s_{2}, s_{3}) \) a scale matrix. Hence, the ellipsoid associated with the primitive can be described by the equation:
\begin{equation}
(\mathbf{x} - \mathbf{\mu})^T {\tilde{\mathbf{\Sigma}}}^{-1} (\mathbf{x} - \mathbf{\mu})\leq 1
\label{eq:ellips_global}
\end{equation}
such that $\tilde{\mathbf{\Sigma}} = \mathbf{R} \,\tilde{\mathbf{S}}\, \tilde{\mathbf{S}}^T  \mathbf{R}^T$, where $\tilde{\mathbf{S}} = \sqrt{2\,\ln\!\Bigl(\frac{\tilde\sigma}{\sigma_{\epsilon}}\Bigr)} \cdot \mathbf{S}$.
The volume of ellipsoid $\mathcal{E}$ is then:
\[
\text{Vol}(\mathcal{E})
=
\frac{4\pi}{3}
\,
\Bigl[2\ln\!\Bigl(\tfrac{\tilde{\sigma}}{\sigma_{\epsilon}}\Bigr)\Bigr]^{3/2}
\,
s_{1}\,s_{2}\,s_{3}
\]
Also, it can be shown that the volume of the associated AABB, $\text{Vol}(\text{AABB}(\mathcal{E}))$, is equal to (see Subsection~\ref{subsec:minimum_aabb} for details):

\begin{equation}
8\,\Bigl(2\,\ln\!\Bigl(\tfrac{\tilde\sigma}{\sigma_\epsilon}\Bigr)\Bigr)^{\frac{3}{2}}
\sqrt{
\prod_{i=1}^{3}
\Bigl(
r_{i,1}^2\,s_1^2
+
r_{i,2}^2 s_2^2
+
r_{i,3}^2s_3^2
\Bigr)
}
\end{equation}
And therefore, the ratio of the volumes is:
\begin{equation}
r
=
\frac{6}{\pi}
\,
\frac{
\sqrt{
\prod_{i=1}^3
\bigl(r_{i,1}^2\,s_1^2 + r_{i,2}^2\,s_2^2 + r_{i,3}^2\,s_3^2\bigr)
}
}{
s_1\,s_2\,s_3
}
\end{equation}
independent of the scale factor introduced by $\sigma_{\epsilon}$ and $\tilde{\sigma}$. Subsequently, we ensure that this ratio remains rotation-invariant to avoid bias toward any particular Gaussian orientation, and we constrain the worst-case scenario by computing an upper bound (see Subsection~\ref{subsec:upper_bound_aabb} for details):

\begin{equation}
\frac{\text{Vol}(\text{AABB}(\mathcal{E}))}{\text{Vol}(\mathcal{E})}
\;\le\;
\frac{2}{\pi\,\sqrt{3}}
\;\frac{\bigl(s_{1}^{2} + s_{2}^{2} + s_{3}^{2}\bigr)^{\tfrac{3}{2}}}{s_{1}\,s_{2}\,s_{3}}
\end{equation}

Thus, we obtain the expression for the upper bound used in the main article in the section \textbf{Limiting Highly Anisotropic Gaussian}.

\subsection{Computation of the Minimum Axis-Aligned Bounding Box for an Ellipsoid}
\label{subsec:minimum_aabb}
Considering the ellipsoid $\mathcal{E}$ defined in global coordinates by
\begin{equation}
(x - \mu)^T \tilde{\Sigma}^{-1} (x - \mu) \leq 1    
\end{equation}
with: $\tilde{\Sigma} = R\, \tilde{S}\, \tilde{S}^T R^T$  and $\tilde{S} = \operatorname{diag}(\tilde{s}_1, \tilde{s}_2, \tilde{s}_3),
$
where the rotation matrix \( R = (r_{ij}) \) performs the transformation of the ellipsoid (aligned in its local coordinate system) into the global coordinate system.

In this section, we determine the minimum axis-aligned bounding box for the ellipsoid \(\mathcal{E}\). To obtain the axis-aligned bounding box in the world reference frame, we need to determine, for each axis \( i \) in the world frame (with \( i = 1,2,3 \) corresponding to \( x, y, \) and \( z \)), the maximum half-length \( L_i \) such that

\begin{equation}
L_i = \max_{x \in \mathbb{R}^3} \left\{ \left| x_i - \mu_i \right| \, : \, (x - \mu)^T \tilde{\Sigma}^{-1} (x - \mu) \leq 1 \right\}    
\end{equation}

The volume of the axis-aligned bounding box (AABB) will then be computed as

\begin{equation}
\text{Vol}(\text{AABB}(\mathcal{E})) = \prod_{i=1}^{3} 2L_i    
\end{equation}

First, we introduce the local variable in the ellipsoid-centered reference frame: $y = R^T (x - \mu)$ and  thus we obtain: $(x - \mu) = R y$ thanks to the orthogonality of \( R \). Similarly, the half-lengths can be re-expressed using the local coordinate $y$ in the ellipsoid's reference frame, using the fact that:
\begin{equation}
\left| x_i - \mu_i \right| = \left|\left[ x - \mu \right]_i \right| =  \left |\sum_{j=1}^{3} r_{ij} y_j \right|
\end{equation}
and:
\begin{equation}
(x - \mu)^T \tilde{\Sigma}^{-1} (x - \mu) = y^T (\tilde{S} \tilde{S}^T)^{-1} y \leq 1    
\end{equation}
Since \( \tilde{S} \) is diagonal, this constraint can be rewritten as:

\[
\sum_{j=1}^{3} \left( \frac{y_j}{\tilde{s}_j} \right)^2 \leq 1
\]
Thus, we obtain the new expression for the half-lengths:
\begin{equation}
L_i = \max_{y \in \mathbb{R}^3} \left\{ \left |\sum_{j=1}^{3} r_{ij} y_j \right| \; : \; \sum_{j=1}^{3} \left( \frac{y_j}{\tilde{s}_j} \right)^2 \leq 1 \right\}.
\end{equation}

We now introduce the variable \( z_j = \frac{y_j}{\tilde{s}_j} \) and reformulate the problem in terms of the variable \( z \), which belongs to the unit ball in \( \mathbb{R}^3 \), as follows:

\begin{equation}
L_i = \max_{z \in \mathbb{R}^3} \left\{ \left |\sum_{j=1}^{3} r_{ij} \tilde{s}_j z_j \right| \; : \; \sum_{j=1}^{3} \left(z_j\right)^2 \leq 1 \right\}.
\end{equation}
Here, using the Cauchy-Schwarz inequality, we obtain:
\begin{equation}
    L_i = \sqrt{\sum_{j=1}^{3} r_{ij}^2 \tilde{s}_j^2}
\end{equation}
where the maximum is attained for the vector collinear to \( \mathbf{r}_{i,\cdot} \tilde{\mathbf{s}} \) with unit norm.
Consequently, the volume of the axis-aligned bounding box can be computed as:
\begin{equation}
\text{Vol}(\text{AABB}(\mathcal{E})) = 8 \sqrt{\prod_{i=1}^{3}\sum_{j=1}^{3} r_{ij}^2 \tilde{s}_j^2}    
\end{equation}
Thus, we derive the formula used to compute the expression of \( r \) in Section~\ref{sec:ratio_upper_bound}.

\subsection{Upper Bound on the Volume of the Enclosing Axis-Aligned Bounding Box}
\label{subsec:upper_bound_aabb}
We express the volume of the AABB enclosing the ellipsoid in the form:
\begin{equation}
\text{Vol}(\text{AABB}(\mathcal{E})) = 8 \sqrt{\prod_{i=1}^{3}\sum_{j=1}^{3} r_{ij}^2 s_j^2}    
\end{equation}
We use \( s \) instead of \( \tilde{s} \) because, at the stage of the section~\ref{sec:ratio_upper_bound} requiring this upper bound, the scaling accounting for \( \sigma \) and \( \sigma_{\epsilon} \) is no longer necessary. 

Here, we apply the arithmetic-geometric inequality to \( l_i = \sum_{j=1}^{3} r_{ij}^2 s_j^2 \), yielding:
\begin{equation}
l_1 l_2 l_3 \leq \left( \frac{l_1 + l_2 + l_3}{3} \right)^3
\label{eq:GM_AM}
\end{equation}
with equality when \( l_1 = l_2 = l_3 \).  Then, we have:
\begin{equation}
\sum_{i=1}^{3}l_i=\sum_{i=1}^3\sum_{j=1}^{3} r_{ij}^2 s_j^2
\end{equation}
We can note here that, since the matrix \( R \) is orthogonal, its columns form an orthonormal basis, thus \( \sum_{i=1}^{3} r_{ij}^2 = 1 \), yielding:

\begin{equation}
\sum_{i=1}^{3}l_i=\sum_{j=1}^{3} s_j^2
\end{equation}
and:
\begin{equation}
l_1 l_2 l_3 \leq \left( \frac{\sum_{j=1}^{3} s_j^2}{3} \right)^3
\label{eq:GM_AM_last}
\end{equation}
Thus, knowing that \( \text{Vol}(\text{AABB}(\mathcal{E})) = 8\sqrt{l_1 l_2 l_3} \), and by using the equations \eqref{eq:GM_AM_last}, we obtain:
\begin{equation}
    \text{Vol}(\text{AABB}(\mathcal{E}))\leq \frac{8}{3\sqrt{3}} \left( \sum_{j=1}^{3} s_j^2 \right)^{\frac{3}{2}}
\end{equation}
This expression enables us to establish an upper bound on the ratio \( r \) in Section~\ref{sec:ratio_upper_bound}. Moreover, this upper bound is achieved when the half-lengths are equal $L_1=L_2=L_3$ (since $L_i=\sqrt{l_i}$).

\begin{figure*}[ht]
  \centering
  \begin{tabular}{@{}c@{\hspace{0.5cm}}c@{\hspace{0.5cm}}c@{\hspace{0.5cm}}c@{}}
    \begin{tikzpicture}[baseline]
      \node[anchor=south west, inner sep=0] (img) 
        {\includegraphics[height=3cm]{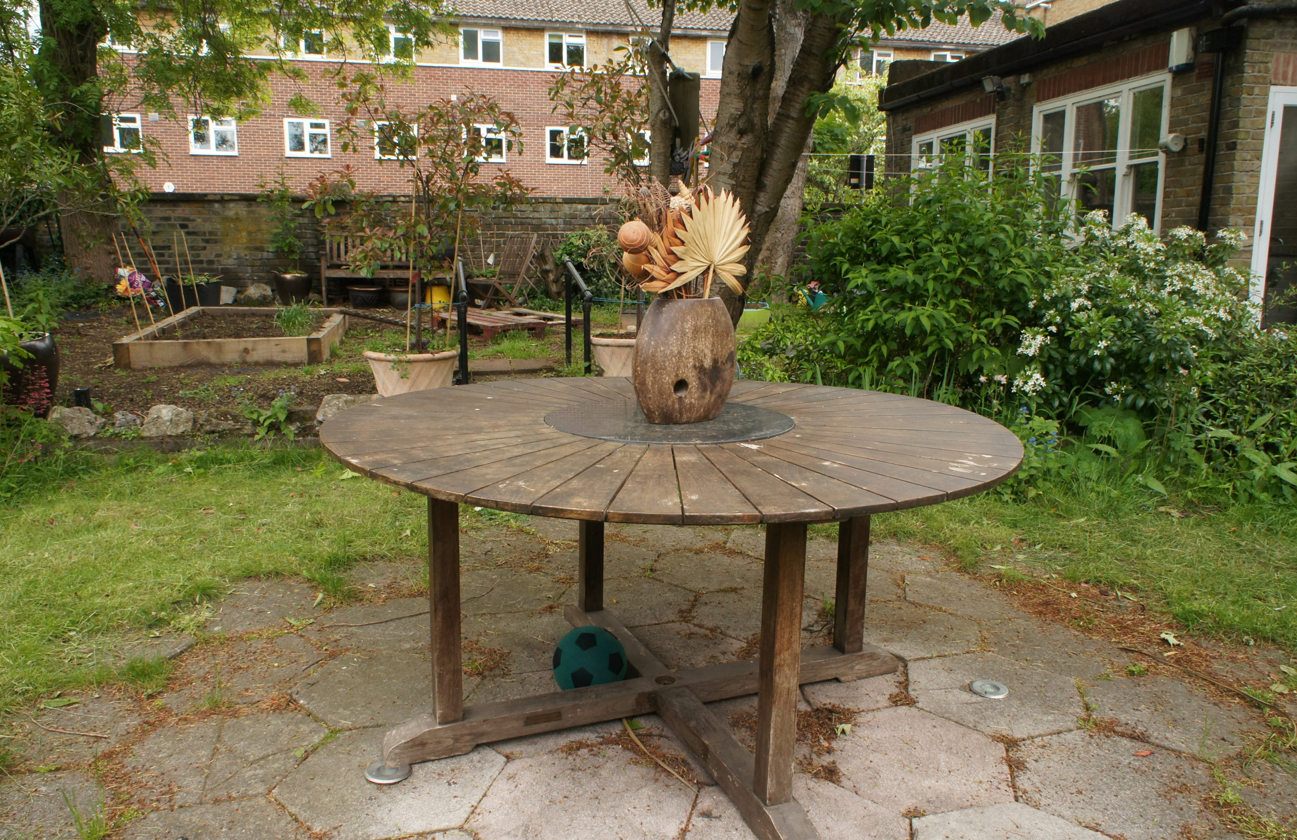}};
      \begin{scope}[x={(img.south east)}, y={(img.north west)}]
        \draw[red,ultra thick]
          ({78/1297},{(840-13)/840}) rectangle ({(78+280)/1297},{(840-13-280)/840});
      \end{scope}
    \end{tikzpicture}
    &
    \includegraphics[height=3cm]{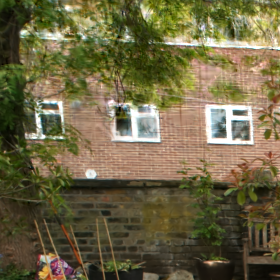}
    &
    \includegraphics[height=3cm]{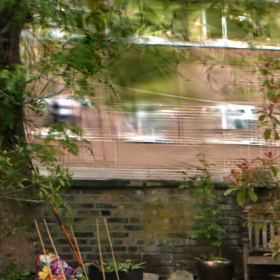}
    &
    \includegraphics[height=3cm]{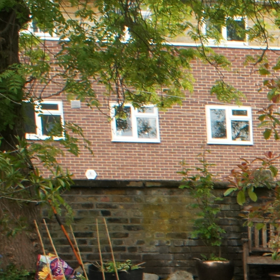}
    \\[0.5ex]
     & New densification & Old densification & Ground‑truth
  \end{tabular}
  \caption{Comparison of background reconstruction for image 1 in the \textit{garden} test scene. From left to right, the full scene with the region of interest highlighted in red, the reconstruction obtained with the new densification criterion, the reconstruction obtained with the previous densification criterion, and the corresponding ground‑truth image crop.}
  \label{fig:background_comparison}
\end{figure*}

\section{Connecting the Densification Criterion and the Projected Mean Derivative}
\label{sec:densif_connect}
In this section, \( o \) denotes the center of the camera, \( f \) its focal distance, and \( \mu \) the position of a Gaussian. We will see here that the densification criterion can be expressed in terms of the gradient with respect to the position \( \mu \) projected onto the sphere centered at \( o \) with radius \( f \), under well-chosen assumptions.
We define the projection \( P : \mathbb{R}^3 \setminus \{o\} \to \mathbb{R}^3 \) by:
\begin{equation}
    P(\mu) = o + f \,\frac{\mu - o}{\|\mu - o\|}
\end{equation}
where \( f > 0 \) is the camera's focal distance and \( o \) is the camera center. Thus, \( P(\mu) \) is always a point on the sphere of radius \( f \) centered at \( o \). Here, we introduce a new parameterization based on the position on the sphere. We define $\tilde{\mu} = (\mu_P, r)$ where:
\begin{itemize}
    \item \( \mu_P = P(\mu) = o + f \,\dfrac{\mu - o}{\|\mu - o\|} \) is the projection of \( \mu \in \mathbb{R}^3 \setminus \{o\} \) onto the sphere \( S^2(o, f) \) (the sphere of radius \( f>0 \) centered at \( o \)).
    \item \( r = \|\mu - o\| \) is the radial distance of \( \mu \) from \( o \).
\end{itemize}
We then define the reprojection function $F: S^2(o, f) \times \mathbb{R}_+ \to \mathbb{R}^3$ such that:
\begin{equation}
\mu = F(\mu_P, r) = o + \frac{r}{f} \bigl( \mu_P - o \bigr)
\end{equation}
Also, by introducing the unit vector $u = \frac{\mu - o}{\|\mu - o\|}$, we note that the derivatives of \( F \) with respect to the parameters can be expressed as follows:
\begin{equation}
\frac{\partial F}{\partial \mu_P} = \frac{r}{f}\, (I-uu^T)
\end{equation}
where \( I - u u^T \) is the projector onto \( T_{\mu_{P}} S^2(o, f) =\{ v \in \mathbb{R}^3 \mid v \cdot u = 0 \} \), the tangent space to the sphere centered at \( o \) and with radius \( f \) at the point \( \mu_{P} = o + f u \). This multiplication by the projection is necessary because \(\mu_P\) is constrained to lie on the sphere \( S^2(o, f) \); hence, its derivatives must lie in the tangent space \( T_{\mu_P} S^2(o, f) \).
Furthermore, we have:
\begin{equation}
\frac{\partial F}{\partial r} = \frac{1}{f}\,\left(\mu_P - o\right)
\end{equation}

Thus, we obtain the gradients, with respect to \(\mu_P\):
\begin{equation}
\nabla_{\mu_P} L = \frac{r}{f}(I - u u^T) \nabla_{\mu} L
\end{equation}
and the radial component:
\begin{equation}
\nabla_{r} L = \frac{1}{f}\, (\mu_P - o)^T \, \nabla_{\mu} L
\end{equation}
Experimentally, we observe that the derivative along the radial direction is significantly smaller compared to the tangential components. Also, we assume in the following analysis that \( \nabla_{\mu} L \cdot u = 0 \). Thus, we have:
\begin{equation}
\nabla_{\mu} L \in T_{\mu_P} S^2(o, f)
\end{equation}
Moreover, since \( (I - u u^T) \) is a projector onto \( T_{\mu_P} S^2(o, f) \) and \( \nabla_{\mu} L \) lies in this tangent space, we have:

\begin{equation}
(I - u u^T) \nabla_{\mu} L = \nabla_{\mu} L.
\end{equation}
and:
\begin{equation}
 (\mu_P - o)^T \, \nabla_{\mu} L = 0
\end{equation}
Thus, we finally obtain that:
\begin{equation}
\nabla_{\mu_P} L = \frac{r}{f} \nabla_{\mu} L
\end{equation}
and:
\begin{equation}
\nabla_{r} L = 0
\end{equation}
Therefore, the gradient with respect to the projection on the sphere is, in this case, proportional to the gradient in the 3D space, and by taking the norm, we obtain:
\begin{equation}
    \left\| \nabla_{\mu_P} L \right\|= \frac{r}{f}\left\| \nabla_{\mu} L \right\|
\end{equation}

\section{Experimental Evaluation of Densification}
\label{sec:eval_densif}
\begin{figure}[H]
  \centering
  \begin{subfigure}[b]{0.48\columnwidth}
    \centering
    \includegraphics[clip,width=\linewidth]{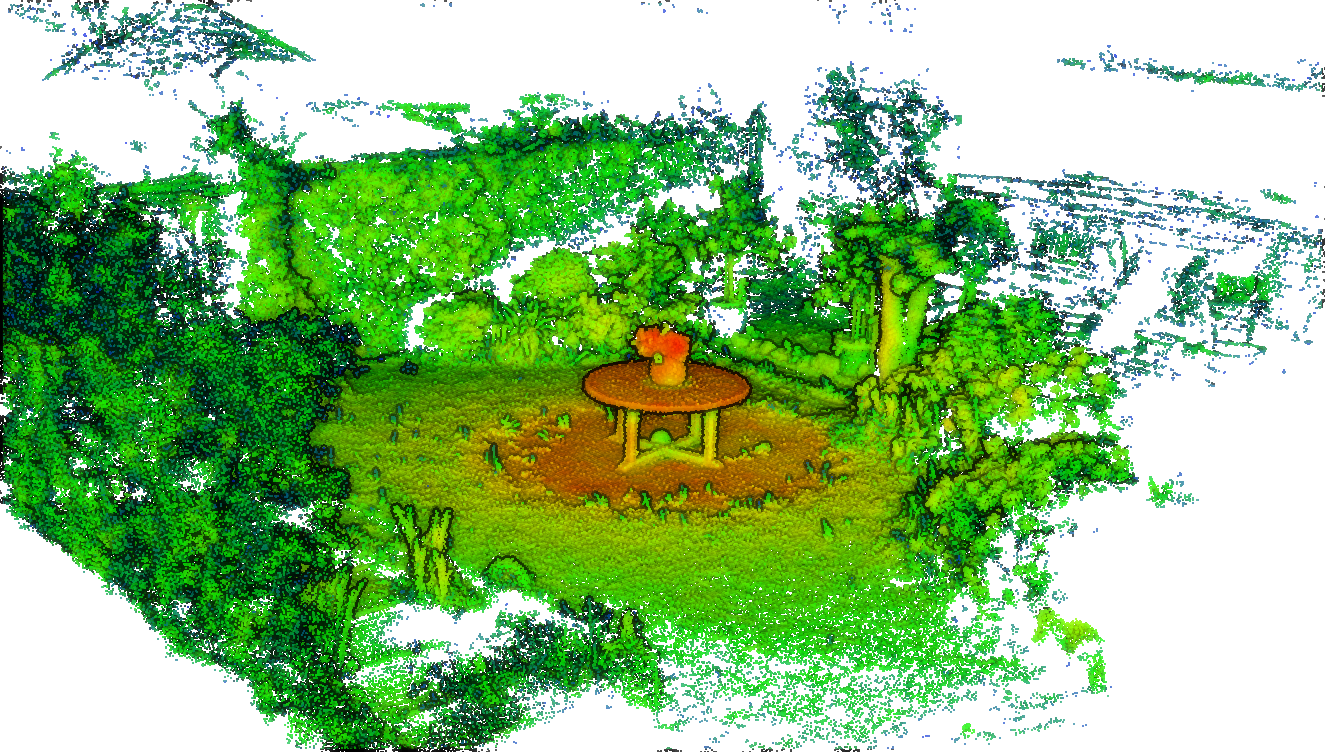}
    \caption{Raygauss densification}
  \end{subfigure}
  \hfill
  \begin{subfigure}[b]{0.48\columnwidth}
    \centering
    \includegraphics[clip,width=\linewidth]{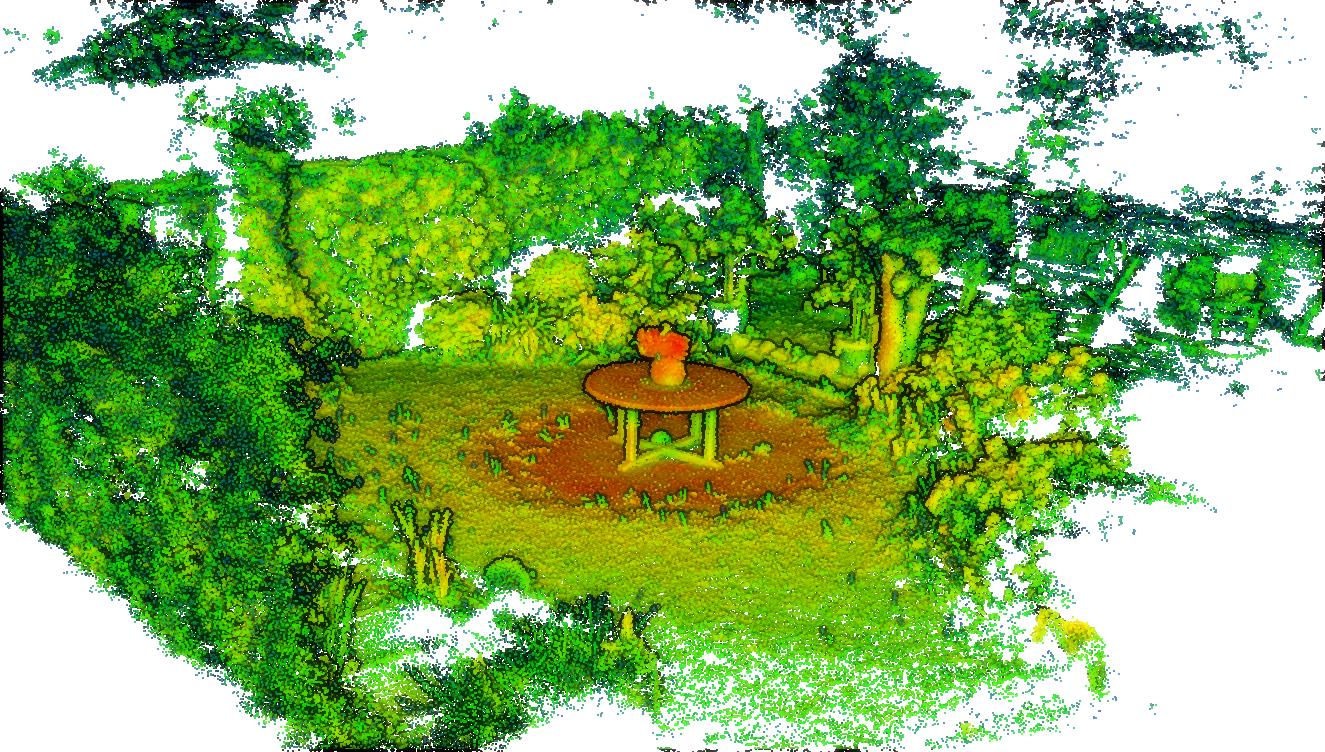}
    \caption{RaygaussX densification}
  \end{subfigure}
  \caption{Densification comparison in the MipNeRF360 \textit{garden} scene using Raygauss and RaygaussX criteria. The color gradient from blue through green, yellow, and red encodes on a logarithmic scale the increasing number of Gaussian neighbors within a radius of \(R = 0.125\).}
  \label{fig:densification_pc}
\end{figure}

In this section, we experimentally demonstrate the effectiveness of the new densification criterion. To this end, we train the \textit{garden} scene from the MipNeRF360 dataset first using the original criterion and then using the new densification criterion, with densification parameters tuned to yield approximately the same number of Gaussians in the scene (around 4 million). Subsequently, by considering the point clouds associated with the centers of the Gaussians, we compute the number of neighbors of each Gaussian within a radius of \(R = 0.125\) and use this metric to denote the density of Gaussians around a given Gaussian. The point clouds associated with this scalar field are shown in Fig.~\ref{fig:densification_pc}. In this scene, the cameras are predominantly positioned around the central table. Under the new densification criterion, there remains a significant number of Gaussians around the table, which corresponds to the area with the most information, and there are also more points distributed in the distant regions. In particular, the right-hand side of the \textit{garden} scene contains a higher density of Gaussians, confirming the effectiveness of the new criterion in better densifying distant areas. 

Moreover, we demonstrate the effect of the new densification on rendering in Fig.~\ref{fig:background_comparison} by comparing the background reconstruction of image~1 in the \textit{garden} scene test set. As shown, the previous densification produces a poorly reconstructed background that appears blurry and lacks detail. In contrast, our new approach yields a much more accurate background reconstruction, with fine details such as the reflection on the glass pane clearly visible.

\section{Detailed ablation study of each contribution}
\label{sec:detailed_ablations}
In this section, we present a detailed ablation study of the various contributions introduced in RayGaussX to complement the study presented in the main paper by detailing the influence of each component individually. Specifically, we investigate the influence of the main contributions: the novel densification strategy (D) which yields a more uniform distribution of Gaussians in regions distant from the camera poses, empty‐space skipping (E), adaptive sampling (A), spatial reordering of Gaussians via a Z‐order curve (Z), ray coherence optimization (R) and the isotropic loss (L).

  \begin{table*}[t]
  \centering
  \resizebox{\textwidth}{!}{%
    \begin{tabular}{l | c ||c c|| c c || c | c c c c c}
      & D & E & A & Z & R & L 
      & PSNR$\uparrow$ & SSIM$\uparrow$ & LPIPS$\downarrow$ 
      & Train$\downarrow$ & FPS$\uparrow$ \\
      \hline
      (1) &      &      &      &      &      &      
          & 28.06 & 0.879 & 0.103 & 585\,min & 0.5  \\
      (1*) &      &      &      &      &      &      
          & 28.14 & 0.885 & 0.090 & 297\,min & 1.5  \\
      (2) &      & \checkmark &  &      &      &      
          & 28.14 & 0.884 & 0.090 & 218\,min & 4.4  \\
      (3) &      &  & \checkmark &      &      &      
          & 28.13 & 0.885 & 0.089 & 191\,min & 4.7  \\
      (4) &      & \checkmark & \checkmark &      &      &      
          & 28.14 & 0.884 & 0.090 & 180\,min & 5.9  \\
      (5) &      &      &      & \checkmark &  &      
          & 28.14 & 0.885 & 0.089 & 237\,min & 2.4  \\
      (6) &      &      &      &  & \checkmark &      
          & 28.12 & 0.884 & 0.090 & 238\,min & 2.4  \\          
      (7) &      &      &      & \checkmark & \checkmark &      
          & 28.15 & 0.885 & 0.089 & 200\,min & 3.6  \\
      (8) &      & \checkmark & \checkmark & \checkmark & \checkmark &  
          & 28.15 & 0.885 & 0.089 & 125\,min & 10.2  \\
      (9) &      &      &      &      &      & \checkmark 
          & 28.12 & 0.885 & 0.090 & 133\,min & 10.1  \\
      (10) &      & \checkmark & \checkmark & \checkmark & \checkmark & \checkmark 
          & 28.16 & 0.885 & 0.090 &  88\,min & 26.1 \\
      (11) & \checkmark &      &      &      &      &      
          & 28.38 & 0.887 & 0.090 & 294\,min & 1.8  \\
      (12) & \checkmark & \checkmark & \checkmark &      &      &      
          & 28.35 & 0.887 & 0.089 & 168\,min & 6.3  \\
      (13) & \checkmark & \checkmark & \checkmark 
             & \checkmark & \checkmark &      
          & 28.36 & 0.888 & 0.089 & 116\,min & 10.5 \\
      (14) & \checkmark & \checkmark & \checkmark 
             & \checkmark & \checkmark & \checkmark 
          & 28.35 & 0.887 & 0.090 &  84\,min & 27.4 \\
    \end{tabular}%
  }
  \caption{Ablation study of key contributions on the \textit{garden} scene from the Mip-NeRF360 dataset (D: new Densification criterion; E: Empty-space Skipping; A: Adaptive sampling; Z: Z-curve re-indexing of Gaussians; R: Ray coherence; L: Loss for isotropic Gaussians). Row (1) represents the original RayGauss method~\cite{raygauss}, Row (1*) corresponds to the optimized RayGauss baseline underlying RayGaussX, incorporating implementation-level optimizations but excluding the main contributions (D, E, A, Z, R, L) and row (14) corresponds to our full method, RayGaussX.}
  \label{tab:ablation_study}
\end{table*}

This ablation study, presented in Table~\ref{tab:ablation_study}, is designed to determine how each individual contribution and their combinations influence rendering quality (PSNR, SSIM, LPIPS), training time, and inference speed in frames per second (FPS). First, rows (1) and (1*) demonstrate that implementation-level optimizations provide a stronger baseline than RayGauss in terms of training time and FPS, while preserving comparable rendering performance. These improvements are not described in the paper as they consist of using faster operations and pertain to implementation details. However, interested readers may compare our code, available online, with the implementation from the RayGauss paper. 

Next, rows (2), (3), (5), (6), and (9) confirm that the core contributions of the paper effectively reduce both training and inference times. We also observe that overall rendering quality is minimally affected by these contributions, demonstrating that they accelerate rendering without any significant trade-off in quality. The isotropic loss appears to slightly reduce PSNR, but this effect is minor.

Rows (4), (7), (8), and (10) illustrate the cumulative effect of the individual contributions. We observe no accumulation of errors that could degrade the rendering‐quality metrics; indeed, the PSNR remains approximately 28.14~dB. This confirms that, with the chosen hyperparameters, the contributions do not compromise rendering quality. Furthermore, training and inference are accelerated when combining these contributions, achieving a training time of 88 minutes and an inference speed of 26.1 FPS when all speed‐related contributions are applied.

Finally, rows (11), (12), (13), and (14) highlight the influence of the new densification on the algorithm’s performance. It is clear that the new densification yields superior rendering quality by better densifying distant regions, as shown in Section~\ref{sec:eval_densif}. In particular, with the new densification we achieve a PSNR of approximately 28.36 dB compared to 28.14 dB obtained with (1*). Furthermore, by comparing rows (12), (13), and (14) to their counterparts without the new densification, we observe that the new densification also appears to slightly accelerate rendering and inference times.

\section{Selective comparison with related methods}
\label{sec:comparison_related_methods}
In this section, we present a targeted comparison of our approach with two related methods: 3D Gaussian Splatting (3D-GS)~\cite{3D_Gaussian_Splatting} and 3D Gaussian Ray Tracing (3DGRT)~\cite{3dgrt2024}. First, we compare rendering quality using the same appearance model, spherical harmonics/spherical Gaussians, for both 3D-GS and our method to isolate the impact of their rendering pipelines. Second, we analyze the different constraints of our approach and 3DGRT that motivate our choice of enclosing volumes.

\begin{table*}[h]
  \centering
  \resizebox{\textwidth}{!}{%
    \begin{tabular}{lccccc}
      \toprule
      Method           & NeRF-Synth.  & NSVF-Synth.  & Mip-NeRF360   & Tanks \& Temp. & Deep Blending \\
      \midrule
      3D-GS            & 33.39/0.968  & 37.07/0.987  & 27.80/0.825   & 23.72/0.848    & 29.92/0.905   \\
      3D-GS (SH, SG)   & 33.88/0.971  & 38.06/0.988  & 28.12/0.827   & \textbf{23.81}/0.855 & 29.88/0.908 \\
      RayGaussX (ours) & \textbf{34.54/0.974} & \textbf{38.75/0.991} & \textbf{28.43/0.842} & 23.76/\textbf{0.865} & \textbf{30.32/0.915} \\
      \bottomrule
    \end{tabular}%
  }
  \caption{Comparison in terms of PSNR/SSIM  of RayGaussX (ours) and 3D-GS methods (with and without spherical harmonics/gaussians) on NeRF-Synthetic, NSVF-Synthetic, Mip-NeRF360, Tanks and Temples, and Deep Blending datasets. Best results in bold.}
  \label{tab:psnr_ssim_comparison}
\end{table*}

\subsection{Rendering quality comparison with a similar appearance model}
\label{sec:rendering_comparison}
In this subsection, we compare RayGaussX with 3D Gaussian Splatting (3D-GS) under the same appearance model (spherical harmonics and spherical Gaussians). Using an identical appearance representation isolates the rendering algorithm’s impact on output quality, since both methods employ Gaussian primitives within an identical appearance framework. We evaluate both methods on the datasets introduced in the main paper: NeRF-Synthetic, NSVF-Synthetic, Mip-NeRF360, Tanks\&Temples, and Deep Blending. The appearance model comprises nine spherical harmonics and seven spherical Gaussians per primitive. The resulting PSNR and SSIM values are reported in Fig.~\ref{tab:psnr_ssim_comparison}. Results show that the SH/SG formulation also enhances 3D-GS’s performance. However, except for PSNR on Tanks\&Temples, across all datasets and for every standard metric, RayGaussX outperforms the modified 3D-GS in terms of rendering quality, confirming the advantage of employing Volume Ray Marching. 
This aligns with theory, as ray marching provides a less approximate formulation that avoids rendering artifacts such as flickering compared to 3D-GS rasterization. Additionally, as noted by the authors of~\cite{Mip_NeRF_360}, the Tanks \& Temples~\cite{Knapitsch2017} dataset exhibits significant lighting variations, making conclusions drawn from this dataset somewhat uncertain.

\subsection{Enclosing volume selection}
The recent 3DGRT method~\cite{3dgrt2024} also uses the OptiX API to ray trace Gaussian primitives. However, their choice of enclosing volume differs from ours: they use bounding polyhedra and we use axis aligned bounding boxes(AABB). We explain the reasons for this difference in this subsection. 

The 3D Gaussian Ray Tracing approach performs ray tracing of Gaussians with a single sample per Gaussian; this simplified approach approximates the rendering equation and, unlike RayGauss, does not account for the overlap of multiple Gaussians. Additionally, compared with AABBs, 3DGRT’s complex bounding primitives raise two issues. First, constructing them and building the associated BVH are several orders of magnitude slower (Fig. 9 in 3DGRT paper), so the frequent BVH rebuilds required during training become a bottleneck in scenes with many primitives. A second drawback stems from using polyhedra with the OptiX API for ray marching: indeed, we integrate each ray segment sequentially. However, if a segment lies entirely within a bounding polyhedron, OptiX reports no intersection, wrongly treating it as empty because it crosses no faces. In contrast, with AABBs, an OptiX-specific behavior is that it reports an intersection even for fully enclosed segments. For these reasons, AABBs are better suited to our Volume Ray Marching algorithm than the bounding polyhedra proposed by 3DGRT.
\section{Additional experimental results}
\label{sec:additional_results}

\subsection{Qualitative results}
\label{sec:qualitative_results}

Fig.~\ref{fig:qualtitative_mipnerf360} presents qualitative results of our RayGaussX approach compared to RayGauss~\cite{raygauss}, Zip-NeRF~\cite{zipnerf}, and 3D-GS~\cite{3D_Gaussian_Splatting} on the Mip-NeRF360~\cite{Mip_NeRF_360} dataset. First, we observe that the qualitative results are very similar between RayGaussX and RayGauss for indoor scenes. However, for outdoor scenes, our approach shows a clear improvement in background rendering thanks to our new densification criterion. The Zip-NeRF method achieves very good visual results but produces grainy images (particularly noticeable in the \textit{counter} scene) and sometimes generates floaters, as seen on the wooden foot in the \textit{treehill} scene. Finally, 3D-GS exhibits lower reconstruction quality, with less accurately reconstructed objects, which is especially visible in the shelf of the \textit{room} scene.

As noted by the authors of~\cite{Mip_NeRF_360}, the Tanks\&Temples~\cite{Knapitsch2017} dataset contains images with significant variations in lighting conditions, making conclusions drawn from this dataset somewhat uncertain. Moreover, we can observe that the Zip-NeRF~\cite{zipnerf} method failed on this dataset.

\newlength{\imagewidth}
\settowidth{\imagewidth}{\includegraphics{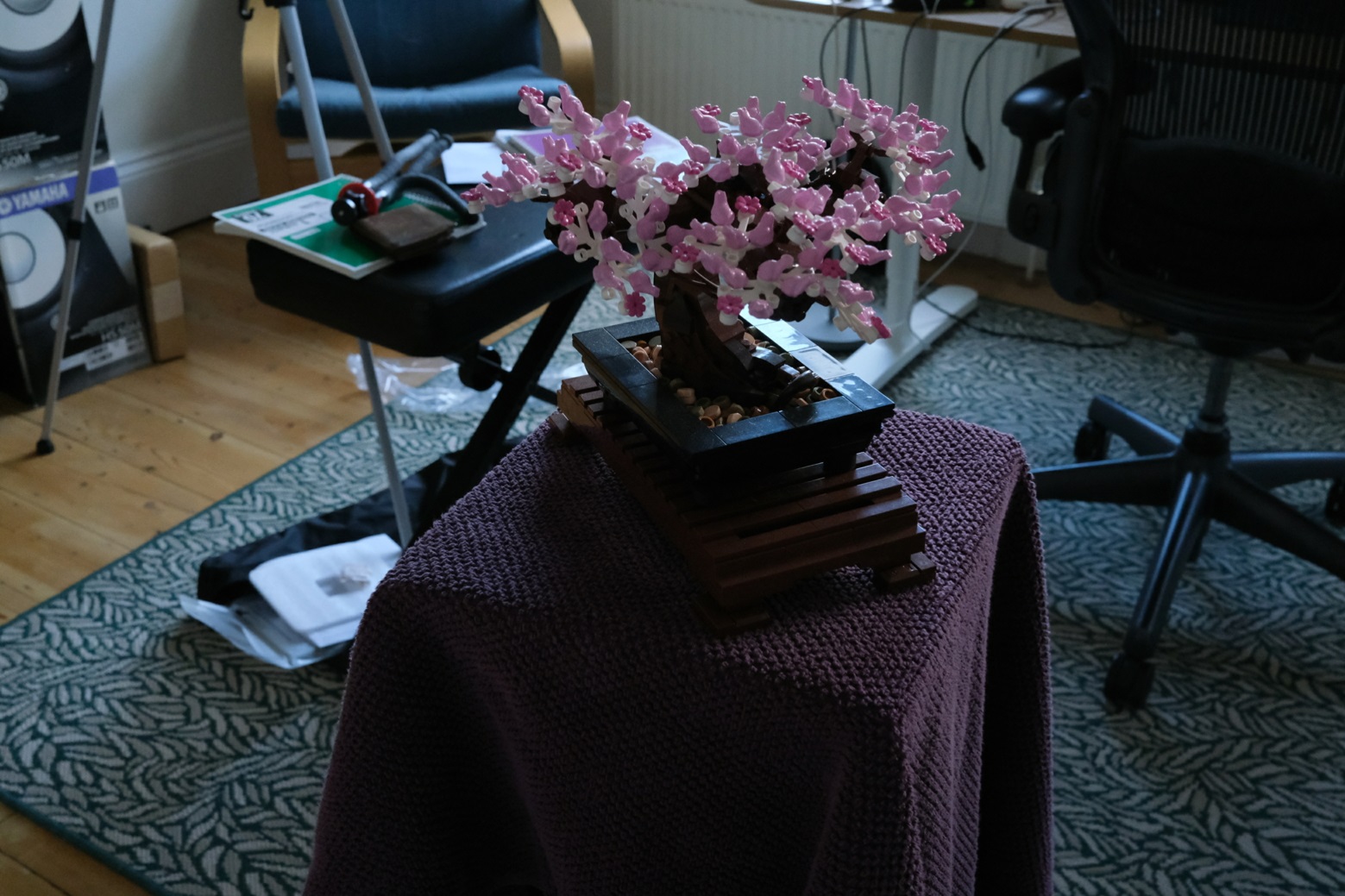}}
\newlength{\imageheight}
\settoheight{\imageheight}{\includegraphics{figures/figures_supp/mipnerf360/bonsai_gt_16.jpg}}

\begin{figure*}[t!]
\centering
\begin{subfigure}[t]{.19\textwidth}
    \centering
    \includegraphics[trim=0.5\imagewidth{} 0.5\imageheight{} 0.2\imagewidth{} 0.1\imageheight{},clip, height=2.9cm]{figures/figures_supp/mipnerf360/bonsai_gt_16.jpg}
\end{subfigure}
\begin{subfigure}[t]{.19\textwidth}
    \centering
    \includegraphics[trim=0.5\imagewidth{} 0.5\imageheight{} 0.2\imagewidth{} 0.1\imageheight{},clip, height=2.9cm]{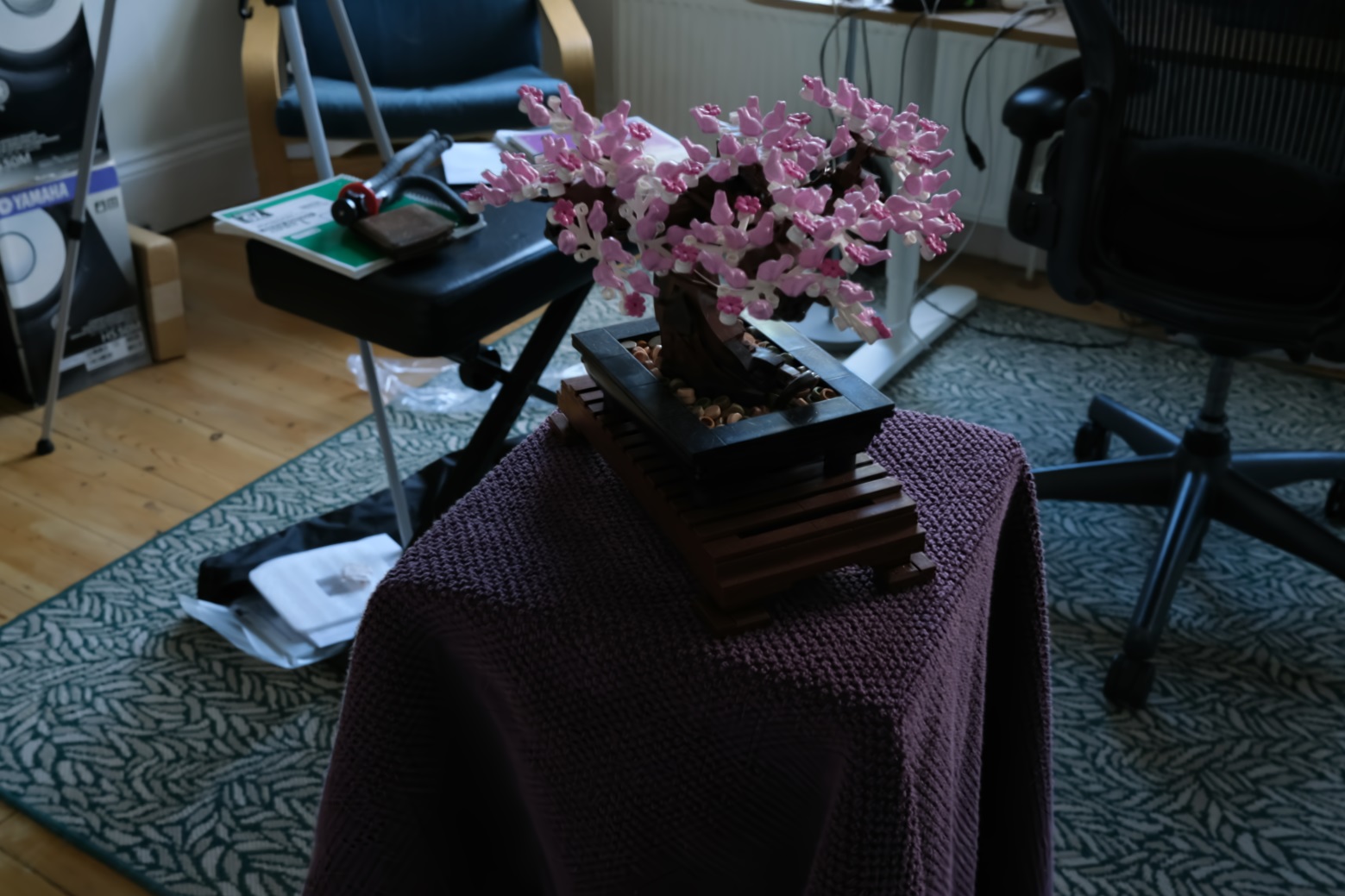}
\end{subfigure}
\begin{subfigure}[t]{.19\textwidth}
    \centering
    \includegraphics[trim=0.5\imagewidth{} 0.5\imageheight{} 0.2\imagewidth{} 0.1\imageheight{},clip, height=2.9cm]{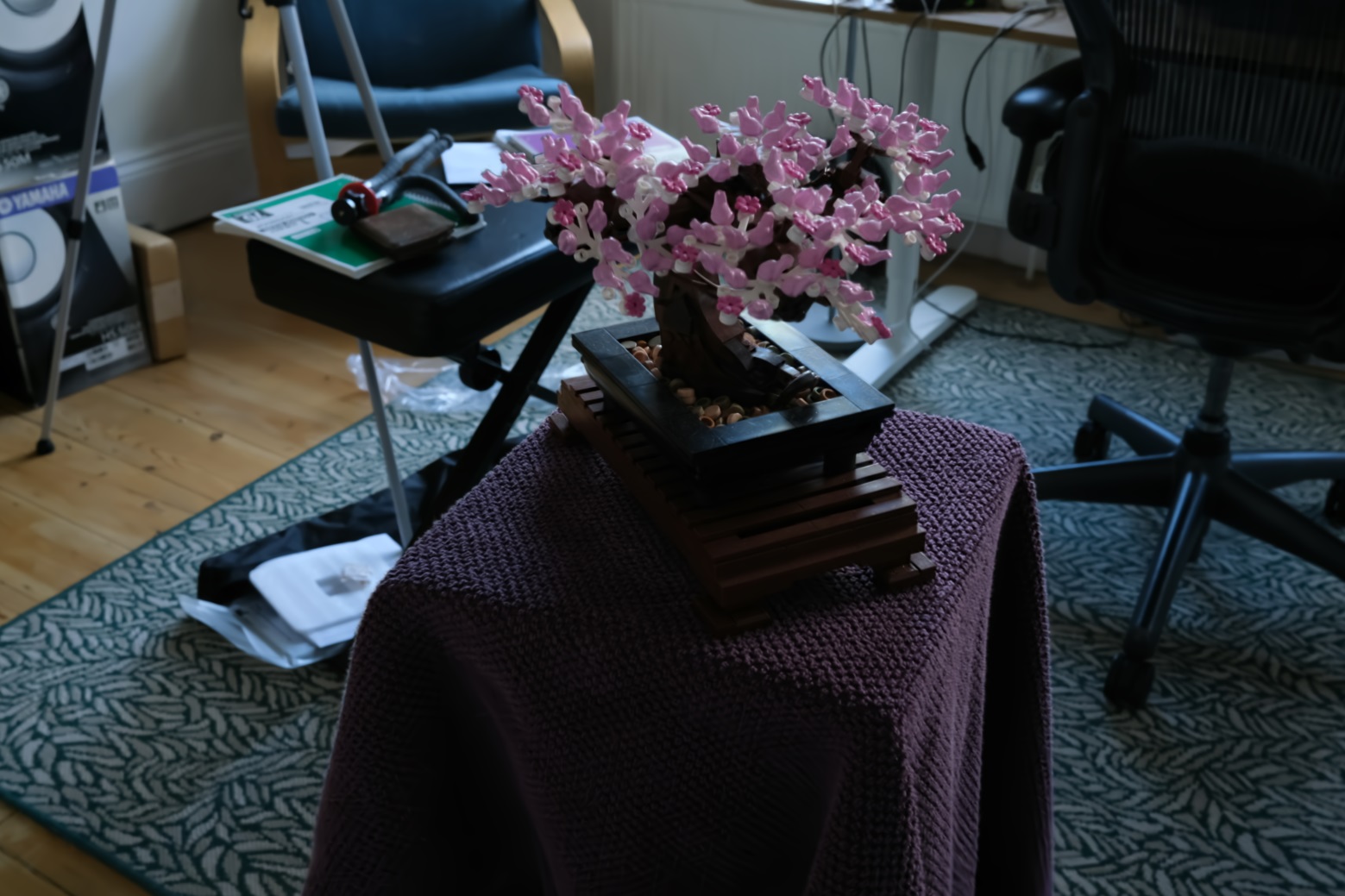}
\end{subfigure}
\begin{subfigure}[t]{.19\textwidth}
    \centering
    \includegraphics[trim=0.5\imagewidth{} 0.5\imageheight{} 0.2\imagewidth{} 0.1\imageheight{},clip, height=2.9cm]{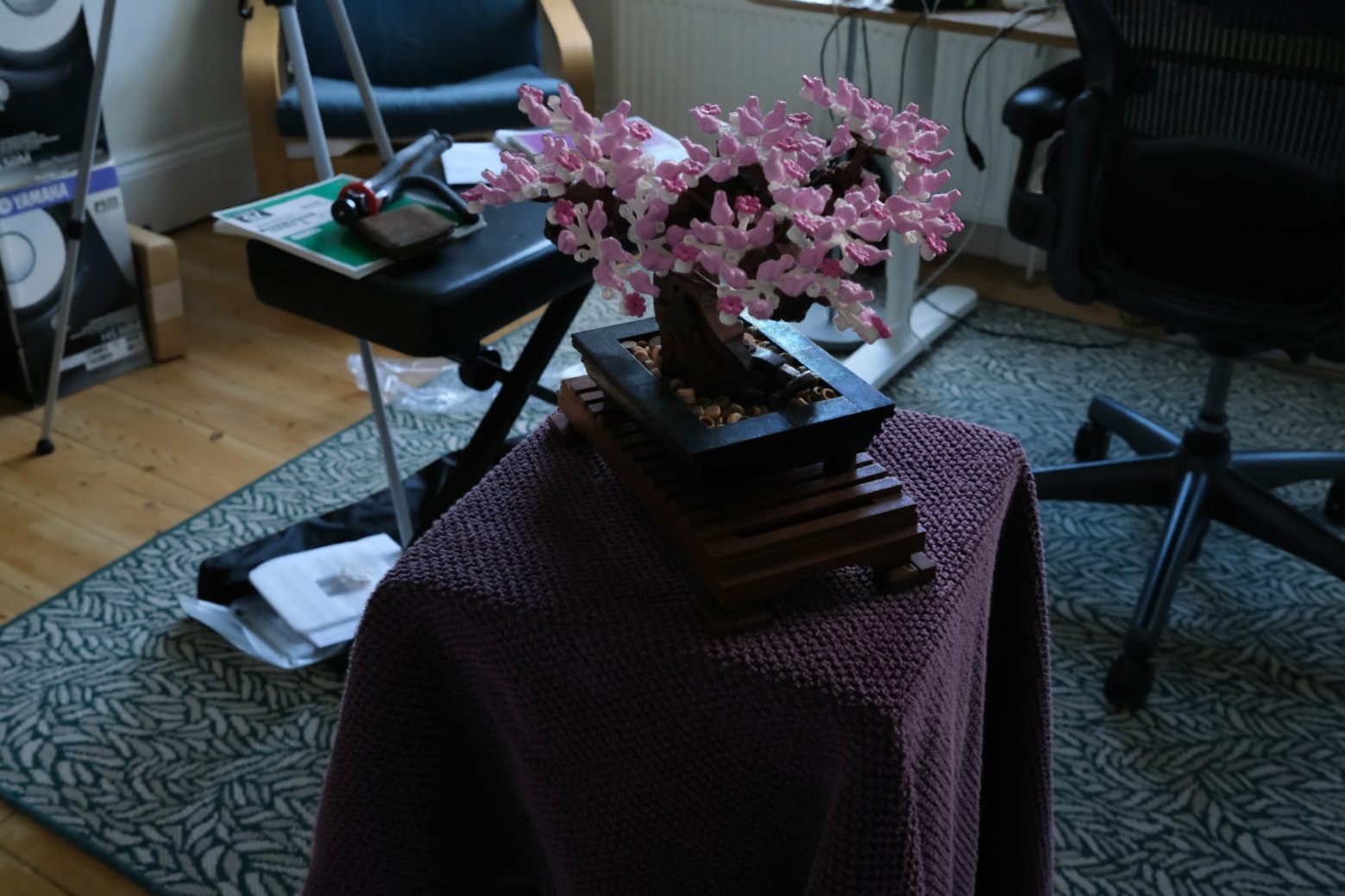}
\end{subfigure}
\begin{subfigure}[t]{.19\textwidth}
    \centering
    \includegraphics[trim=0.5\imagewidth{} 0.5\imageheight{} 0.2\imagewidth{} 0.1\imageheight{},clip, height=2.9cm]{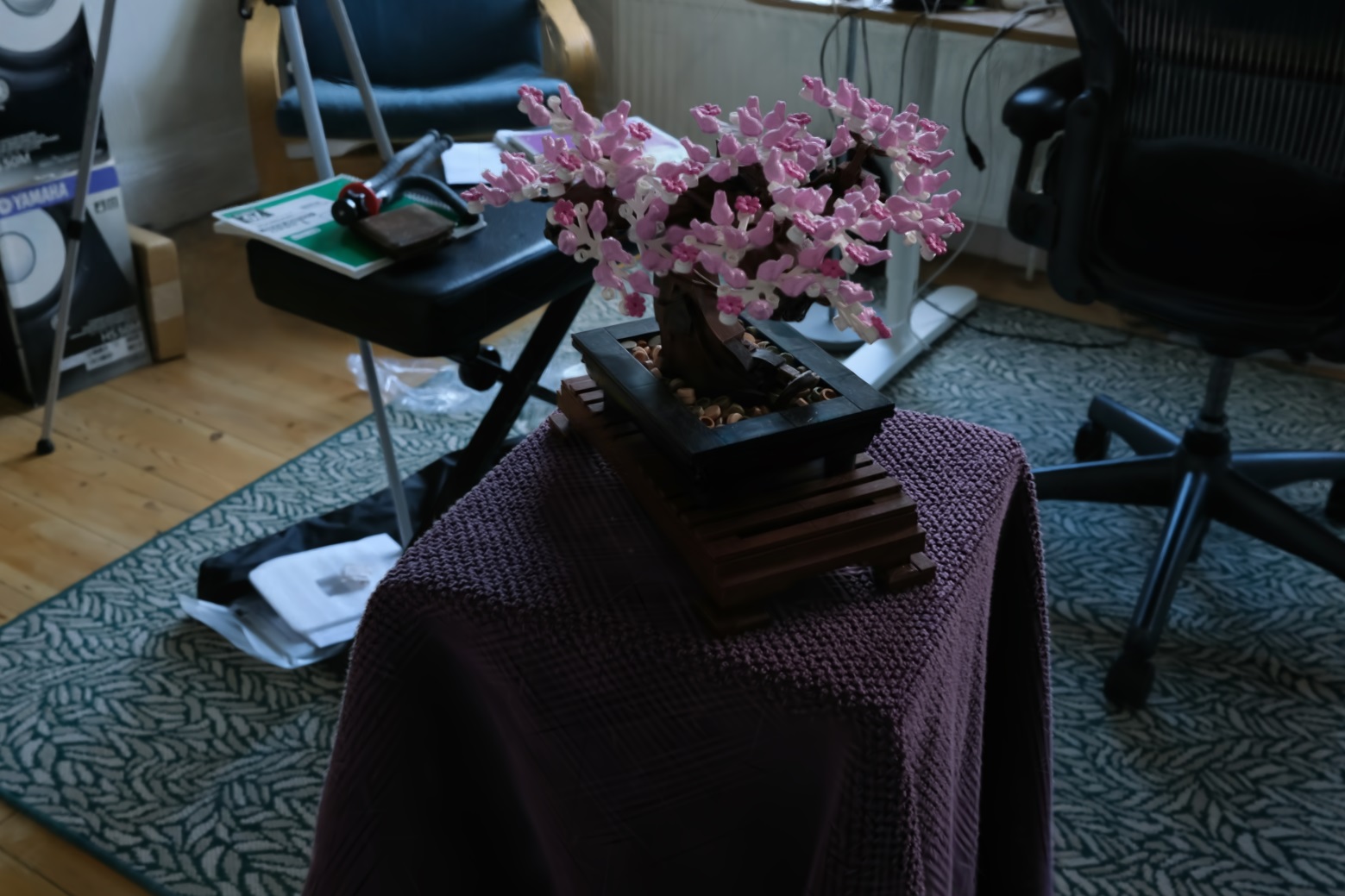}
\end{subfigure}

\vspace{0.2cm}

\begin{subfigure}[t]{.19\textwidth}
    \centering
    \includegraphics[trim=0.5\imagewidth{} 0.45\imageheight{} 0.3\imagewidth{} 0.28\imageheight{},clip, height=2.9cm]{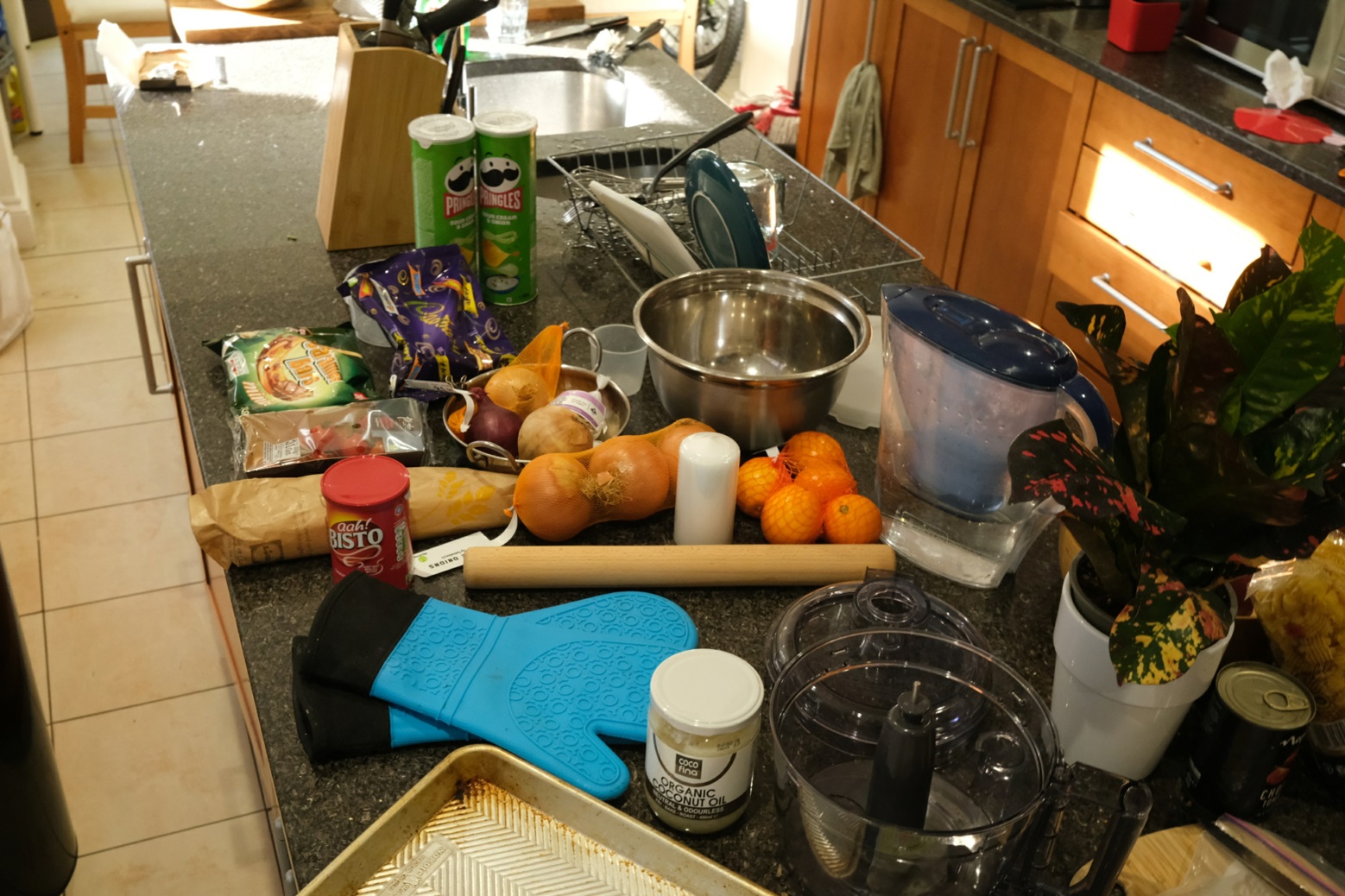}
\end{subfigure}
\begin{subfigure}[t]{.19\textwidth}
    \centering
    \includegraphics[trim=0.5\imagewidth{} 0.45\imageheight{} 0.3\imagewidth{} 0.28\imageheight{},clip, height=2.9cm]{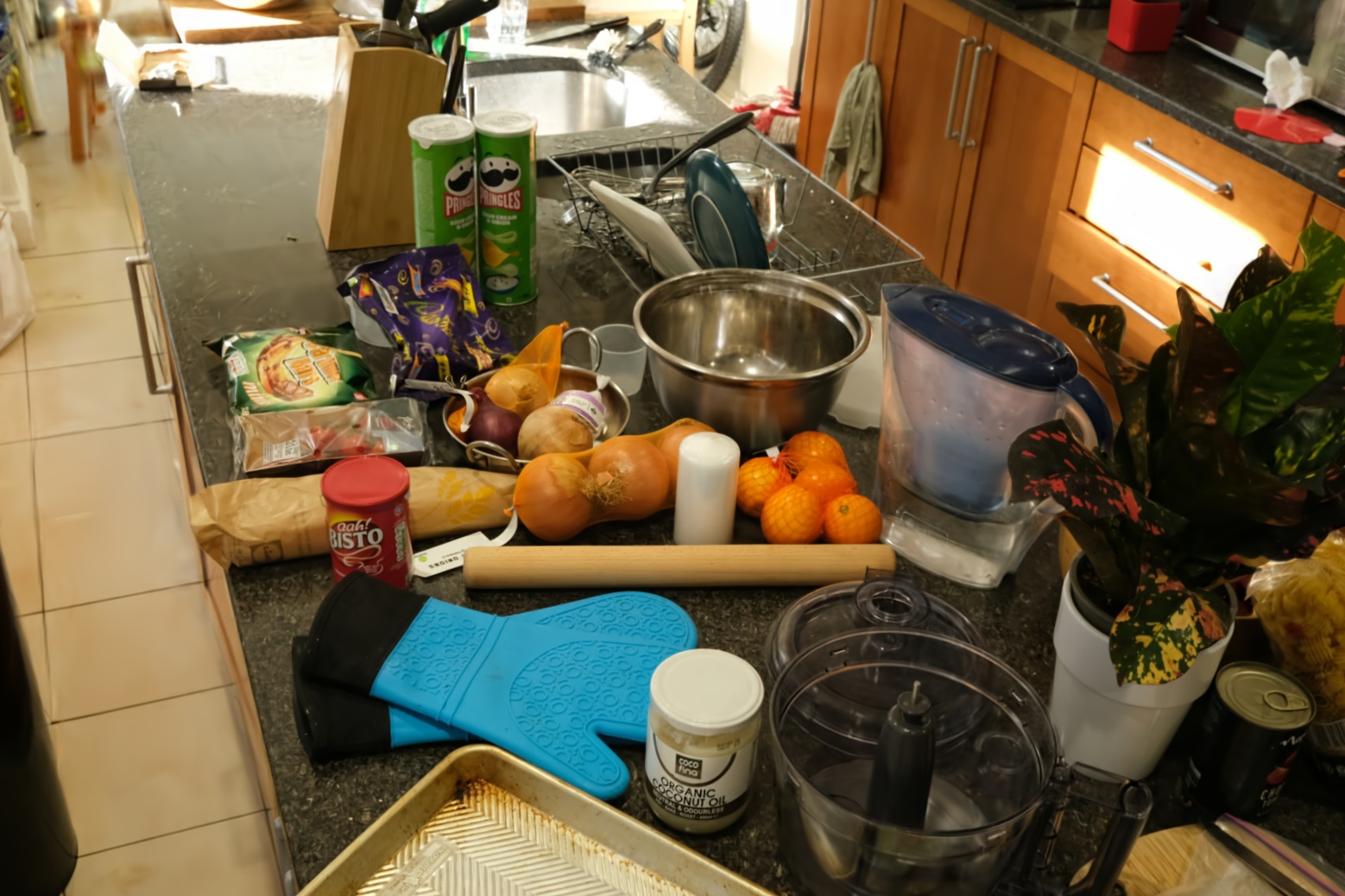}
\end{subfigure}
\begin{subfigure}[t]{.19\textwidth}
    \centering
    \includegraphics[trim=0.5\imagewidth{} 0.45\imageheight{} 0.3\imagewidth{} 0.28\imageheight{},clip, height=2.9cm]{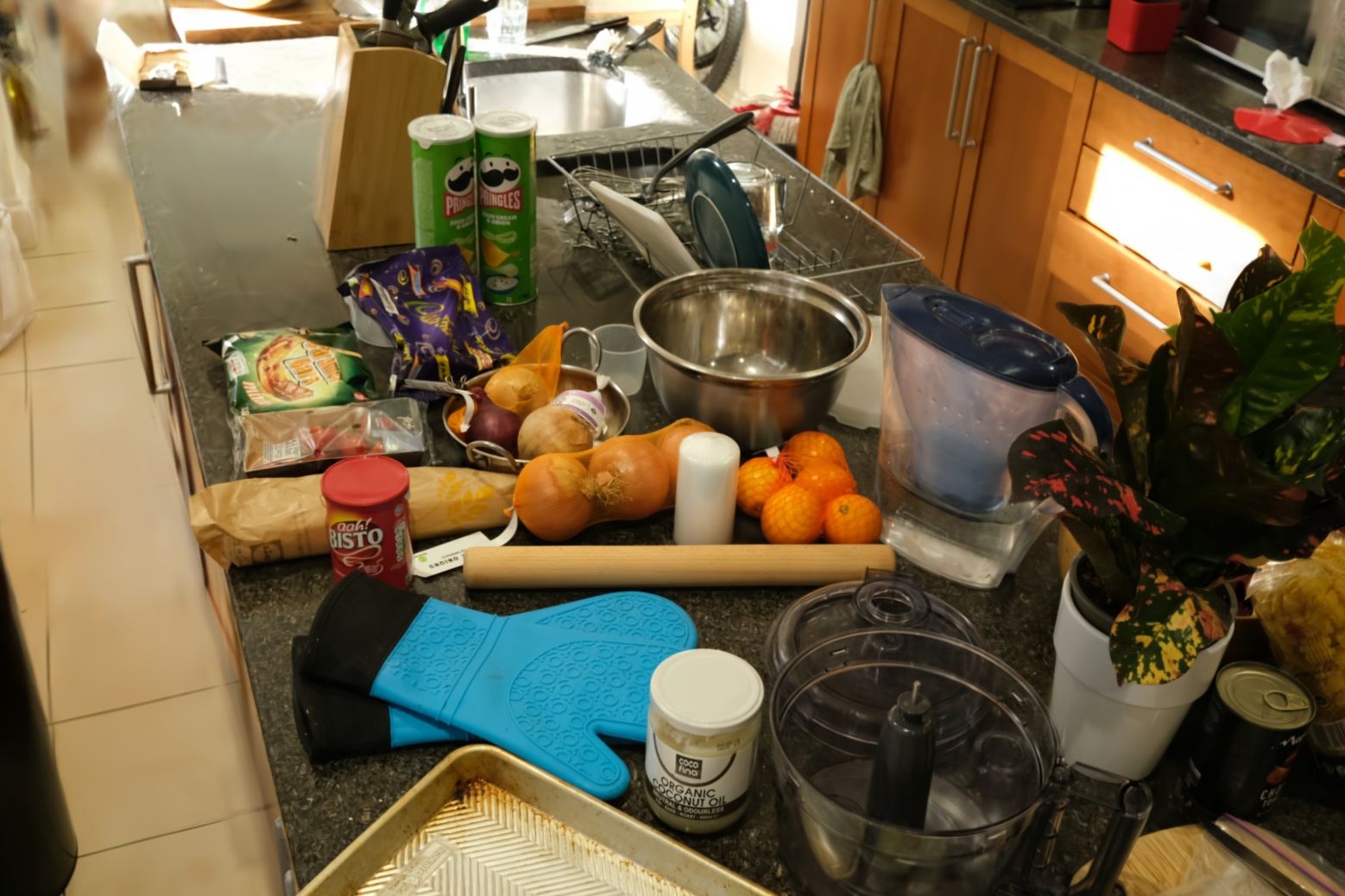}
\end{subfigure}
\begin{subfigure}[t]{.19\textwidth}
    \centering
    \includegraphics[trim=0.5\imagewidth{} 0.45\imageheight{} 0.3\imagewidth{} 0.28\imageheight{},clip, height=2.9cm]{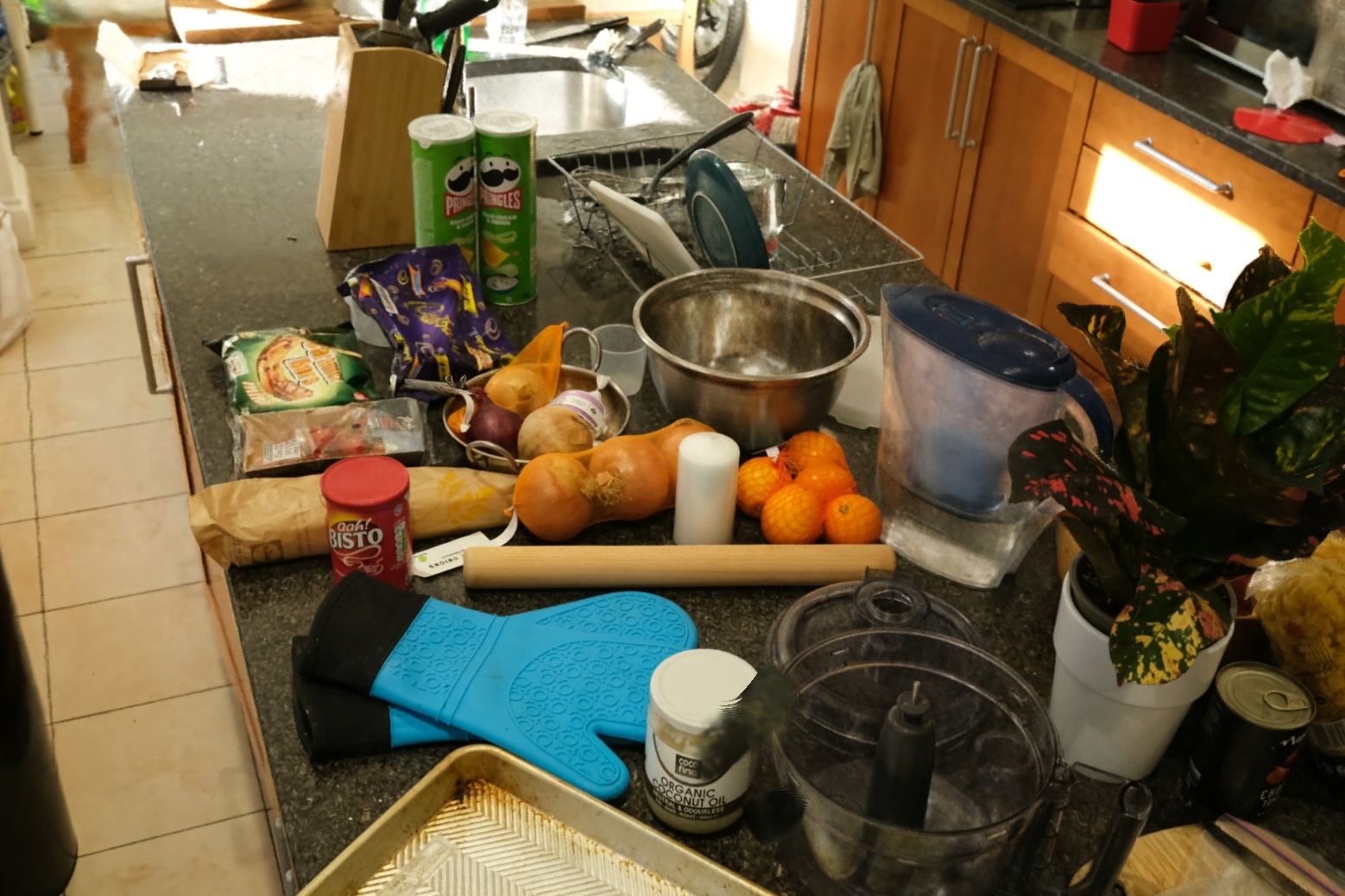}
\end{subfigure}
\begin{subfigure}[t]{.19\textwidth}
    \centering
    \includegraphics[trim=0.5\imagewidth{} 0.45\imageheight{} 0.3\imagewidth{} 0.28\imageheight{},clip, height=2.9cm]{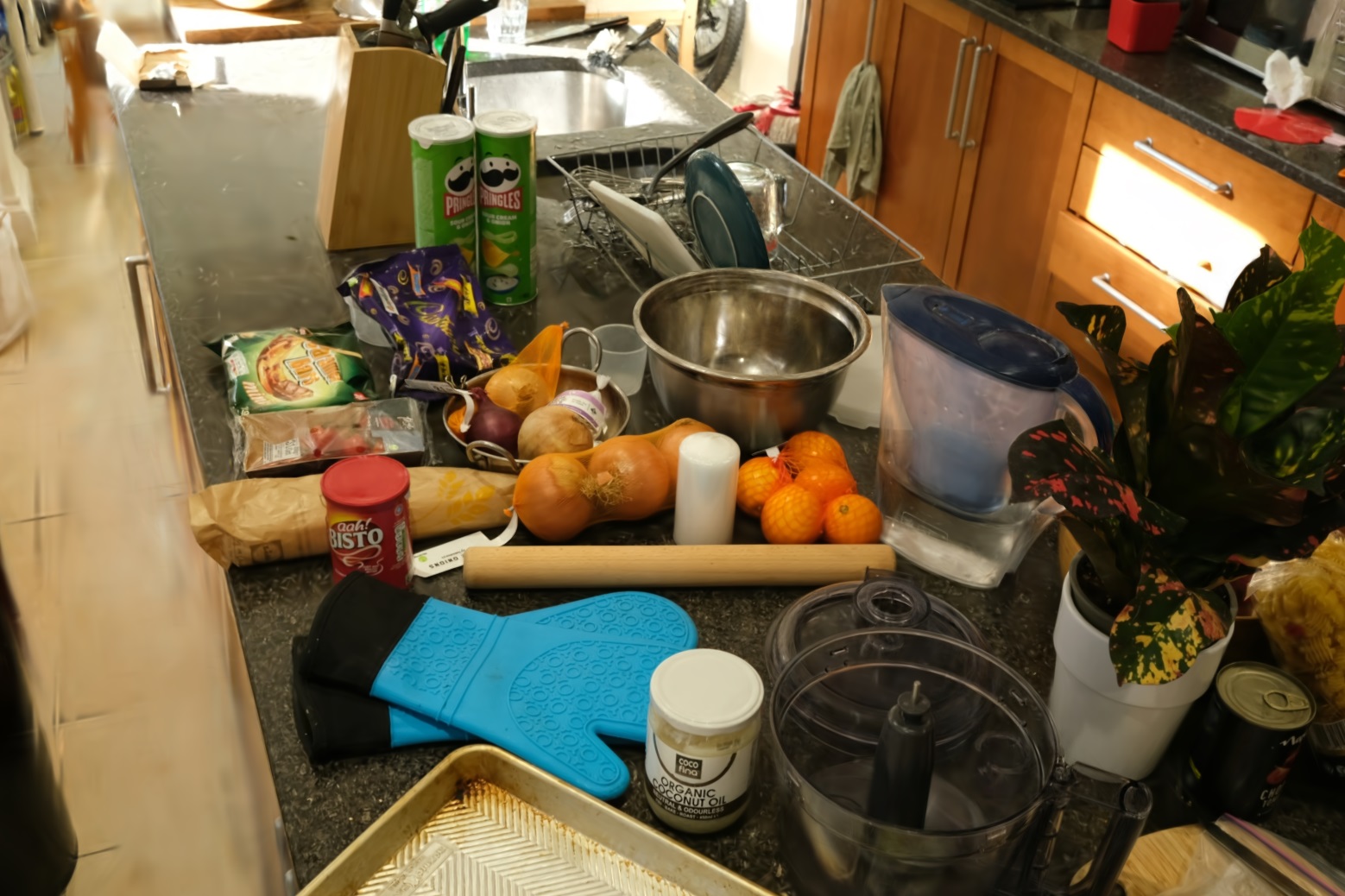}
\end{subfigure}

\vspace{0.2cm}

\begin{subfigure}[t]{.19\textwidth}
    \centering
    \includegraphics[trim=0.2\imagewidth{} 0.55\imageheight{} 0.6\imagewidth{} 0.18\imageheight{},clip, height=2.9cm]{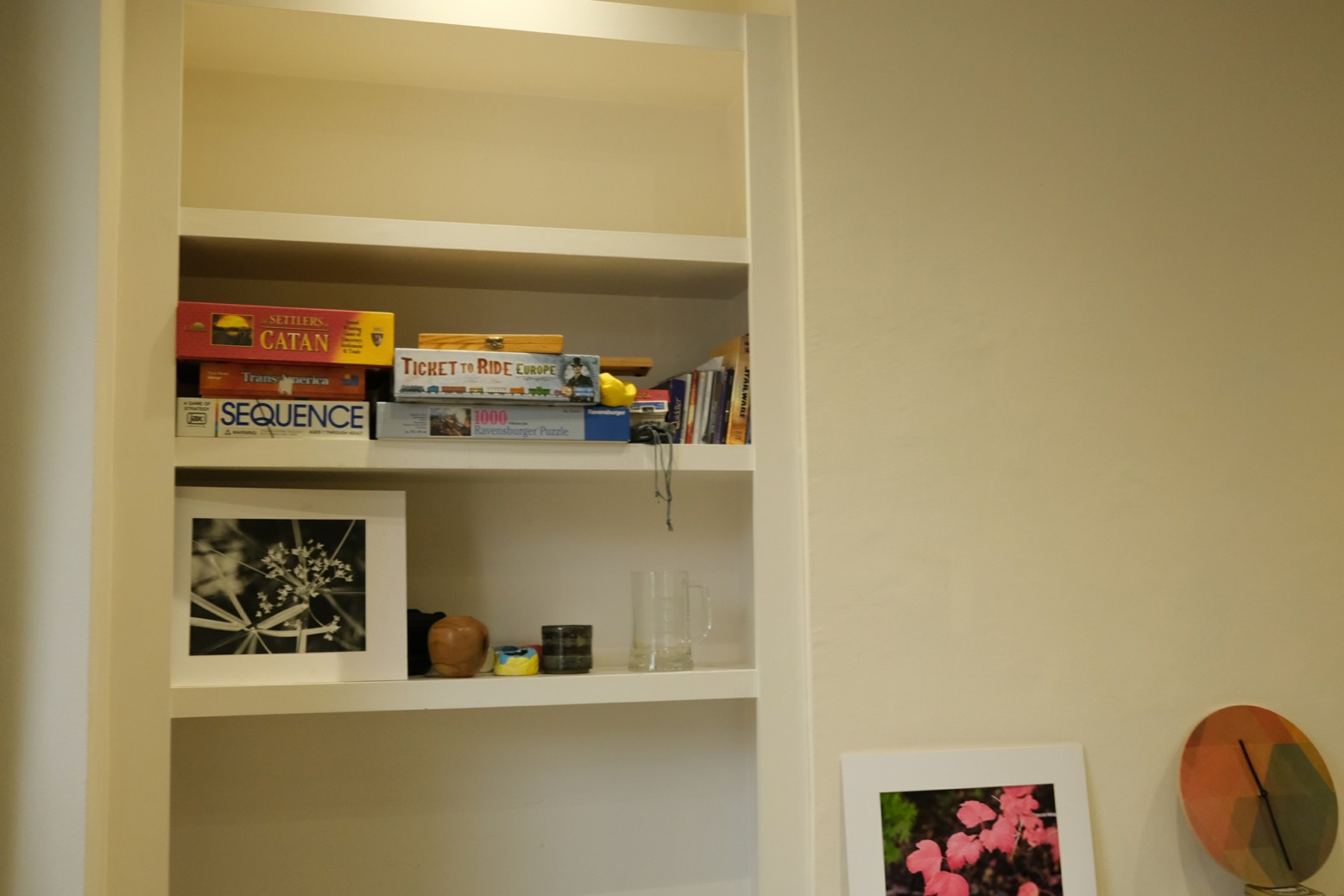}
    \caption{Ground Truth}
\end{subfigure}
\begin{subfigure}[t]{.19\textwidth}
    \centering
    \includegraphics[trim=0.2\imagewidth{} 0.55\imageheight{} 0.6\imagewidth{} 0.18\imageheight{},clip, height=2.9cm]{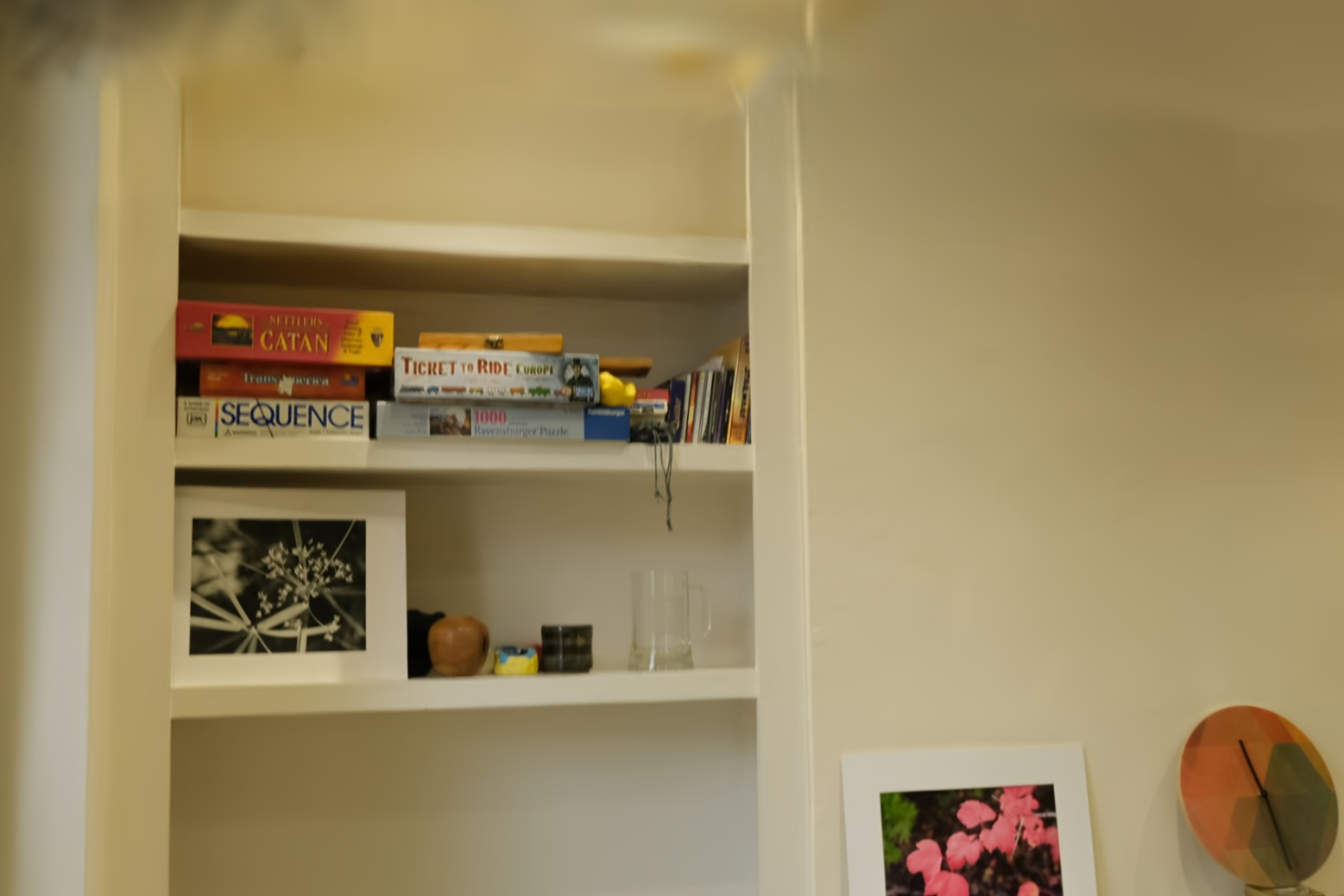}
    \caption{RayGaussX (ours)}
\end{subfigure}
\begin{subfigure}[t]{.19\textwidth}
    \centering
    \includegraphics[trim=0.2\imagewidth{} 0.55\imageheight{} 0.6\imagewidth{} 0.18\imageheight{},clip, height=2.9cm]{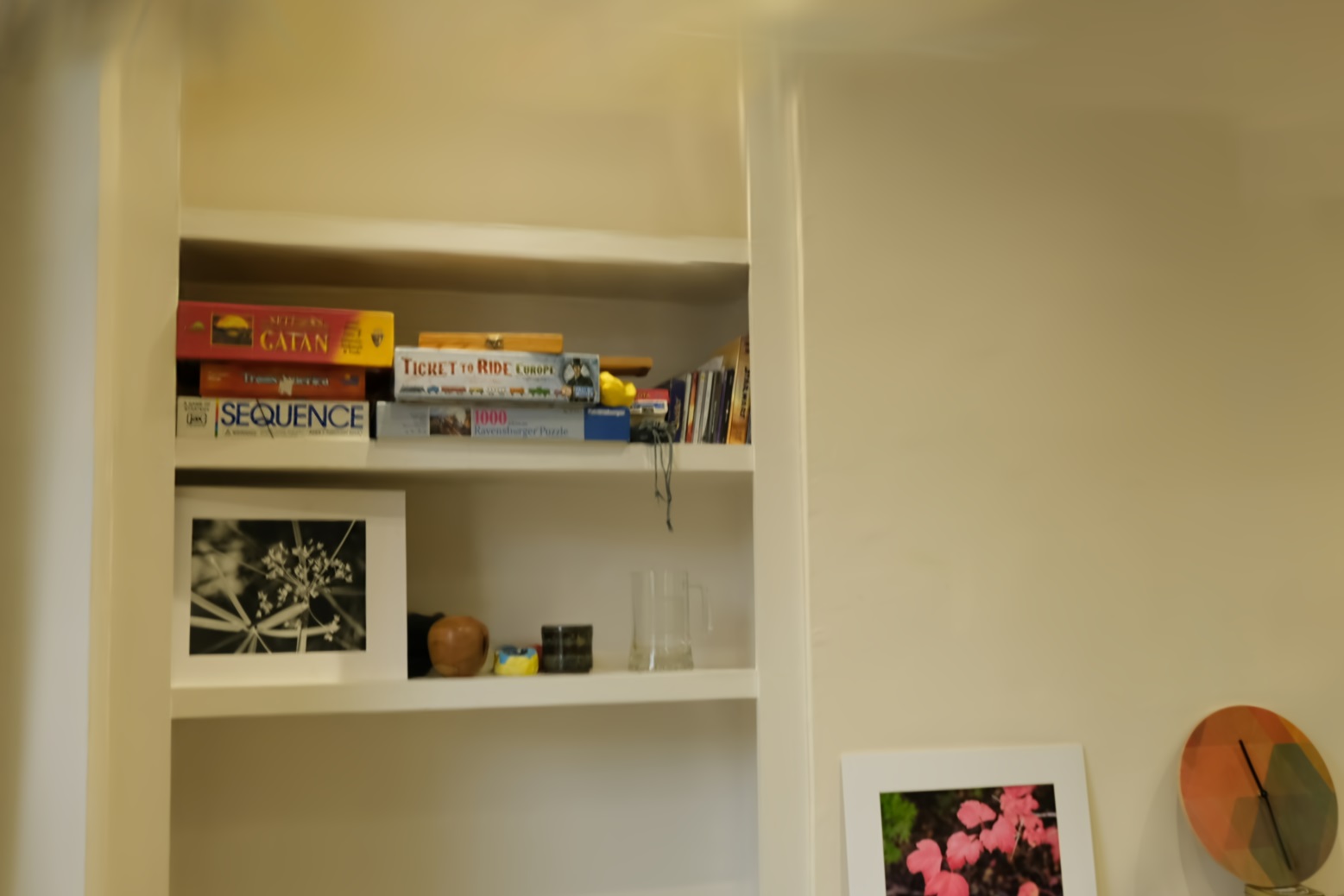}
     \caption{RayGauss~\cite{raygauss}}
\end{subfigure}
\begin{subfigure}[t]{.19\textwidth}
    \centering
    \includegraphics[trim=0.2\imagewidth{} 0.55\imageheight{} 0.6\imagewidth{} 0.18\imageheight{},clip, height=2.9cm]{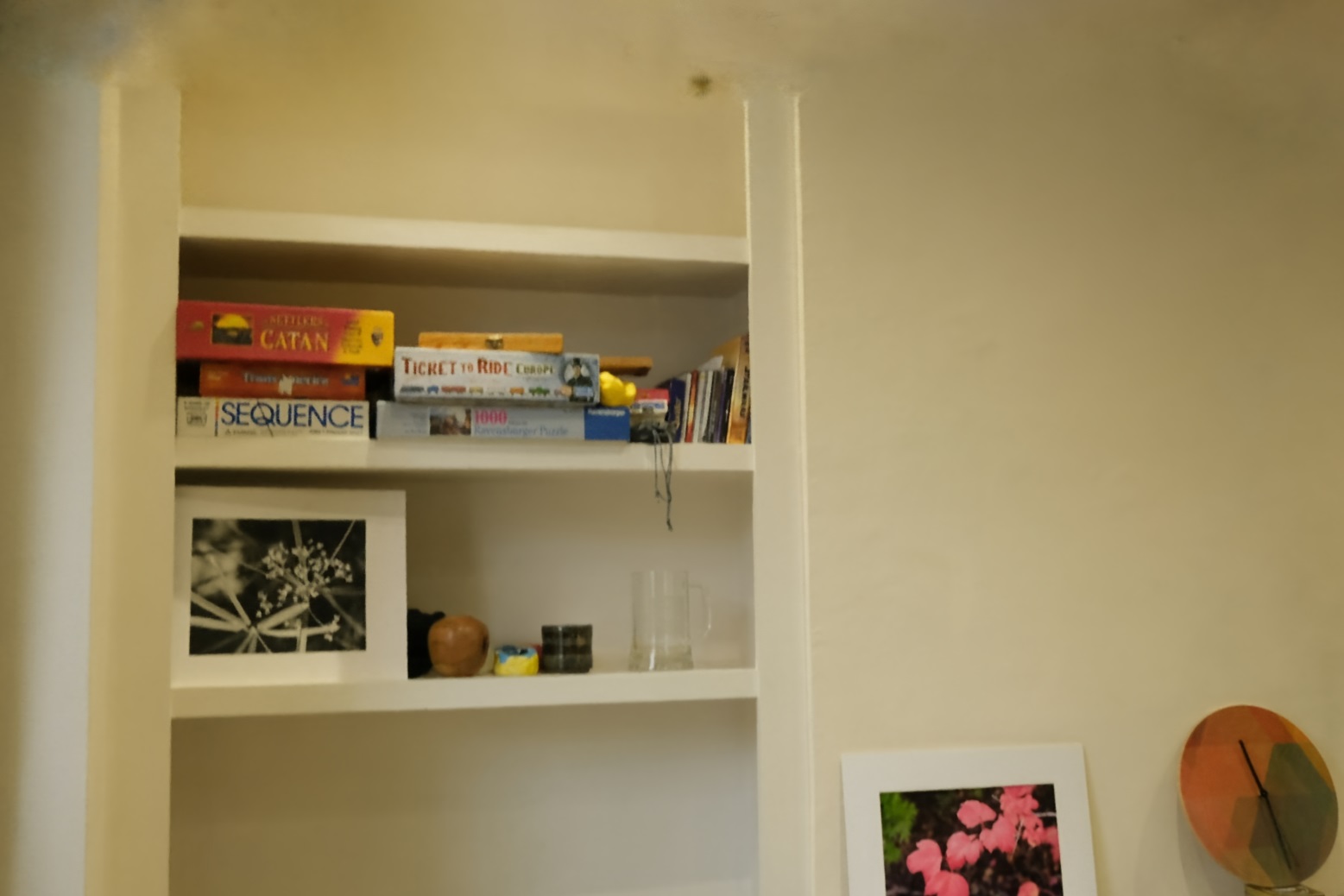}
    \caption{Zip-NeRF~\cite{zipnerf}}
\end{subfigure}
\begin{subfigure}[t]{.19\textwidth}
    \centering
    \includegraphics[trim=0.2\imagewidth{} 0.55\imageheight{} 0.6\imagewidth{} 0.18\imageheight{},clip, height=2.9cm]{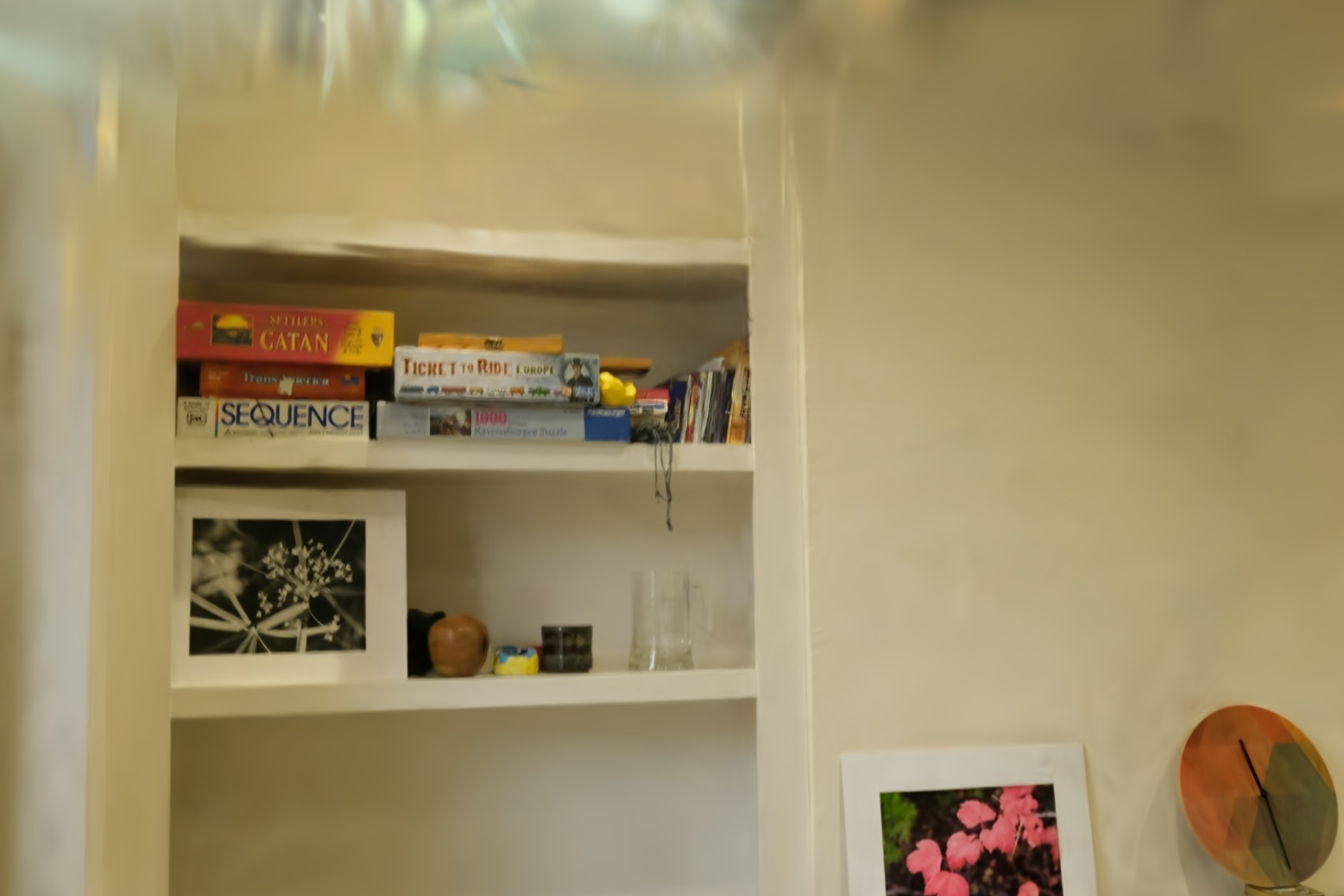}
    \caption{3D-GS~\cite{3D_Gaussian_Splatting}}
\end{subfigure}

\vspace{1.0cm}

\begin{subfigure}[t]{.19\textwidth}
    \centering
    \includegraphics[trim=0.495\imagewidth{} 0.6\imageheight{} 0.18\imagewidth{} 0.0\imageheight{},clip, height=2.9cm]{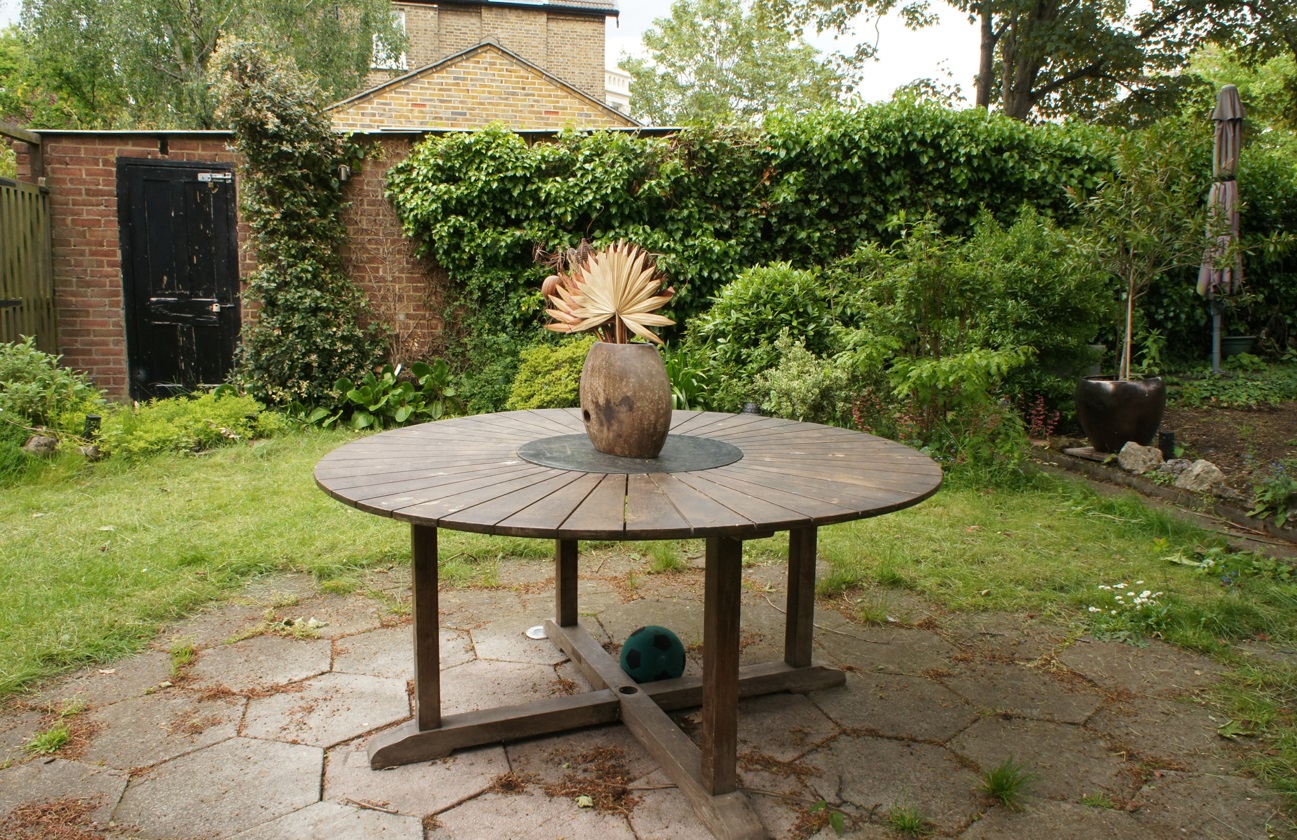}
\end{subfigure}
\begin{subfigure}[t]{.19\textwidth}
    \centering
    \includegraphics[trim=0.495\imagewidth{} 0.6\imageheight{} 0.18\imagewidth{} 0.0\imageheight{},clip, height=2.9cm]{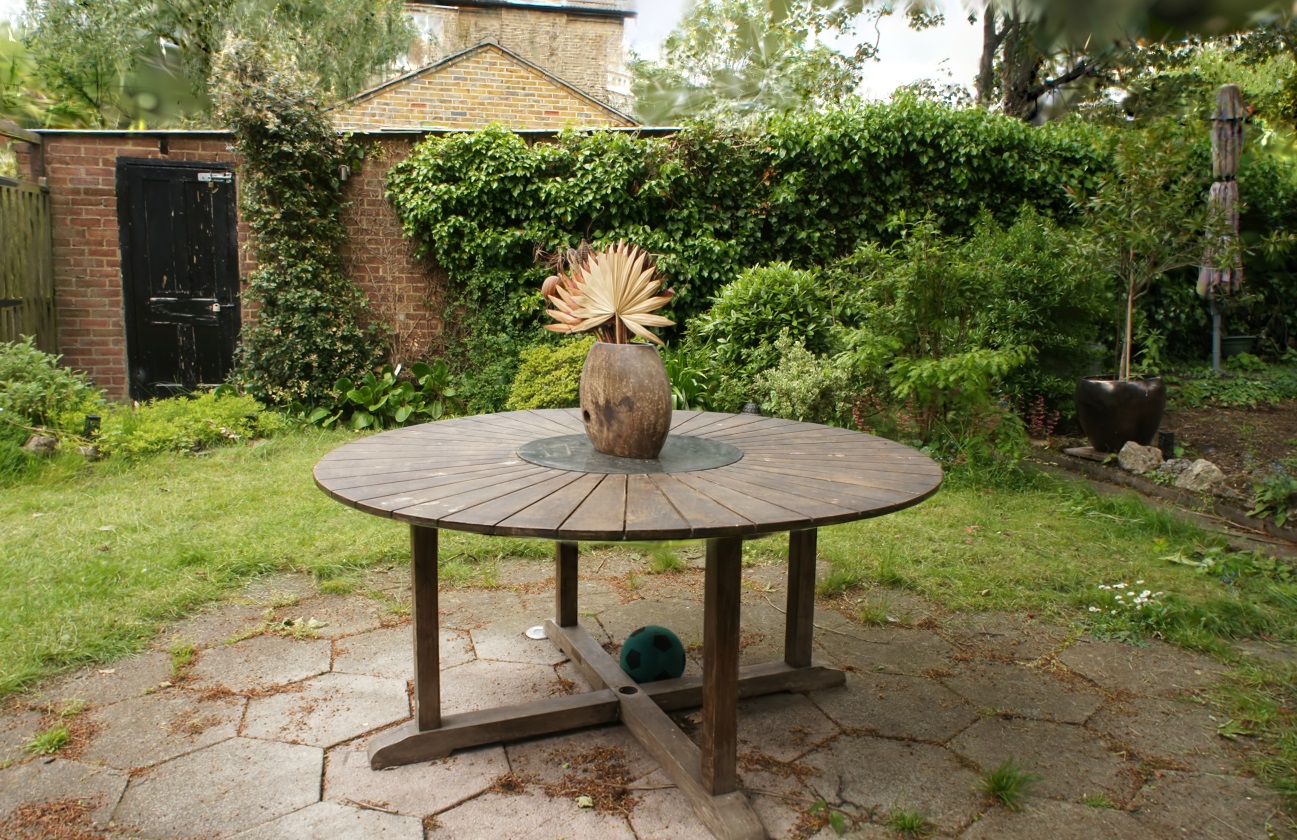}
\end{subfigure}
\begin{subfigure}[t]{.19\textwidth}
    \centering
    \includegraphics[trim=0.495\imagewidth{} 0.6\imageheight{} 0.18\imagewidth{} 0.0\imageheight{},clip, height=2.9cm]{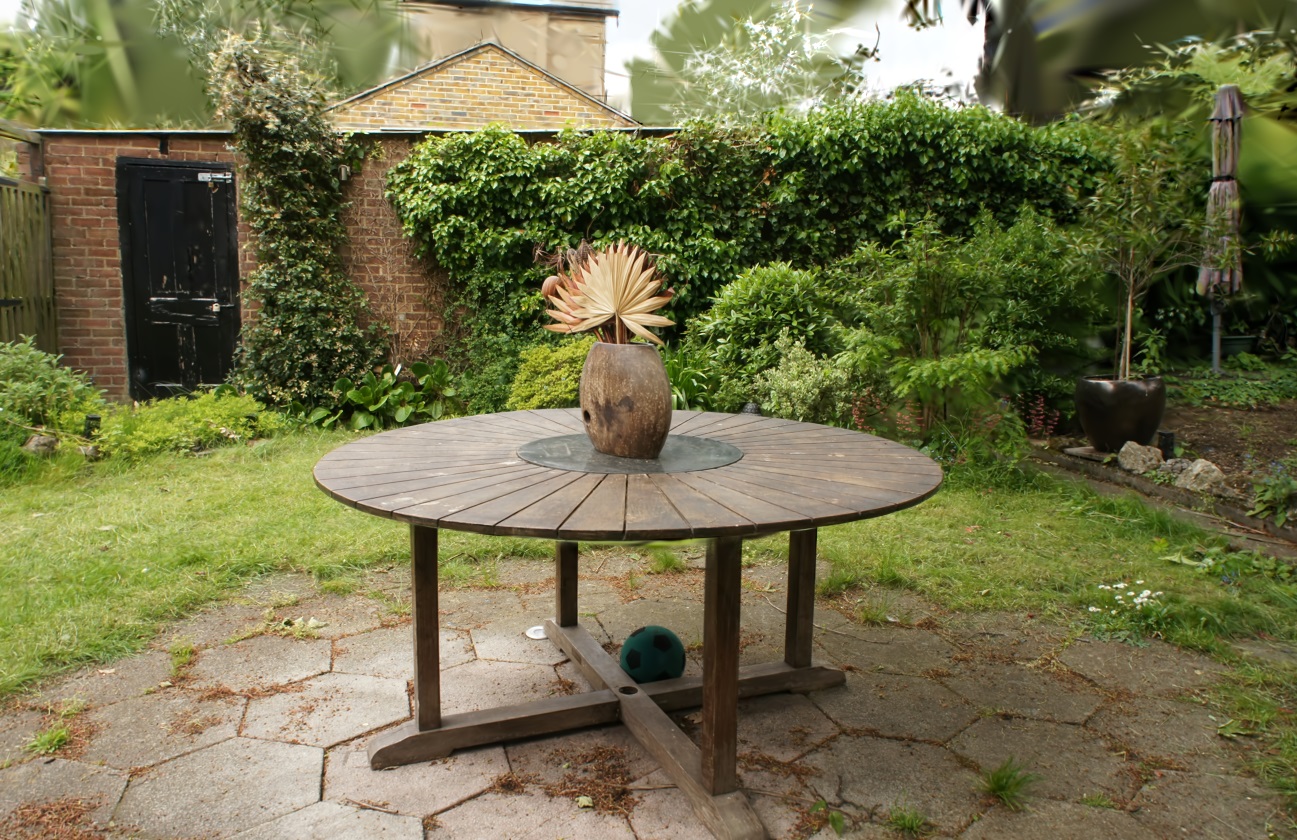}
\end{subfigure}
\begin{subfigure}[t]{.19\textwidth}
    \centering
    \includegraphics[trim=0.495\imagewidth{} 0.6\imageheight{} 0.18\imagewidth{} 0.0\imageheight{},clip, height=2.9cm]{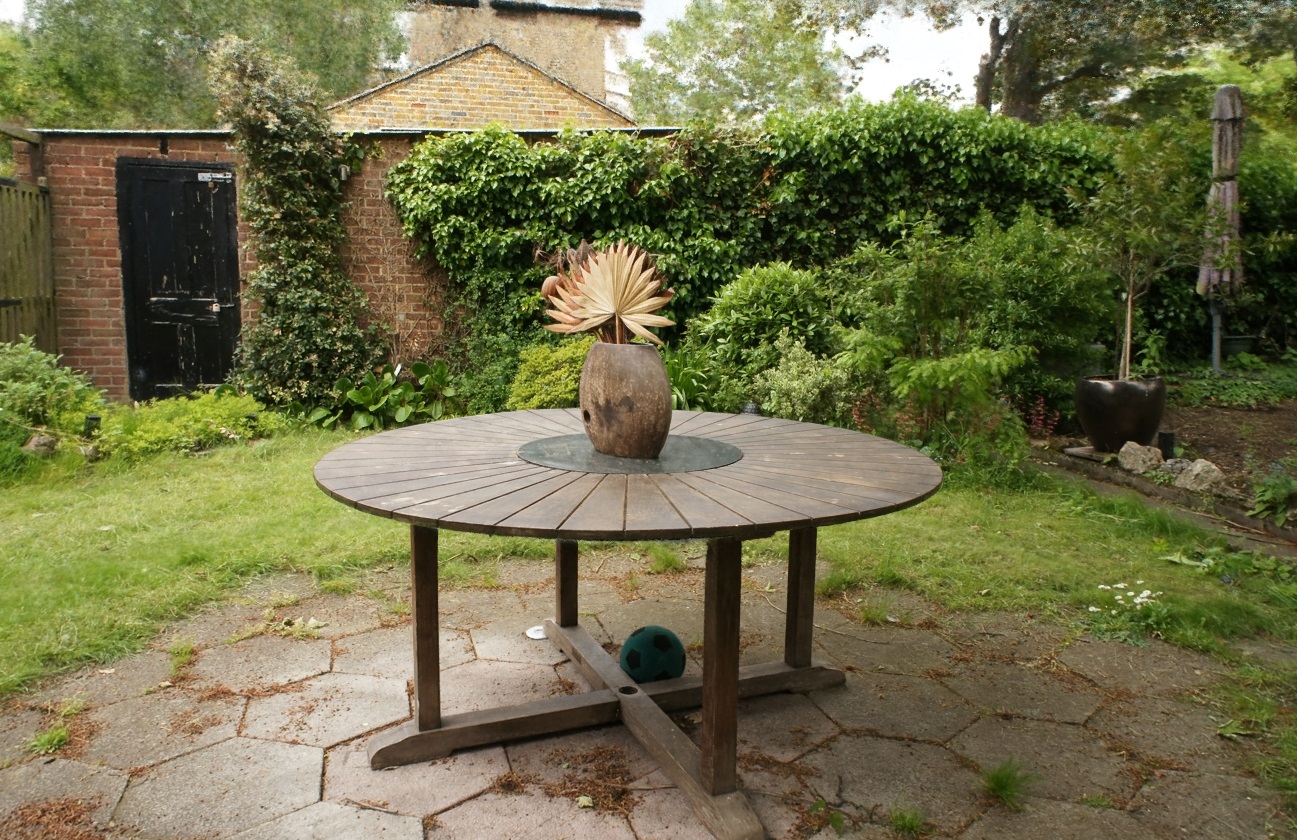}
\end{subfigure}
\begin{subfigure}[t]{.19\textwidth}
    \centering
    \includegraphics[trim=0.495\imagewidth{} 0.6\imageheight{} 0.18\imagewidth{} 0.0\imageheight{},clip, height=2.9cm]{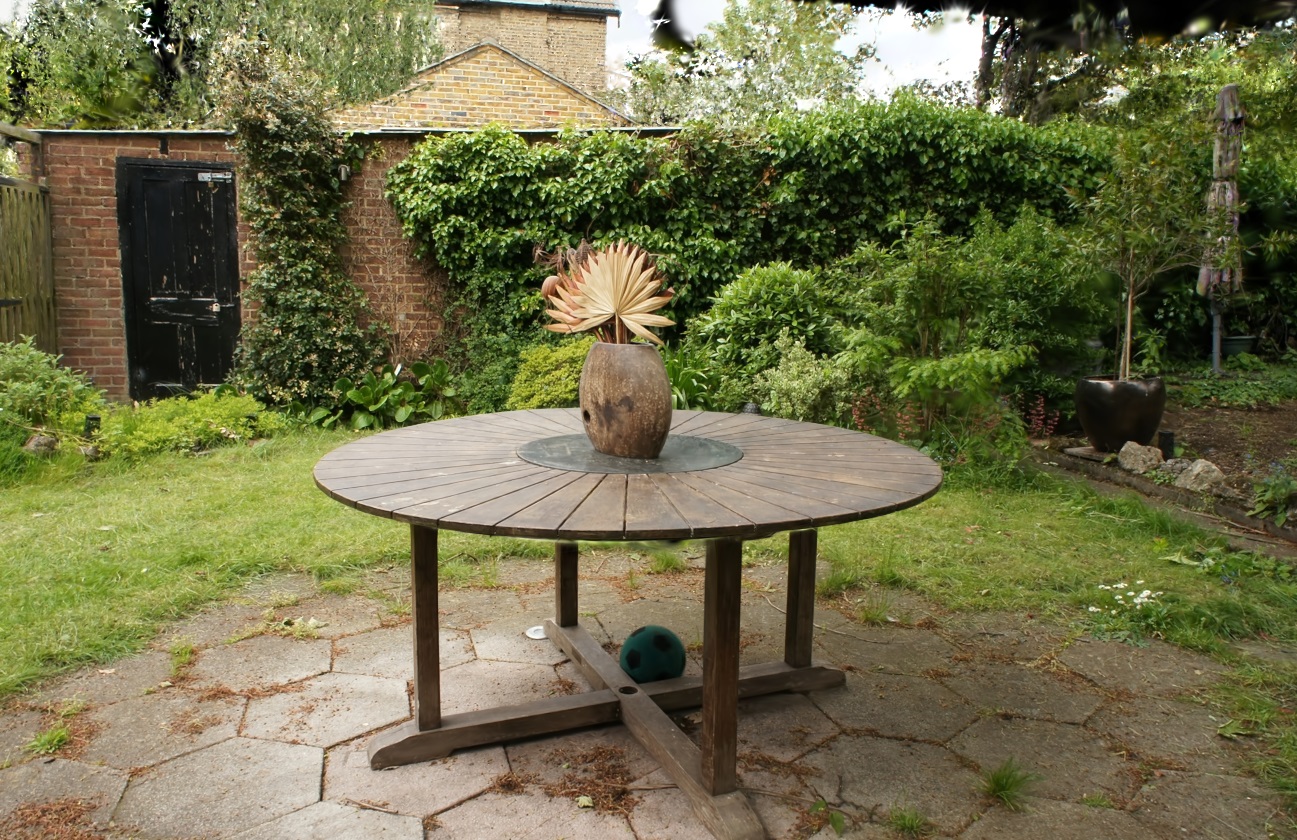}
\end{subfigure}

\vspace{0.2cm}

\begin{subfigure}[t]{.19\textwidth}
    \centering
    \includegraphics[trim=0.498\imagewidth{} 0.4\imageheight{} 0.15\imagewidth{} 0.2\imageheight{},clip, height=2.9cm]{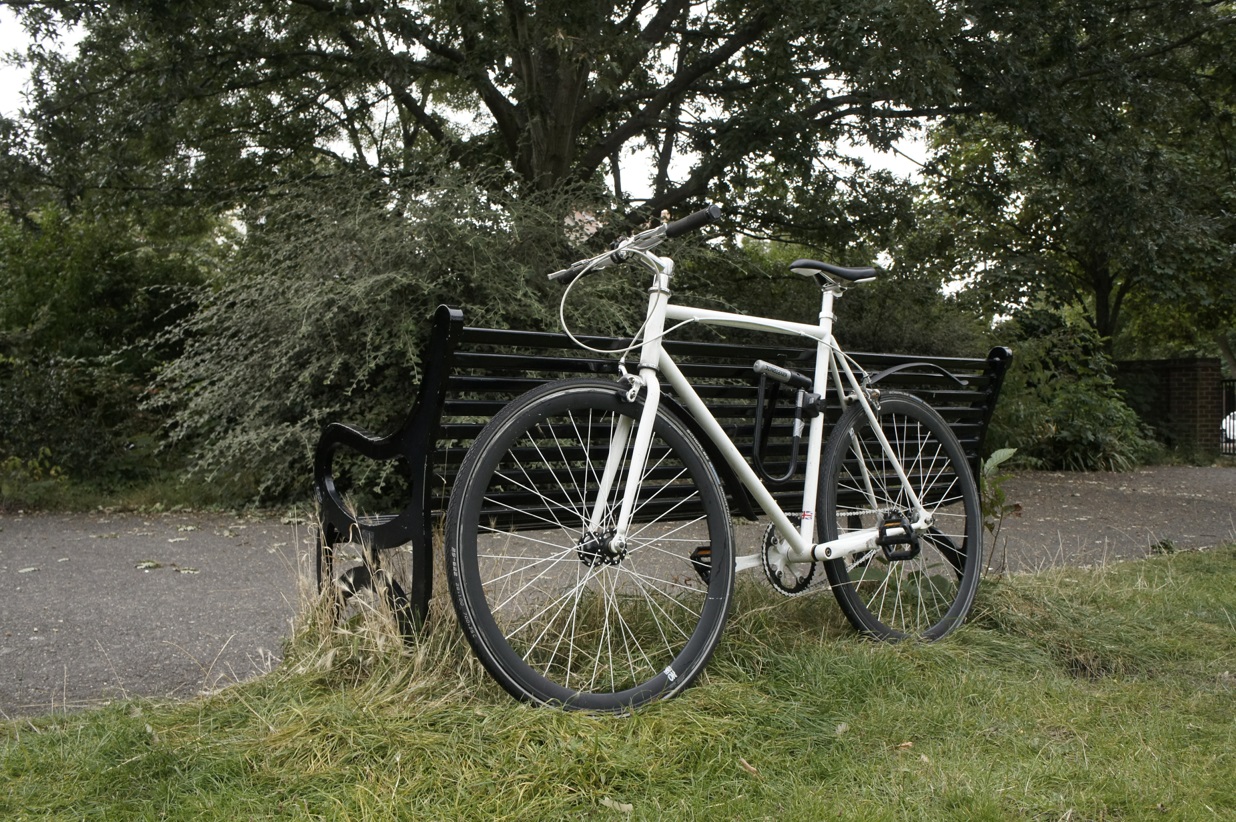}
\end{subfigure}
\begin{subfigure}[t]{.19\textwidth}
    \centering
    \includegraphics[trim=0.498\imagewidth{} 0.4\imageheight{} 0.15\imagewidth{} 0.2\imageheight{},clip, height=2.9cm]{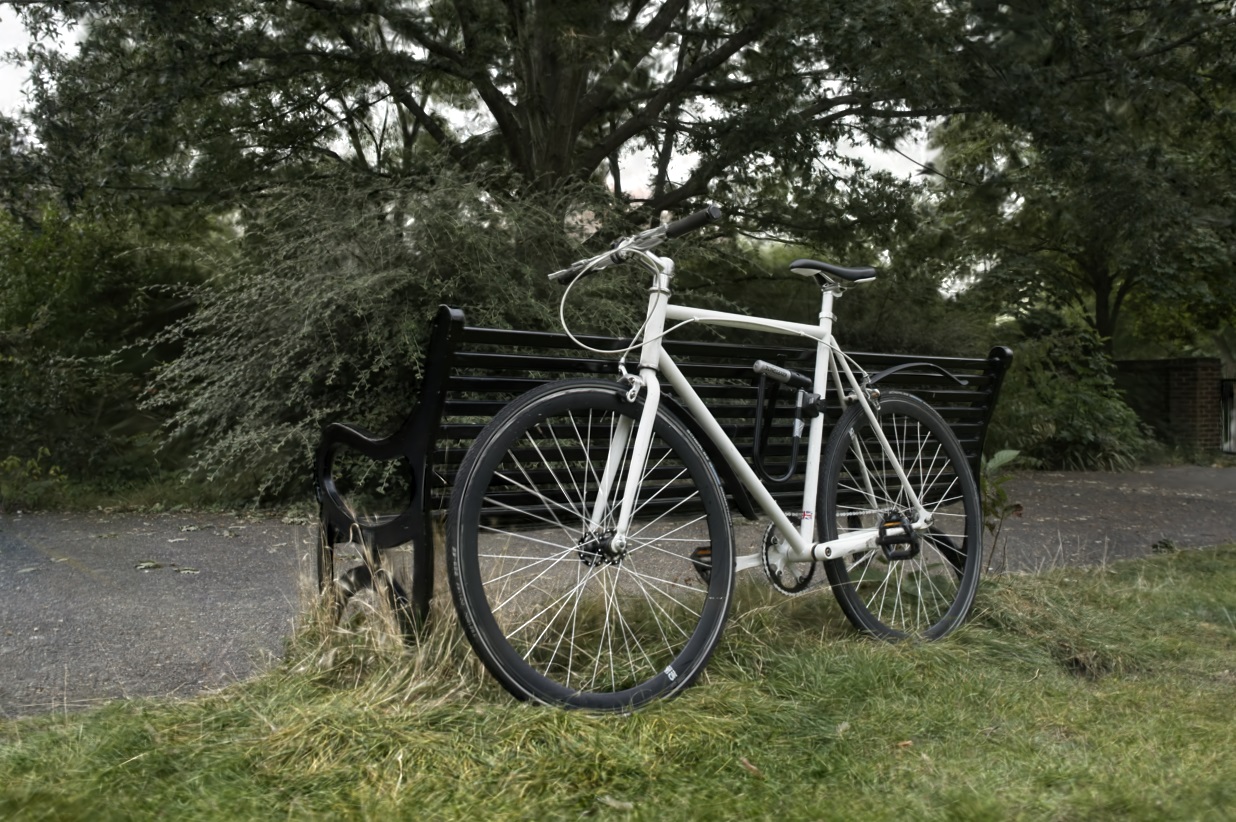}
\end{subfigure}
\begin{subfigure}[t]{.19\textwidth}
    \centering
    \includegraphics[trim=0.498\imagewidth{} 0.4\imageheight{} 0.15\imagewidth{} 0.2\imageheight{},clip, height=2.9cm]{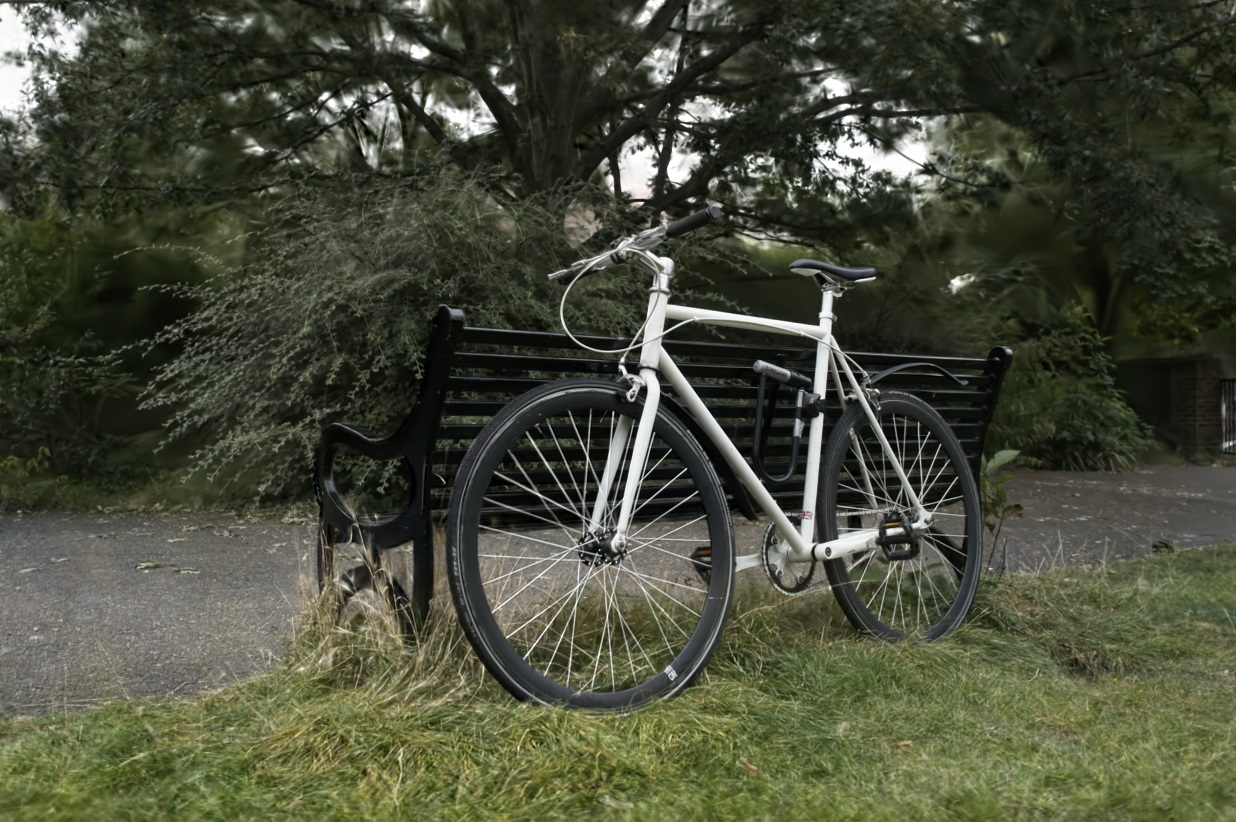}
\end{subfigure}
\begin{subfigure}[t]{.19\textwidth}
    \centering
    \includegraphics[trim=0.498\imagewidth{} 0.4\imageheight{} 0.15\imagewidth{} 0.2\imageheight{},clip, height=2.9cm]{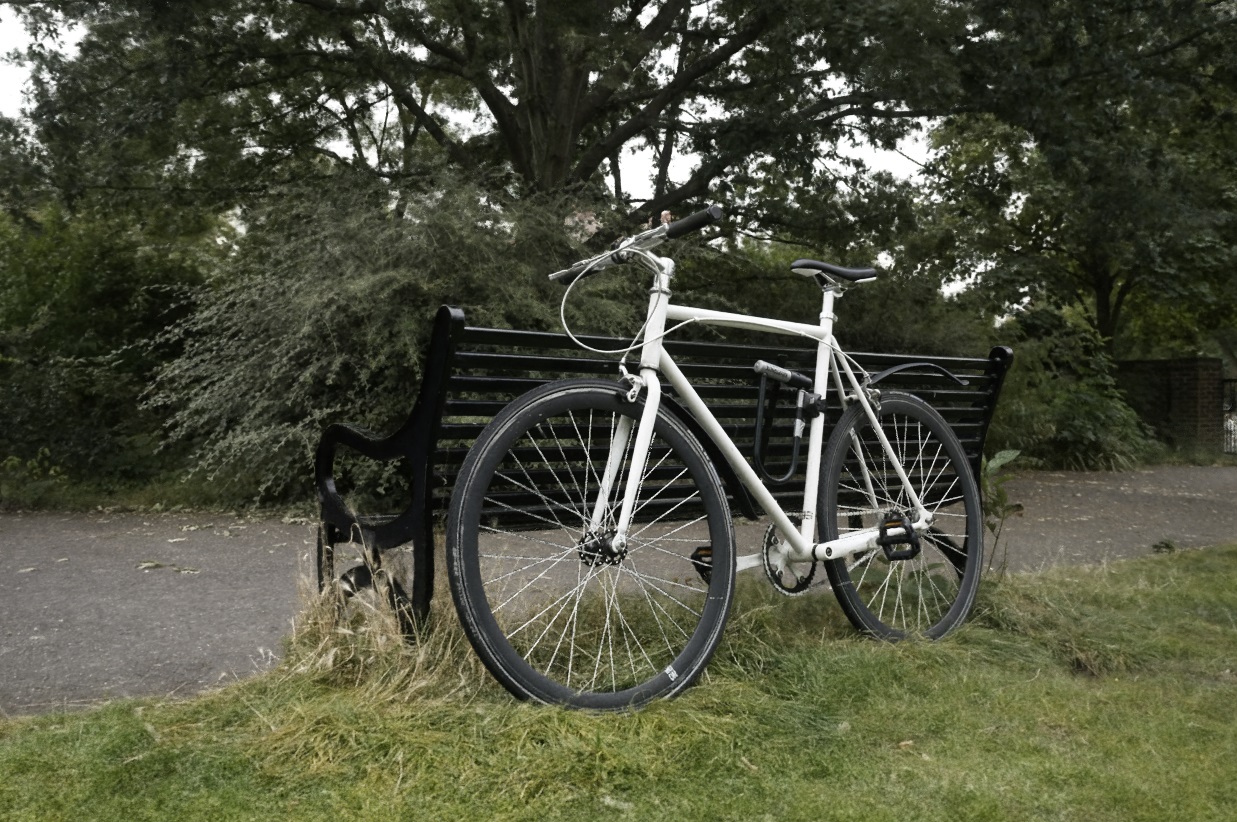}
\end{subfigure}
\begin{subfigure}[t]{.19\textwidth}
    \centering
    \includegraphics[trim=0.498\imagewidth{} 0.4\imageheight{} 0.15\imagewidth{} 0.2\imageheight{},clip, height=2.9cm]{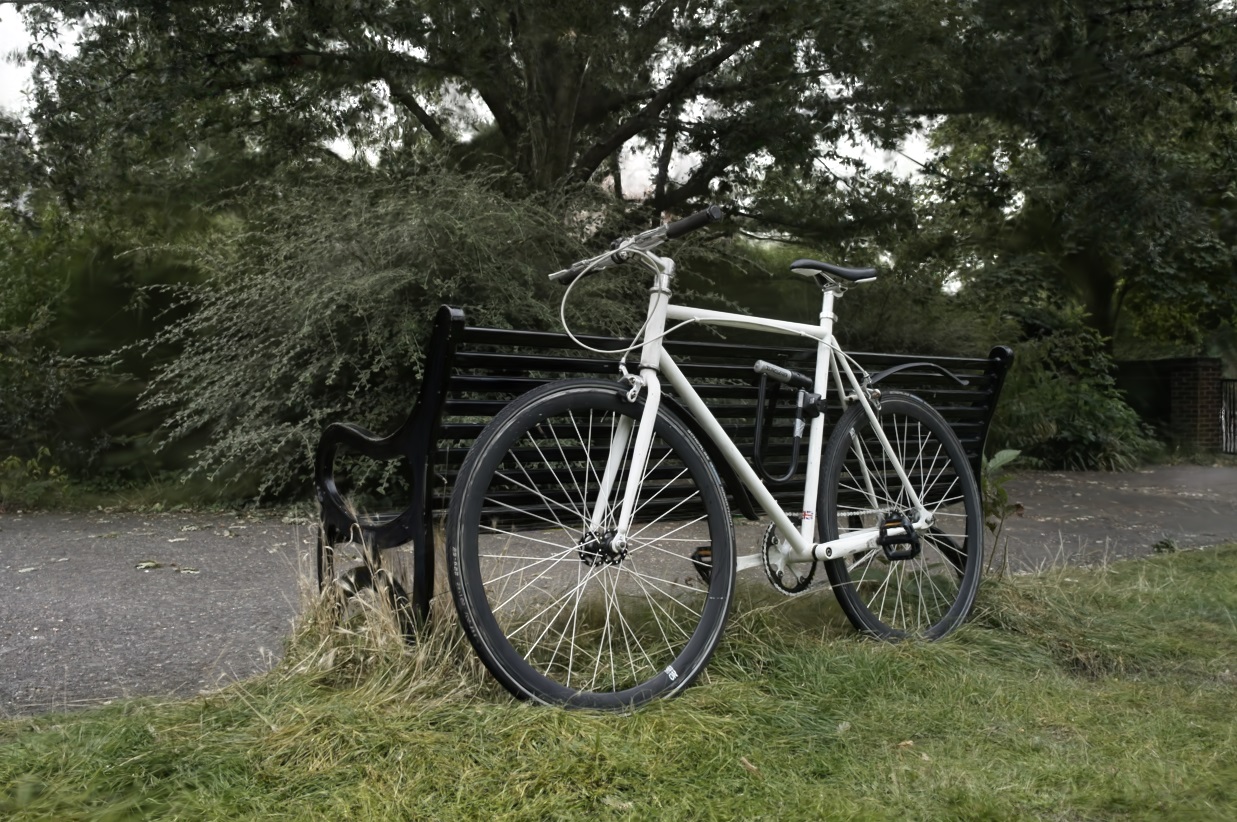}
\end{subfigure} \\

\vspace{0.2cm}

\begin{subfigure}[t]{.19\textwidth}
    \centering
    \includegraphics[trim=0.21\imagewidth{} 0.2\imageheight{} 0.45\imagewidth{} 0.4\imageheight{},clip, height=2.9cm]{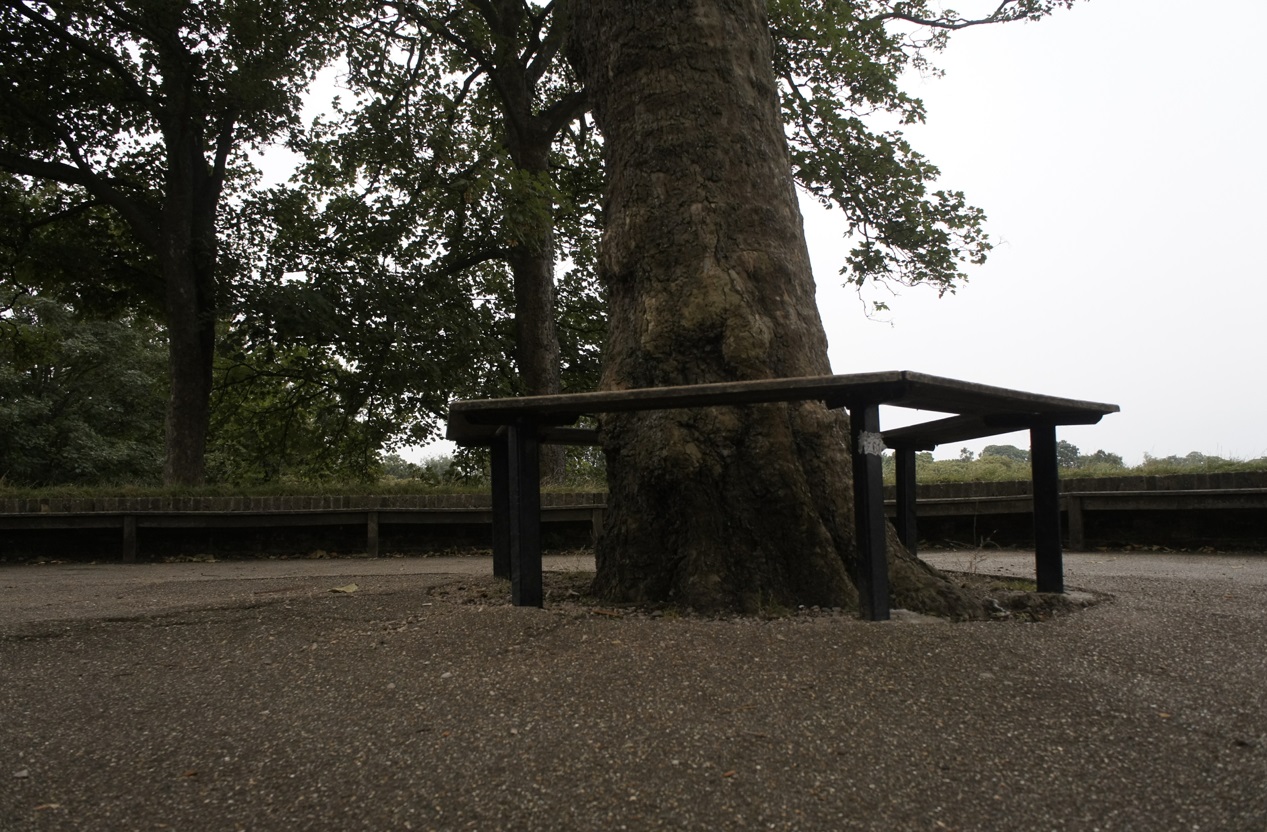}
    \caption{Ground Truth}
\end{subfigure}
\begin{subfigure}[t]{.19\textwidth}
    \centering
    \includegraphics[trim=0.21\imagewidth{} 0.2\imageheight{} 0.45\imagewidth{} 0.4\imageheight{},clip, height=2.9cm]{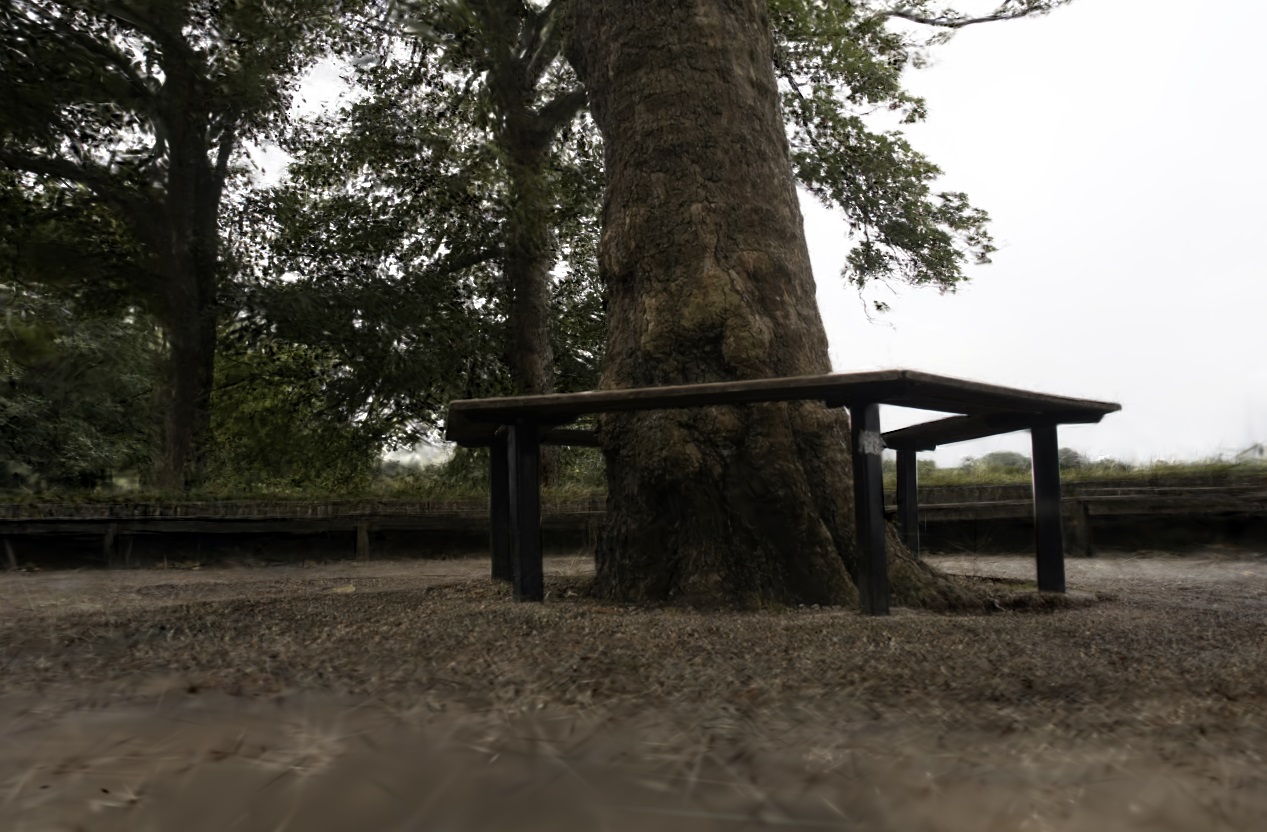}
    \caption{RayGaussX (ours)}
\end{subfigure}
\begin{subfigure}[t]{.19\textwidth}
    \centering
    \includegraphics[trim=0.21\imagewidth{} 0.2\imageheight{} 0.45\imagewidth{} 0.4\imageheight{},clip, height=2.9cm]{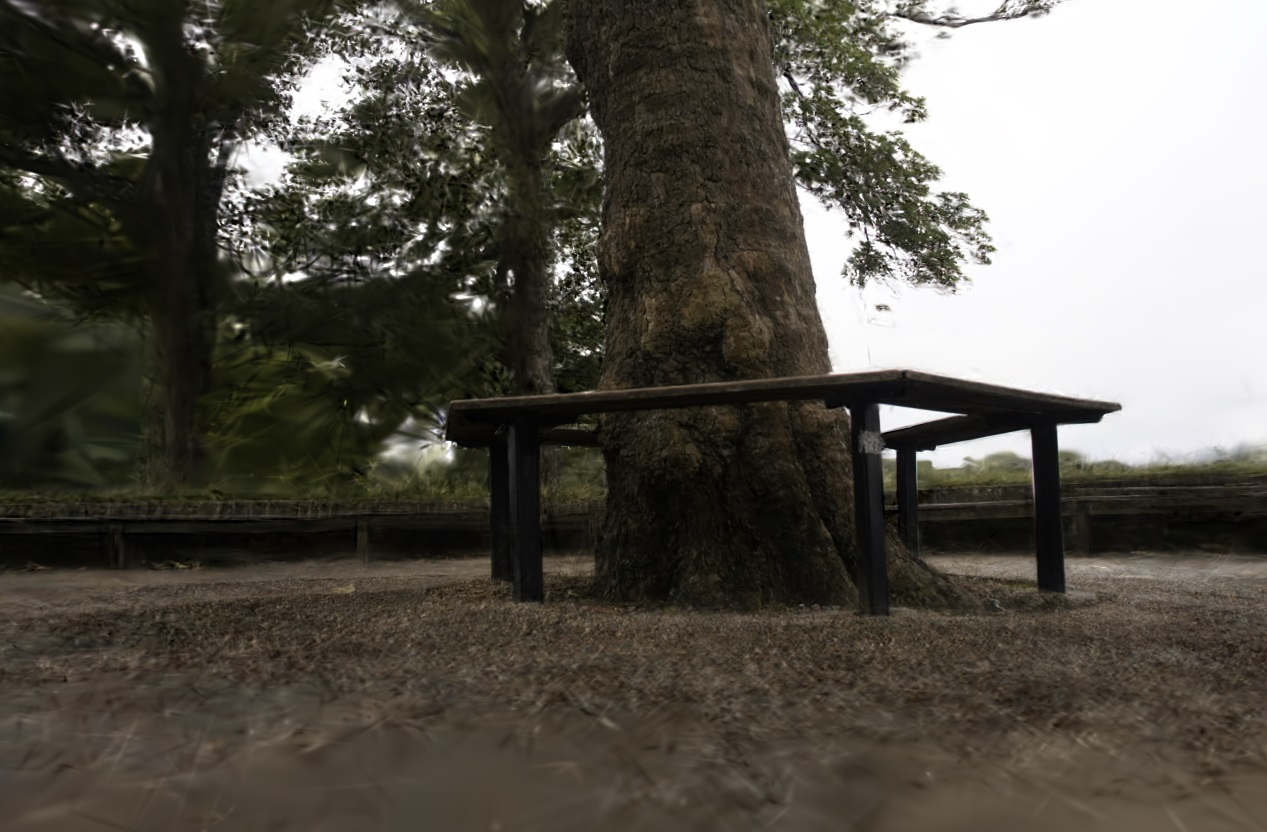}
    \caption{RayGauss~\cite{raygauss}}
\end{subfigure}
\begin{subfigure}[t]{.19\textwidth}
    \centering
    \includegraphics[trim=0.21\imagewidth{} 0.2\imageheight{} 0.45\imagewidth{} 0.4\imageheight{},clip, height=2.9cm]{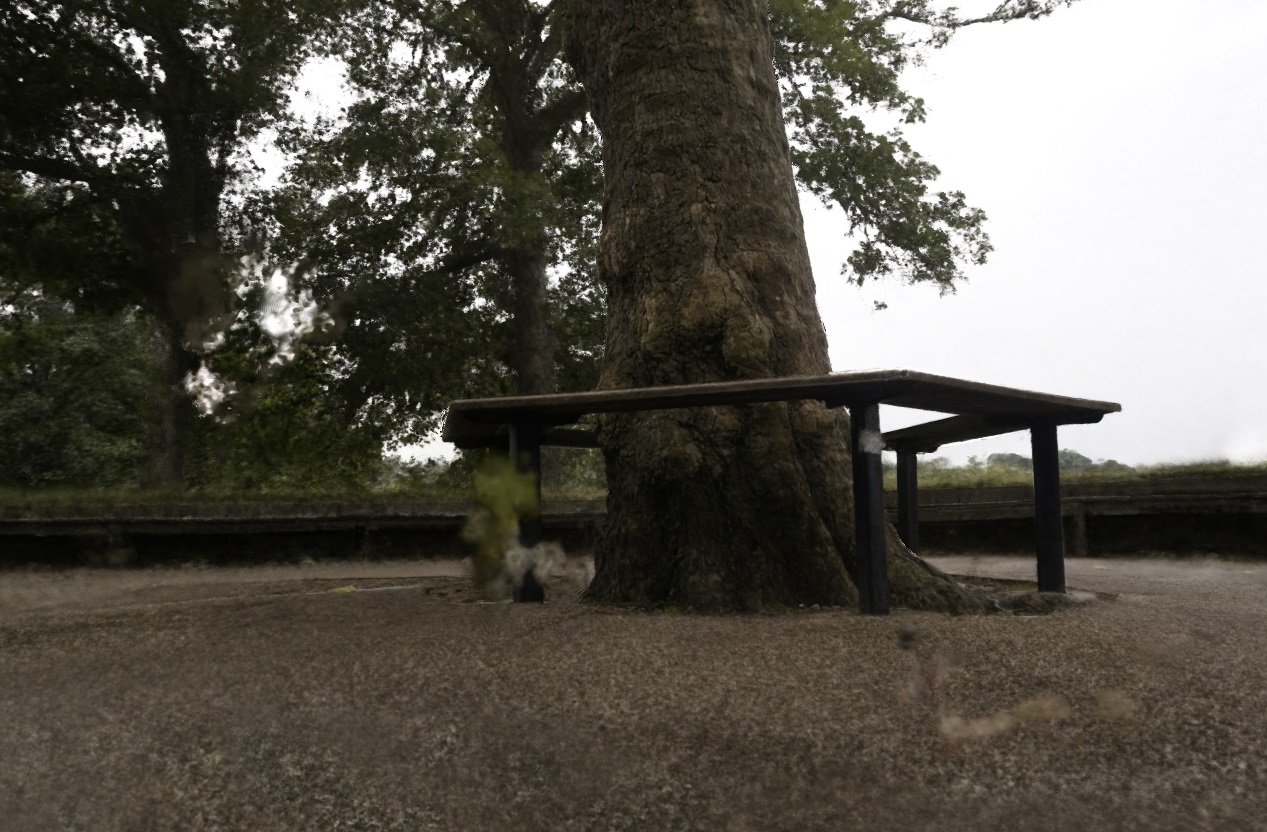}
    \caption{Zip-NeRF~\cite{zipnerf}}
\end{subfigure}
\begin{subfigure}[t]{.19\textwidth}
    \centering
    \includegraphics[trim=0.21\imagewidth{} 0.2\imageheight{} 0.45\imagewidth{} 0.4\imageheight{},clip, height=2.9cm]{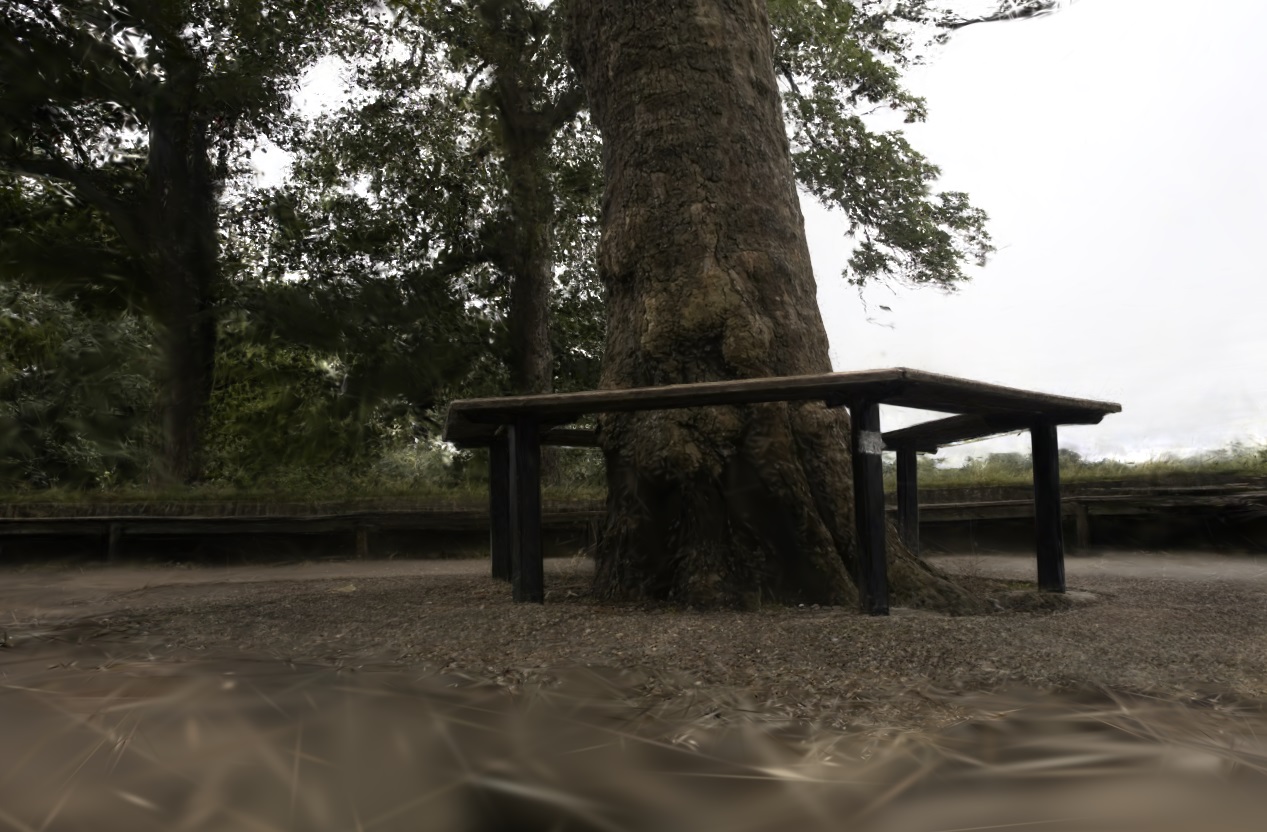}
    \caption{3D-GS~\cite{3D_Gaussian_Splatting}}
\end{subfigure} \\

\caption{Qualitative results on the Mip-NeRF360~\cite{Mip_NeRF_360} dataset. The scenes are from the top down: \textit{bonsai}, \textit{counter}, \textit{room}, \textit{garden}, \textit{bicycle}, \textit{treehill}.} 
\label{fig:qualtitative_mipnerf360}
\end{figure*}

\subsection{Detailed quantitative results}
\label{sec:detailed_quantitative_results}

Tab.~\ref{tab:detailed_blender}, Tab.~\ref{tab:detailed_nsvf}, Tab.~\ref{tab:detailed_mipnerf360}, and Tab.~\ref{tab:detailed_tandt_db} show the detailed per scene results of the main paper with metrics PSNR, SSIM, and LPIPS on the NeRF Synthetic~\cite{Neural_Radiance_Fields}, NSVF Synthetic~\cite{NSVF}, Mip-NeRF360~\cite{Mip_NeRF_360}, Tanks\&Temples~\cite{Knapitsch2017} and Deep Blending~\cite{Hedman2018} datasets. 

All methods with an * in tables of the main paper and supplementary have been re-trained using the available online code of respective methods:
\begin{itemize}
 \item 3D Gaussian Splatting: \url{https://github.com/graphdeco-inria/gaussian-splatting}
 \item Mip-Splatting: \url{https://github.com/autonomousvision/mip-splatting}
 \item Spec-Gaussian: \url{https://github.com/ingra14m/Specular-Gaussians}
 \item Zip-NeRF: \url{https://github.com/jonbarron/camp_zipnerf}
 \item RayGauss: \url{https://github.com/hugobl1/ray_gauss}
\end{itemize}

\begin{table*}[ht]
  \centering
  \resizebox{0.8\linewidth}{!}{%
  \begin{tabular}{l | c c c c c c c c | c }
   & \multicolumn{8}{c}{\textbf{PSNR ↑}} \\
    & chair & drums & ficus & hotdog & lego & materials & mic & ship & \textbf{Avg.}\\
    \hline
Instant-NGP~\cite{Instant_NGP} & 35.00 & 26.02 & 33.51 & 37.40 & 36.39 & 29.78 & 36.22 & 31.10 & 33.18 \\ 
Mip-NeRF360~\cite{Mip_NeRF_360} &  35.65 & 25.60 & 33.19 & 37.71 & 36.10 & 29.90 & \cellcolor{yellow!40}36.52 & 31.26 & 33.24 \\
Point-NeRF~\cite{Point_NeRF} & 35.40 & 26.06 & \cellcolor{orange!40}36.13 & 37.30 & 35.04 & 29.61 & 35.95 & 30.97 & 33.30 \\
3D-GS~\cite{3D_Gaussian_Splatting}* & 35.85 & 26.22 & 35.00 & 37.81 & 35.87 & 30.00 & 35.40 & 30.95 & 33.39\\
Zip-NeRF~\cite{zipnerf}* &  35.78 & 25.91 & 34.72 & 38.05 & 35.79 & 31.05 & 35.92 & \cellcolor{red!40}32.33 & 33.69 \\
Mip-Splatting~\cite{Yu2023MipSplatting}* & 36.23 & 26.29 & 35.33 & 38.04 & 36.13 & 30.37 & 36.22 & 31.32 & 33.74 \\
Spec-Gaussian~\cite{Spec-gaussian}* & 36.20 & \cellcolor{yellow!40}26.76 & \cellcolor{yellow!40}35.56 & 38.04 & 36.06 & 30.61 & 35.78 & 31.45 & 33.80 \\
3DGRT~\cite{3dgrt2024}* & 35.69 & 25.88 & \cellcolor{red!40}36.54 & 37.98 & 36.80 & 30.40 & 35.88 & 31.73 & 33.86 \\
NeuRBF~\cite{chen2023neurbf}*& \cellcolor{yellow!40}36.54 & 26.38 & 35.01 & \cellcolor{orange!40}38.44 & \cellcolor{red!40}37.35 & \cellcolor{red!40}34.12 & 36.16 & 31.73 & \cellcolor{yellow!40}34.47 \\
\hdashline
RayGauss~\cite{raygauss}* & \cellcolor{red!40}37.20 & \cellcolor{red!40}27.14 & 35.11 & \cellcolor{yellow!40}38.30 & \cellcolor{orange!40}37.10 & \cellcolor{orange!40}31.36 & \cellcolor{red!40}38.11 & \cellcolor{yellow!40}31.95 & \cellcolor{orange!40}34.53\\
    \hline
RayGaussX (ours) & \cellcolor{orange!40}37.19 & \cellcolor{orange!40}27.10 & 35.09 & \cellcolor{red!40}38.46 & \cellcolor{yellow!40}37.01 & \cellcolor{yellow!40}31.33 & \cellcolor{orange!40}38.02 & \cellcolor{orange!40}32.13 & \cellcolor{red!40}34.54\\
  \end{tabular}
  }
  
\bigskip
   
  \resizebox{0.8\linewidth}{!}{%
  \begin{tabular}{l | c c c c c c c c | c }
   & \multicolumn{8}{c}{\textbf{SSIM ↑}} \\
    & chair & drums & ficus & hotdog & lego & materials & mic & ship & \textbf{Avg.}\\
    \hline
Instant-NGP~\cite{Instant_NGP} & 0.979 & 0.937 & 0.981 & 0.982 & 0.982 & 0.951 & 0.990 & 0.896 & 0.963 \\
Mip-NeRF360~\cite{Mip_NeRF_360} & 0.983 & 0.931 &  0.979 & 0.982&  0.980 & 0.949 & 0.991 & 0.893 & 0.961\\
Point-NeRF~\cite{Point_NeRF} & 0.984 & 0.935 & \cellcolor{yellow!40}0.987 & 0.982 & 0.978 & 0.948 & 0.990 & 0.892 & 0.962\\
3D-GS~\cite{3D_Gaussian_Splatting}* & \cellcolor{orange!40}0.988 & \cellcolor{yellow!40}0.955 & \cellcolor{orange!40}0.988 & \cellcolor{yellow!40}0.986 &  0.983 & 0.960 & \cellcolor{yellow!40}0.992 & 0.893 & 0.968  \\
Zip-NeRF~\cite{zipnerf}* & \cellcolor{yellow!40}0.987 & 0.948 & \cellcolor{yellow!40}0.987 & \cellcolor{orange!40}0.987 & 0.983 & \cellcolor{yellow!40}0.968 & \cellcolor{yellow!40}0.992 & \cellcolor{red!40}0.937 & \cellcolor{red!40}0.974 \\
Mip-Splatting~\cite{Yu2023MipSplatting}* & \cellcolor{orange!40}0.988 & \cellcolor{yellow!40}0.955 & \cellcolor{orange!40}0.988 & \cellcolor{orange!40}0.987 & \cellcolor{yellow!40}0.984 & 0.963 & \cellcolor{orange!40}0.993 & 0.908 & \cellcolor{orange!40}0.971  \\
Spec-Gaussian~\cite{Spec-gaussian}* & \cellcolor{yellow!40}0.987 & \cellcolor{yellow!40}0.955 & \cellcolor{orange!40}0.988 & \cellcolor{yellow!40}0.986 & 0.983 & 0.964 & \cellcolor{yellow!40}0.992 & 0.906 & \cellcolor{yellow!40}0.970\\
3DGRT~\cite{3dgrt2024}* & \cellcolor{yellow!40}0.987 & 0.954 & \cellcolor{red!40}0.989 & \cellcolor{yellow!40}0.986 & \cellcolor{orange!40}0.985 & 0.961 & 0.991 & 0.909 & \cellcolor{yellow!40}0.970 \\
NeuRBF\cite{chen2023neurbf}* & \cellcolor{orange!40}0.988 & 0.944 & \cellcolor{yellow!40}0.987 & \cellcolor{orange!40}0.987 & \cellcolor{red!40}0.986 & \cellcolor{red!40}0.979 & \cellcolor{yellow!40}0.992 & \cellcolor{orange!40}0.925 & \cellcolor{red!40}0.974 \\
\hdashline
RayGauss~\cite{raygauss}* & \cellcolor{red!40}0.990 & \cellcolor{red!40}0.960 & \cellcolor{orange!40}0.988 & \cellcolor{red!40}0.988 & \cellcolor{red!40}0.986 & \cellcolor{orange!40}0.969 & \cellcolor{red!40}0.995 & \cellcolor{yellow!40}0.914 & \cellcolor{red!40}0.974 \\
    \hline
RayGaussX (ours) & \cellcolor{red!40}0.990 & \cellcolor{orange!40}0.959 & \cellcolor{orange!40}0.988 & \cellcolor{red!40}0.988 & \cellcolor{red!40}0.986 & \cellcolor{orange!40}0.969 & \cellcolor{red!40}0.995 & \cellcolor{yellow!40}0.914 & \cellcolor{red!40}0.974\\
  \end{tabular}
  }

\bigskip

  \resizebox{0.8\linewidth}{!}{%
  \begin{tabular}{l | c c c c c c c c | c }
   & \multicolumn{8}{c}{\textbf{LPIPS ↓}} \\
   & chair & drums & ficus & hotdog & lego & materials & mic & ship & \textbf{Avg.}\\
    \hline
Instant-NGP~\cite{Instant_NGP} & 0.022 & 0.071 & 0.023 & 0.027 & 0.017 & 0.060 & 0.010 & 0.132 & 0.045 \\
Mip-NeRF360~\cite{Mip_NeRF_360} & 0.018 & 0.069 & 0.022 & 0.024 & 0.018 &  0.053 & 0.011 & 0.119 & 0.042 \\
Point-NeRF~\cite{Point_NeRF} & 0.023 & 0.078 & 0.022 & 0.037 & 0.024 & 0.072 & 0.014 & 0.124 & 0.049\\
3D-GS~\cite{3D_Gaussian_Splatting}* & \cellcolor{orange!40}0.011 & \cellcolor{yellow!40}0.037 & \cellcolor{orange!40}0.011 & \cellcolor{orange!40}0.017 & \cellcolor{orange!40}0.015 & 0.034 & \cellcolor{yellow!40}0.006 & 0.118 & 0.031  \\
Zip-NeRF~\cite{zipnerf}* & \cellcolor{yellow!40}0.013 & 0.045 & \cellcolor{yellow!40}0.013 & \cellcolor{orange!40}0.017 &\cellcolor{orange!40}0.015 & \cellcolor{yellow!40}0.031 & \cellcolor{yellow!40}0.006 & \cellcolor{red!40}0.082 & \cellcolor{yellow!40}0.028\\
Mip-Splatting~\cite{Yu2023MipSplatting}* & \cellcolor{yellow!40}0.013 & 0.038 & \cellcolor{orange!40}0.011 & \cellcolor{orange!40}0.017 & \cellcolor{orange!40}0.015 & 0.033 & \cellcolor{yellow!40}0.006 & 0.098 & 0.029 \\
Spec-Gaussian~\cite{Spec-gaussian}* & \cellcolor{orange!40}0.011 & \cellcolor{orange!40}0.034 & \cellcolor{orange!40}0.011 & \cellcolor{orange!40}0.017 & \cellcolor{orange!40}0.015 & \cellcolor{orange!40}0.030 & \cellcolor{orange!40}0.005 & \cellcolor{yellow!40}0.094 & \cellcolor{orange!40}0.027 \\
3DGRT~\cite{3dgrt2024}* & 0.016 & 0.047 & \cellcolor{yellow!40}0.013 & 0.024 & \cellcolor{yellow!40}0.016 & 0.046 & 0.009 & 0.123 & 0.037 \\
NeuRBF\cite{chen2023neurbf}*  & 0.016 & 0.061 & 0.016 & \cellcolor{yellow!40}0.021 & \cellcolor{orange!40}0.015 & 0.032 & 0.008 & 0.114 & 0.035 \\
\hdashline
RayGauss~\cite{raygauss}* & \cellcolor{red!40}0.009 & \cellcolor{red!40}0.030 & \cellcolor{orange!40}0.011 & \cellcolor{red!40}0.015 & \cellcolor{red!40}0.012 & \cellcolor{red!40}0.026 & \cellcolor{red!40}0.004 & \cellcolor{orange!40}0.088 & \cellcolor{red!40}0.024 \\
    \hline
RayGaussX (ours) & \cellcolor{red!40}0.009 & \cellcolor{red!40}0.030 & \cellcolor{red!40}0.010 & \cellcolor{red!40}0.015 & \cellcolor{red!40}0.012 & \cellcolor{red!40}0.026 & \cellcolor{red!40}0.004 & \cellcolor{orange!40}0.088 & \cellcolor{red!40}0.024\\
  \end{tabular}
  }
  \caption{\textbf{PSNR, SSIM and LPIPS (with VGG network) scores on the NeRF Synthetic dataset~\cite{Neural_Radiance_Fields}.} All methods are trained on the train set with full-resolution images (800x800 pixels) and evaluated on the test set with full-resolution images (800x800 pixels). Methods marked * were retrained on an NVIDIA RTX4090.}
  \label{tab:detailed_blender}
\end{table*}

\begin{table*}[ht]
  \centering
  \resizebox{0.8\linewidth}{!}{%
  \begin{tabular}{l | c c c c c c c c | c }
   & \multicolumn{8}{c}{\textbf{PSNR ↑}} \\
    & Bike & Life & Palace & Robot & Space & Steam & Toad & Wine & \textbf{Avg.}\\
    \hline
TensoRF~\cite{TensoRF} & 39.23 & 34.51 & 37.56 & 38.26 & 38.60 & 37.87 & 34.85 & 31.32 & 36.53 \\
3D-GS~\cite{3D_Gaussian_Splatting}* & 40.76 & 33.19 & 38.89 & 39.16 & 36.80 & 37.67 & \cellcolor{orange!40}37.33 & 32.76 & 37.07 \\
Mip-Splatting~\cite{Yu2023MipSplatting}* & 40.34 & 35.06 & 39.00 & 39.34 & 37.11 & \cellcolor{yellow!40}38.42 & \cellcolor{yellow!40}36.64 & 32.30 & 37.28\\
NeuRBF~\cite{chen2023neurbf} & 40.71 & 36.08 & 38.93 & 39.13 & \cellcolor{red!40}40.44 & 38.35 & 35.73 & \cellcolor{yellow!40}32.99 & 37.80 \\
Spec-Gaussian~\cite{Spec-gaussian}* & \cellcolor{yellow!40}40.93 & \cellcolor{yellow!40}36.17 & \cellcolor{yellow!40}39.39 & \cellcolor{yellow!40}39.64 & \cellcolor{orange!40}40.12 & 38.08 & 36.63 & 32.82 & \cellcolor{yellow!40}37.97 \\
\hdashline
RayGauss~\cite{raygauss}* & \cellcolor{orange!40}41.25 & \cellcolor{orange!40}36.21 & \cellcolor{red!40}40.30 & \cellcolor{red!40}40.37 & 40.03 & \cellcolor{orange!40}39.08 & \cellcolor{red!40}38.39 & \cellcolor{red!40}34.12 & \cellcolor{orange!40}38.72\\
    \hline
RayGaussX (ours) & \cellcolor{red!40}41.34 & \cellcolor{red!40}36.39 & \cellcolor{orange!40}40.28 & \cellcolor{orange!40}40.32 & \cellcolor{yellow!40}40.11 & \cellcolor{red!40}39.12 & \cellcolor{red!40}38.39 & \cellcolor{orange!40}34.01 & \cellcolor{red!40}38.75\\
  \end{tabular}
  }
  
\bigskip
   
  \resizebox{0.8\linewidth}{!}{%
  \begin{tabular}{l | c c c c c c c c | c }
   & \multicolumn{8}{c}{\textbf{SSIM ↑}} \\
    & Bike & Life & Palace & Robot & Space & Steam & Toad & Wine & \textbf{Avg.}\\
    \hline
TensoRF~\cite{TensoRF} & 0.993 & 0.968 & 0.979 & \cellcolor{yellow!40}0.994 & 0.989 & 0.991 & 0.978 & 0.961 & 0.982 \\
3D-GS~\cite{3D_Gaussian_Splatting}* & \cellcolor{yellow!40}0.994 & 0.979 & 0.983 & \cellcolor{yellow!40}0.994 & 0.991 & \cellcolor{yellow!40}0.993 & \cellcolor{orange!40}0.985 & 0.975 & 0.987 \\
Mip-Splatting~\cite{Yu2023MipSplatting}* &  \cellcolor{orange!40}0.995 & \cellcolor{yellow!40}0.982 & 0.985 & \cellcolor{red!40}0.996 & 0.992 & \cellcolor{orange!40}0.994 & \cellcolor{orange!40}0.985 & \cellcolor{orange!40}0.978 & 0.988 \\
NeuRBF\cite{chen2023neurbf} & \cellcolor{orange!40}0.995 & 0.977 & 0.985 & \cellcolor{orange!40}0.995 & \cellcolor{yellow!40}0.993 & \cellcolor{yellow!40}0.993 & \cellcolor{yellow!40}0.983 & 0.972 & 0.987 \\
Spec-Gaussian~\cite{Spec-gaussian}* & \cellcolor{orange!40}0.995 & \cellcolor{yellow!40}0.982 & \cellcolor{yellow!40}0.986 & \cellcolor{red!40}0.996 & \cellcolor{orange!40}0.994 & \cellcolor{orange!40}0.994 & \cellcolor{orange!40}0.985 & \cellcolor{yellow!40}0.977 & \cellcolor{yellow!40}0.989 \\
\hdashline
RayGauss~\cite{raygauss}* & \cellcolor{red!40}0.996 & \cellcolor{orange!40}0.983 & \cellcolor{orange!40}0.987 & \cellcolor{red!40}0.996 & \cellcolor{orange!40}0.994 & \cellcolor{orange!40}0.994 & \cellcolor{red!40}0.989 & \cellcolor{red!40}0.981 & \cellcolor{orange!40}0.990\\
    \hline
RayGaussX (ours) & \cellcolor{red!40}0.996 & \cellcolor{red!40}0.984 & \cellcolor{red!40}0.989 & \cellcolor{red!40}0.996 & \cellcolor{red!40}0.995 & \cellcolor{red!40}0.995 & \cellcolor{red!40}0.989 & \cellcolor{red!40}0.981 & \cellcolor{red!40}0.991\\
  \end{tabular}
  }

\bigskip

  \resizebox{0.8\linewidth}{!}{%
  \begin{tabular}{l | c c c c c c c c | c }
   & \multicolumn{8}{c}{\textbf{LPIPS ↓}} \\
    & Bike & Life & Palace & Robot & Space & Steam & Toad & Wine & \textbf{Avg.}\\
    \hline
TensoRF~\cite{TensoRF} &  0.010 & 0.048 & 0.022 & 0.010 & 0.020 & 0.017 & 0.031 & 0.051 & 0.026 \\
3D-GS~\cite{3D_Gaussian_Splatting}* & \cellcolor{yellow!40}0.005 & 0.028 & 0.017 & \cellcolor{red!40}0.006 & \cellcolor{yellow!40}0.009 & \cellcolor{orange!40}0.007 & 0.018 & 0.025 & 0.014 \\
Mip-Splatting~\cite{Yu2023MipSplatting}* & \cellcolor{yellow!40}0.005 & \cellcolor{yellow!40}0.024 & \cellcolor{yellow!40}0.015 & \cellcolor{orange!40}0.007 & 0.010 & \cellcolor{yellow!40}0.008 & 0.019 & 0.022 & 0.014 \\
NeuRBF\cite{chen2023neurbf} & 0.006 & 0.036 & 0.016 & \cellcolor{yellow!40}0.009 & 0.011 & 0.011 & 0.025 & 0.036 & 0.019 \\
Spec-Gaussian~\cite{Spec-gaussian}* & \cellcolor{orange!40}0.004 & \cellcolor{orange!40}0.022 & \cellcolor{orange!40}0.013 & \cellcolor{red!40}0.006 & \cellcolor{orange!40}0.007 & \cellcolor{orange!40}0.007 & \cellcolor{yellow!40}0.016 & \cellcolor{yellow!40}0.020 & \cellcolor{yellow!40}0.012 \\
\hdashline
RayGauss~\cite{raygauss}* & \cellcolor{red!40}0.003 & \cellcolor{orange!40}0.022 & \cellcolor{red!40}0.011 & \cellcolor{red!40}0.006 & \cellcolor{orange!40}0.007 & \cellcolor{red!40}0.006 & \cellcolor{orange!40}0.012 & \cellcolor{red!40}0.017 & \cellcolor{orange!40}0.011\\
    \hline
RayGaussX (ours) & \cellcolor{red!40}0.003 & \cellcolor{red!40}0.021 & \cellcolor{red!40}0.011 & \cellcolor{red!40}0.006 & \cellcolor{red!40}0.006 & \cellcolor{red!40}0.006 & \cellcolor{red!40}0.011 & \cellcolor{orange!40}0.018 & \cellcolor{red!40}0.010\\
  \end{tabular}
  }
  \caption{\textbf{PSNR, SSIM and LPIPS (with VGG network) scores on the NSVF Synthetic dataset~\cite{NSVF}.} All methods are trained on the train set with full-resolution images (800x800 pixels) and evaluated on the test set with full-resolution images (800x800 pixels). Methods marked * were retrained on an NVIDIA RTX4090.}
  \label{tab:detailed_nsvf}
\end{table*}

\begin{table*}[ht]
  \centering
  \resizebox{0.9\linewidth}{!}{%
  \begin{tabular}{l | c c c c | c c c c c | c }
  & \multicolumn{9}{c}{\textbf{PSNR ↑}} \\
    & bonsai & counter & kitchen & room & bicycle & flowers & garden & stump & treehill & \textbf{Avg.}\\
    \hline
Instant-NGP~\cite{Instant_NGP} & 30.69 & 26.69 & 29.48 & 29.69 & 22.17 & 20.65 & 25.07 & 23.47 & 22.37 & 25.59  \\
3DGRT~\cite{3dgrt2024}* & 31.93 & 28.51 & 29.67 & 30.68 & 24.74 & 21.36 & 26.82 & 26.25 & 22.05 & 26.89  \\
Mip-NeRF360~\cite{Mip_NeRF_360} & 33.46 & 29.55 & 32.23 & 31.63 & 24.37 & 21.73 & 26.98 & 26.40 & 22.87 & 27.69 \\
3D-GS~\cite{3D_Gaussian_Splatting}* & 32.62 & 29.17 & 31.39 & 31.96 & 25.65 & 21.77 & 27.62 & 27.00 & 23.00 & 27.80 \\
Mip-Splatting~\cite{Yu2023MipSplatting}* & 32.42 & 29.41 & 31.85 & 31.68 & \cellcolor{red!40}25.97 & \cellcolor{yellow!40}22.08 & 27.97 & \cellcolor{orange!40}27.31 & 22.71 & 27.93 \\
Spec-Gaussian~\cite{Spec-gaussian}* & 33.26 & \cellcolor{yellow!40}29.86 & \cellcolor{yellow!40}32.45 & \cellcolor{orange!40}32.15 & \cellcolor{yellow!40}25.81 & 21.52 & 28.00 & \cellcolor{yellow!40}27.17 & 22.23 & 28.05 \\
Zip-NeRF~\cite{zipnerf}* & \cellcolor{red!40}34.79 & 29.12 & 32.36 & \cellcolor{red!40}33.04 & \cellcolor{orange!40}25.85 & \cellcolor{red!40}22.33 & \cellcolor{orange!40}28.22 & \cellcolor{red!40}27.35 & \cellcolor{red!40}23.95 & \cellcolor{red!40}28.56 \\
\hdashline
RayGauss~\cite{raygauss}* & \cellcolor{yellow!40}33.91 & \cellcolor{orange!40}30.56 & \cellcolor{orange!40}32.83 & 31.83 & 25.51 & 21.85 & \cellcolor{yellow!40}28.06 & 26.33 & \cellcolor{orange!40}23.18 & \cellcolor{yellow!40}28.23 \\
    \hline
RayGaussX (ours) & \cellcolor{orange!40}34.06 & \cellcolor{red!40}30.64 & \cellcolor{red!40}32.92 & \cellcolor{yellow!40}32.11 & 25.71 & \cellcolor{orange!40}22.31 & \cellcolor{red!40}28.35 & 26.66 & \cellcolor{yellow!40}23.15 & \cellcolor{orange!40}28.43\\
  \end{tabular}
  }

\bigskip

  \resizebox{0.9\linewidth}{!}{%
  \begin{tabular}{l | c c c c | c c c c c | c }
  & \multicolumn{9}{c}{\textbf{SSIM ↑}} \\
    & bonsai & counter & kitchen & room & bicycle & flowers & garden & stump & treehill & \textbf{Avg.}\\
    \hline
Instant-NGP~\cite{Instant_NGP} & 0.906 & 0.817 & 0.858 & 0.871 & 0.512 & 0.486 & 0.701 & 0.594 & 0.542 & 0.699   \\
3DGRT~\cite{3dgrt2024}* & 0.941 & 0.904 & 0.914 & 0.914 & 0.744 & 0.612 & 0.848 & 0.767 & 0.621 & 0.807  \\
Mip-NeRF360~\cite{Mip_NeRF_360} & 0.941 & 0.894 & 0.920 & 0.913 & 0.685 & 0.583 & 0.813 & 0.744 & 0.632 & 0.792 \\
3D-GS~\cite{3D_Gaussian_Splatting}* & 0.947 & 0.916 &  0.933 & 0.929 & 0.777 & 0.618 & 0.868 & 0.783 & 0.654 & 0.825 \\
Mip-Splatting~\cite{Yu2023MipSplatting}* & 0.952 & \cellcolor{yellow!40}0.921 & 0.937 & 0.933 & \cellcolor{red!40}0.803 & \cellcolor{red!40}0.656 & \cellcolor{orange!40}0.885 & \cellcolor{red!40}0.801 & 0.657 & \cellcolor{orange!40}0.838  \\
Spec-Gaussian~\cite{Spec-gaussian}* & \cellcolor{yellow!40}0.954 & \cellcolor{orange!40}0.923 & \cellcolor{yellow!40}0.938 & \cellcolor{yellow!40}0.935 & \cellcolor{yellow!40}0.796 & \cellcolor{yellow!40}0.647 & \cellcolor{yellow!40}0.881 & \cellcolor{orange!40}0.796 & 0.645 & \cellcolor{yellow!40}0.835 \\
Zip-NeRF~\cite{zipnerf}* &  0.952 & 0.905 & 0.929 & 0.929 & 0.772 & 0.637 & 0.863 & 0.788 & \cellcolor{orange!40}0.674 & 0.828 \\
\hdashline
RayGauss~\cite{raygauss}* & \cellcolor{orange!40}0.957 & \cellcolor{red!40}0.932 & \cellcolor{orange!40}0.942 & \cellcolor{orange!40}0.936 & 0.782 & 0.634 & 0.879 & 0.775 & \cellcolor{yellow!40}0.672 & 0.834\\
    \hline
RayGaussX (ours) & \cellcolor{red!40}0.959 & \cellcolor{red!40}0.932 & \cellcolor{red!40}0.943 & \cellcolor{red!40}0.939 & \cellcolor{orange!40}0.797 & \cellcolor{orange!40}0.655 & \cellcolor{red!40}0.887 & \cellcolor{yellow!40}0.789 & \cellcolor{red!40}0.677 & \cellcolor{red!40}0.842 \\
  \end{tabular}
  }

\bigskip
  
  \resizebox{0.9\linewidth}{!}{%
  \begin{tabular}{l | c c c c | c c c c c | c }
  & \multicolumn{9}{c}{\textbf{LPIPS ↓}} \\
    & bonsai & counter & kitchen & room & bicycle & flowers & garden & stump & treehill & \textbf{Avg.}\\
    \hline
Instant-NGP~\cite{Instant_NGP} & 0.205 & 0.306 & 0.195 & 0.261 & 0.446 & 0.441 & 0.257 & 0.421 & 0.450 & 0.331 \\
3DGRT~\cite{3dgrt2024}* & 0.249 & 0.258 & 0.169 & 0.295 & 0.254 & 0.335 & 0.145 & 0.251 & 0.372 & 0.259  \\
Mip-NeRF360~\cite{Mip_NeRF_360} & 0.176 & 0.204 & 0.127 & 0.211 & 0.301 & 0.344 & 0.170 & 0.261 & 0.339 & 0.237 \\
3D-GS~\cite{3D_Gaussian_Splatting}* & 0.178 & 0.181 & 0.114 & 0.195 & 0.213 & 0.336 & 0.115 & 0.211 & 0.326 & 0.208\\
Mip-Splatting~\cite{Yu2023MipSplatting}* & \cellcolor{yellow!40}0.161 & 0.167 & 0.109 & \cellcolor{yellow!40}0.177 & \cellcolor{red!40}0.164 & \cellcolor{orange!40}0.266 & \cellcolor{orange!40}0.090 & \cellcolor{red!40}0.182 & \cellcolor{yellow!40}0.270 & \cellcolor{red!40}0.176 \\
Spec-Gaussian~\cite{Spec-gaussian}* & 0.163 & 0.167 & 0.108 & 0.178 & \cellcolor{orange!40}0.167 & \cellcolor{red!40}0.263 & \cellcolor{yellow!40}0.093 & \cellcolor{orange!40}0.185 & \cellcolor{yellow!40}0.270 & \cellcolor{orange!40}0.177 \\
Zip-NeRF~\cite{zipnerf}* & \cellcolor{red!40}0.134 & \cellcolor{yellow!40}0.160 & \cellcolor{red!40}0.102 & \cellcolor{red!40}0.159 & 0.198 & \cellcolor{yellow!40}0.276 & 0.117 & \cellcolor{yellow!40}0.199 & \cellcolor{red!40}0.239 & \cellcolor{red!40}0.176 \\
\hdashline
RayGauss~\cite{raygauss}* & 0.163 & \cellcolor{orange!40}0.157 & \cellcolor{yellow!40}0.105 & 0.178 & 0.205 & 0.306 & 0.103 & 0.221 & 0.297 & \cellcolor{yellow!40}0.193\\
    \hline
RayGaussX (ours) & \cellcolor{orange!40}0.154 & \cellcolor{red!40}0.156 & \cellcolor{orange!40}0.103 & \cellcolor{orange!40}0.171 & \cellcolor{yellow!40}0.177 & 0.285 & \cellcolor{red!40}0.090 & 0.200 & \cellcolor{orange!40}0.262 & \cellcolor{orange!40}0.177\\
  \end{tabular}
  }
  \caption{\textbf{PSNR, SSIM and LPIPS (with VGG network) scores on the Mip-NeRF360 dataset~\cite{Neural_Radiance_Fields}.} All methods are trained and tested on downsampled images by a factor of 2 in indoor (\textit{bonsai}, \textit{counter}, \textit{kitchen}, \textit{room}) and 4 in outdoor (\textit{bicycle}, \textit{flowers}, \textit{garden}, \textit{stump}, \textit{treehill}). Methods marked * were retrained on an NVIDIA RTX4090.}
  \label{tab:detailed_mipnerf360}
\end{table*}

\begin{table*}[ht]
  \centering
  \resizebox{0.7\linewidth}{!}{%
  \begin{tabular}{l | c c | c || c  c  | c }
  & \multicolumn{6}{c}{\textbf{PSNR ↑}} \\
    &  train & truck & \textbf{Avg.} & drjohnson & playroom & \textbf{Avg.}\\
    \hline
Instant-NGP~\cite{Instant_NGP}  & 20.46 & 23.38 & 21.92 & 28.26 & 21.67 & 24.97 \\
3DGRT~\cite{3dgrt2024}* & 21.06 & 24.41 & 22.74 & 29.16 & 30.33 & 29.74  \\
Mip-NeRF360~\cite{Mip_NeRF_360} & 19.52 & 24.91 & 22.22 & 29.14 & 29.66 & 29.40 \\
3D-GS~\cite{3D_Gaussian_Splatting}* & \cellcolor{yellow!40}22.03 & \cellcolor{yellow!40}25.41 & \cellcolor{yellow!40}23.72 & 29.52 & 30.32 & 29.92 \\
Mip-Splatting~\cite{Yu2023MipSplatting}* & 21.78 & \cellcolor{red!40}25.66 & \cellcolor{yellow!40}23.72 & 28.83 & 30.19 & 29.51 \\
Spec-Gaussian~\cite{Spec-gaussian}* & \cellcolor{orange!40}22.10 & \cellcolor{orange!40}25.62 & \cellcolor{red!40}23.86 & 29.03 & 30.31 & 29.67 \\
Zip-NeRF~\cite{zipnerf}* & 10.53 & 11.16 & 10.85 & \cellcolor{red!40}30.28 & \cellcolor{red!40}31.23 & \cellcolor{red!40}30.76 \\
\hdashline
RayGauss~\cite{raygauss}* & 21.80 & 24.59 & 23.20 & \cellcolor{yellow!40}29.74 & \cellcolor{orange!40}30.85 & \cellcolor{yellow!40}30.30 \\
    \hline
RayGaussX (ours) & \cellcolor{red!40}22.24 & 25.27 & \cellcolor{orange!40}23.76 & \cellcolor{orange!40}29.85 & \cellcolor{yellow!40}30.79 & \cellcolor{orange!40}30.32 \\
  \end{tabular}
  }

\bigskip

  \resizebox{0.7\linewidth}{!}{%
  \begin{tabular}{l | c c | c || c  c  | c }
  & \multicolumn{6}{c}{\textbf{SSIM ↑}} \\
    &  train & truck & \textbf{Avg.} & drjohnson & playroom & \textbf{Avg.}\\
    \hline
Instant-NGP~\cite{Instant_NGP} & 0.689 & 0.800 & 0.745 & 0.854 & 0.779 & 0.817 \\
3DGRT~\cite{3dgrt2024}* & 0.814 & 0.873 & 0.844 & \cellcolor{yellow!40}0.904 & 0.906 & 0.905  \\
Mip-NeRF360~\cite{Mip_NeRF_360} &  0.660 & 0.857 & 0.759 & 0.901 & 0.900 & 0.901 \\
3D-GS~\cite{3D_Gaussian_Splatting}* & 0.816 & 0.880 & 0.848 & \cellcolor{yellow!40}0.904 & 0.905 & 0.905 \\
Mip-Splatting~\cite{Yu2023MipSplatting}* & \cellcolor{orange!40}0.827 & \cellcolor{orange!40}0.893 & \cellcolor{orange!40}0.860 & 0.899 & 0.907 & 0.903 \\
Spec-Gaussian~\cite{Spec-gaussian}* & \cellcolor{yellow!40}0.823 & \cellcolor{yellow!40}0.888 & \cellcolor{yellow!40}0.856 & 0.902 & \cellcolor{yellow!40}0.910 & \cellcolor{yellow!40}0.906 \\
Zip-NeRF~\cite{zipnerf}* & 0.300 & 0.351 & 0.326 & \cellcolor{orange!40}0.912 & \cellcolor{orange!40}0.915 & \cellcolor{orange!40}0.914 \\
\hdashline
RayGauss~\cite{raygauss}* & 0.814 & 0.883 & 0.849 & \cellcolor{orange!40}0.912 & \cellcolor{red!40}0.916 & \cellcolor{orange!40}0.914 \\
    \hline
RayGaussX (ours) & \cellcolor{red!40}0.834 & \cellcolor{red!40}0.896 & \cellcolor{red!40}0.865 & \cellcolor{red!40}0.913 & \cellcolor{red!40}0.916 & \cellcolor{red!40}0.915 \\
  \end{tabular}
  }

\bigskip
  
  \resizebox{0.7\linewidth}{!}{%
  \begin{tabular}{l | c c | c || c  c  | c }
  & \multicolumn{6}{c}{\textbf{LPIPS ↓}} \\
    &  train & truck & \textbf{Avg.} & drjohnson & playroom & \textbf{Avg.}\\
    \hline
Instant-NGP~\cite{Instant_NGP}  &  0.360 & 0.249 & 0.305 & 0.352 & 0.428 & 0.390 \\
3DGRT~\cite{3dgrt2024}* & 0.223 & 0.170 & 0.197 & 0.316 & 0.313 & 0.315  \\
Mip-NeRF360~\cite{Mip_NeRF_360} & 0.354 & 0.159 & 0.257 & 0.237 & 0.252 & 0.245 \\
3D-GS~\cite{3D_Gaussian_Splatting}* & 0.205 & 0.150 & 0.178 & \cellcolor{orange!40}0.241 & 0.246 & 0.244\\
Mip-Splatting~\cite{Yu2023MipSplatting}* & \cellcolor{orange!40}0.190 & \cellcolor{orange!40}0.123 & \cellcolor{orange!40}0.157 & 0.246 & \cellcolor{orange!40}0.239 & \cellcolor{yellow!40}0.243 \\
Spec-Gaussian~\cite{Spec-gaussian}* & \cellcolor{yellow!40}0.196 & 0.135 & \cellcolor{yellow!40}0.166 & 0.245 & 0.245 & 0.245 \\
Zip-NeRF~\cite{zipnerf}* & 0.658 & 0.664 & 0.661 & \cellcolor{red!40}0.217 & \cellcolor{red!40}0.201 & \cellcolor{red!40}0.209 \\
\hdashline
RayGauss~\cite{raygauss}* & 0.201 & \cellcolor{yellow!40}0.132 & 0.167 & 0.238 & 0.245 & \cellcolor{orange!40}0.242 \\
    \hline
RayGaussX (ours) & \cellcolor{red!40}0.183 & \cellcolor{red!40}0.117 & \cellcolor{red!40}0.150 & \cellcolor{yellow!40}0.242 & \cellcolor{yellow!40}0.242 & \cellcolor{orange!40}0.242 \\
  \end{tabular}
  }
  \caption{\textbf{PSNR, SSIM and LPIPS (with VGG network) scores on the Tanks\&Temples~\cite{Knapitsch2017} and Deep Blending~\cite{Hedman2018} datasets.} All methods are trained and tested on full images. Methods marked * were retrained on an NVIDIA RTX4090.}
  \label{tab:detailed_tandt_db}
\end{table*}

\FloatBarrier

\clearpage
\newpage

{
    \small
    \bibliographystyle{ieeenat_fullname}
    \bibliography{main}
}